 \documentclass[10pt,twocolumn,letterpaper]{article}

\usepackage{cvpr}
\usepackage{times}
\usepackage{epsfig}
\usepackage{graphicx}
\usepackage{amsmath}
\usepackage{amssymb}

\graphicspath{{./figs/}}

\usepackage{soul}
\usepackage{tabularx}

\newcommand{\norm}[1]{\left\| {#1} \right\|}

\newcommand{\R}{\ensuremath \mathbb{R}}

\renewcommand{\epsilon}{\varepsilon}
\def\utilde#1{\mathord{\vtop{\ialign{##\crcr
$\hfil\displaystyle{#1}\hfil$\crcr\noalign{\kern1.5pt\nointerlineskip}
$\hfil\tilde{}\hfil$\crcr\noalign{\kern1.5pt}}}}}
\usepackage[pagebackref=true,breaklinks=true,letterpaper=true,colorlinks,bookmarks=false]{hyperref}

\usepackage[title]{appendix}

\cvprfinalcopy %

\def\cvprPaperID{4944} %

\ifcvprfinal\fi
\begin{document}

\title{Convolutional Recurrent Network for Road Boundary Extraction}

\author{
	Justin Liang$^{1}$\thanks{Equal contribution.} \quad Namdar Homayounfar$^{1,2*}$\\
	Wei-Chiu Ma$^{1,3}$ \quad Shenlong Wang$^{1,2}$ \quad Raquel Urtasun$^{1,2}$\\
	$^{1}$Uber Advanced Technologies Group \quad $^{2}$University of Toronto \quad $^{3}$ MIT\\
	\small\texttt{justin.j.w.liang@gmail.com, namdar.homayounfar@mail.utoronto.ca} \\
	\small\texttt{weichium@mit.edu, slwang@cs.toronto.edu, urtasun@cs.toronto.edu}
}

\maketitle
\thispagestyle{empty}

\begin{abstract}
Creating high definition maps that contain precise information of static elements of the scene is of utmost importance for enabling self driving cars to drive safely. In this paper, we tackle the problem of drivable road boundary extraction from LiDAR and camera imagery. 
Towards this goal, we design a structured model where a fully convolutional network  obtains deep features encoding  the location and direction of road boundaries and then, a convolutional recurrent network outputs a polyline representation for each one of them.  Importantly, our method is fully automatic and does not require a user in the loop. We showcase the effectiveness of our method on a large North American city where we obtain  perfect topology of road boundaries $99.3\%$ of the time at a high precision and recall. 
\end{abstract}

\section{Introduction}

High definition maps (HD maps) contain useful information about the semantics of the static part of the  scene. They are employed by most self-driving cars as an additional sensor in order to help localization, \cite{deep-gill, ma2017find}, perception \cite{liang2018deep,casas2018intentnet} and motion planning. 
Drawing the maps, is however, a laborious process where annotators look at overhead views of the cities and draw one by one all the elements of the scene. This is an  expensive and time consuming process preventing mapping from being done at scale. 

Crowd-source efforts such as OpenStreetMaps provide scale, but are not very reliable or precise enough for the safe navigation of the self driving cars. 
Although they cover most of the globe and provide valuable data such as the road network topology, speed limits, traffic signals, building contours,  etc, they suffer from low resolution and are not perfectly aligned with respect to the actual physical clues in the world. This is due to the nature of the map topology creation process which is obtained from GPS trajectories or satellite imagery that could have errors in meters as it is typically very low resolution.

\begin{figure}[t]
	\vspace{0.5cm}
	\includegraphics[width=\linewidth]{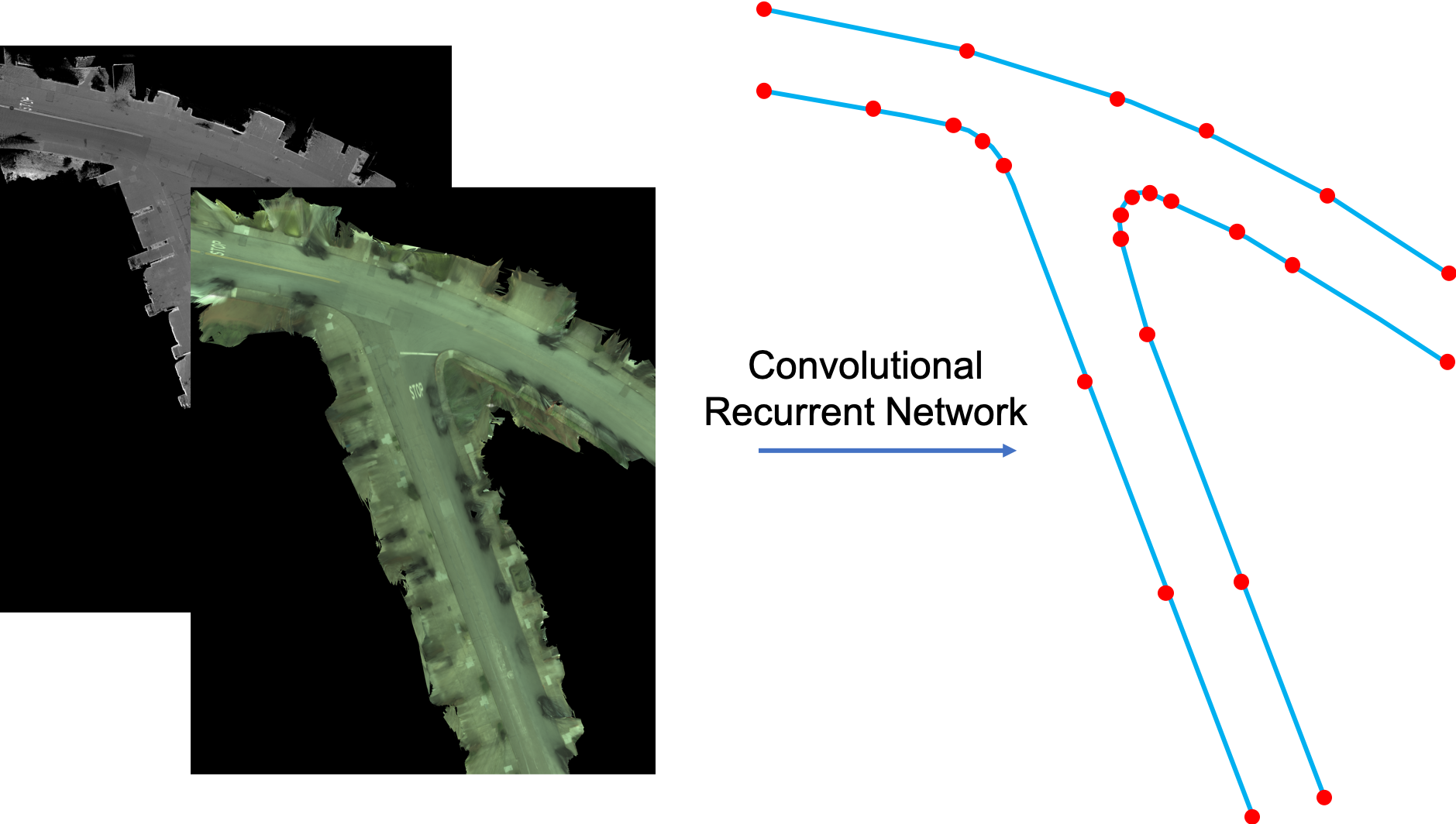}  
	
	\caption{{\bf Overview:} Our Convolutional Reccurrent Network takes as input overhead camera and LiDAR imagery ({\bf left}) and outputs a structured polyline for each road boundary ({\bf right}) that is utilized in creating HD maps for autonomous driving.}
	\label{fig:teaser}
	\vspace{-4mm}
\end{figure}

Many efforts have been devoted to automate the map creation process to achieve scale. Most early approaches treat the problem as semantic segmentation, either from aerial images \cite{mnih2010learning,mnih2012learning,marmanis2016semantic,marmanis2018classification} or from first person views, where the goal is to capture free space \cite{kong2010general,cheng2006lane,wedel2009b, alvarez2012road, kuhnl2012spatial}. However, these techniques do not provide a structured representation that is required in order to be consumed by most self driving software stacks.

Automatically estimating the road topology from aerial imagery has been tackled in \cite{mattyus2017deeproadmapper,ventura2018iterative,bastani2018roadtracer}. In these works, a graph of the road network with nodes being intersections and edges corresponding to the streets connecting them is extracted. Although very useful for routing purposes, these graphs still lack the fine detail and accuracy needed for a safe localization and motion planning of an autonomous car.  

\begin{figure*}[t]
\vspace{-0.5cm}
	\includegraphics[width=\linewidth]{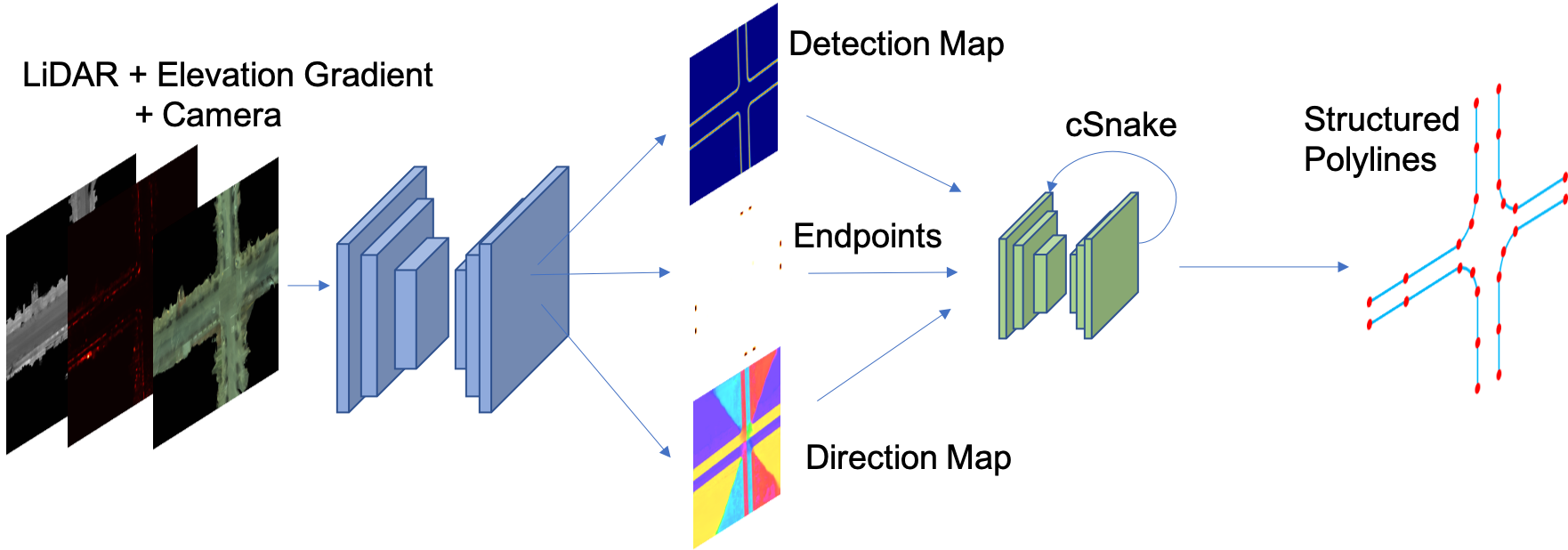}  
	
	\caption{ {\bf Model:} Our model takes as input overhead LiDAR and camera imagery as well as the gradient of the LiDAR's elevation value. Next, a convolutional network outputs three feature maps: A truncated inverse distance transform of the location of road boundaries ({\bf Detection Map}), their endpoints ({\bf Endpoints}) and the vector field of normalized normals to the road boundaries ({\bf Direction Map}) shown here as a flow field \cite{color_code}. Finally, a convolutional recurrent network ({\bf cSnake}) takes these deep features and outputs a structured polyline corresponding to each road boundary. }
	\label{fig:model}
	\vspace{-4mm}
\end{figure*}

In contrast, in this paper we tackle the problem of estimating drivable regions from both LiDAR and camera data, which  provide a very high definition information of the surroundings. 
Towards this goal, we employ convolutional neural networks to predict a set of visual cues that are employed by a convolutional recurrent network to output a variable number of boundary regions of variable size. Importantly, our approach is fully automatic and does not require a user in the loop.  Inspired by how humans performed this task, our convolutional recurrent network outputs a structured polyline corresponding to each road boundary one vertex at a time. 

We demonstrate the effectiveness of our work on a large dataset of a North American city composed of overhead LiDAR and camera imagery of streets and intersections. In comparison to the baselines where a road boundary could be estimated with multiple smaller segments, we predict perfect topology for a road boundary $99.3\%$ of the time with high precision and recall of $87.3\%$ and $87.1\%$ respectively at 5 pixels away. We also perform extensive ablation studies that justify our choices of input representations, training regimes and network architecture. Finally, we propose a novel metric that captures not only precision and recall but also a measure of connectivity of the predictions corresponding to a road boundary.

\section{Related Work}

In the past few decades the computer vision and sensing communities have actively developed a myriad of methods that perform semantic segmentation tasks from  aerial and satellite imagery. We refer the reader to \cite{richards_2013} for a complete introduction of classical approaches. More recently, deep neural networks \cite{mena2005automatic,mnih2010learning,mnih2012learning,marmanis2016semantic,marmanis2018classification} have been applied to this task with considerable success. Such output is however not directly usable by self driving vehicles which require a structured representation instead. 

\begin{table*}[t!]
\vspace{-0.5cm}
\centering
  \begin{tabular}{|l|*{16}{c|}}
  \cline{2-14}
  \multicolumn{1}{c|}{} & \multicolumn{4}{c}{Precision at (px)} & \multicolumn{4}{|c|}{Recall at (px)} & \multicolumn{4}{c|}{F1 score at (px)} & Conn\\ 
  \cline{2-14}
  \multicolumn{1}{c|}{} & 
\multicolumn{1}{c}{2} & \multicolumn{1}{|c}{3} & \multicolumn{1}{|c}{5} & \multicolumn{1}{|c}{10} & 
\multicolumn{1}{|c}{2} & \multicolumn{1}{|c}{3} & \multicolumn{1}{|c}{5} & \multicolumn{1}{|c}{10} & 
\multicolumn{1}{|c}{2} & \multicolumn{1}{|c}{3} & \multicolumn{1}{|c}{5} & \multicolumn{1}{|c|}{10} & \\ 
  \hline 

  \textrm{DT} 
    &$47.5$ &$66.0$ &$85.9$  &$\bf{96.2}$
    &$47.6$ &$65.8$ &$84.7$ &$93.8$
    &$47.4$ &$65.6$ &$85.0$  &$\bf{94.6}$  
    &$89.1$\\
  \hline
  \textrm{\bf{Ours}}
  &$\bf{57.3}$ &$\bf{72.9}$ &$\bf{87.3}$  &$94.5$ 
    &$\bf{57.1}$ &$\bf{72.6}$ &$\bf{87.1}$ &$\bf{94.3}$ 
    &$\bf{57.2}$ &$\bf{72.7}$ &$\bf{87.2}$  &$94.4$ 
     &$\bf{99.2}$ \\
  \hline 
    
  \end{tabular}
  \vspace{1mm}
  \caption{This compares the distance transform (DT) baseline with our model. We show the results for all the models at precision, recall and F1 score thresholds of 2, 3, 5, 10 (4cm/px).}
  \label{tab:baseline}
  \vspace{-5mm}
\end{table*}

Research extracting structured semantic and topological information from satellite and aerial imagery, mainly for consumption in geographic information systems, goes back decades to the earliest works of \cite{simonett1970use,Bajcsy1976ComputerRO} in the 70s. In these works, the authors grow a road from pixels to edges to line segments iteratively by using thresholds on simple features obtained from geometric and spectral properties of roads. \cite{fortier1999survey} compiles a comprehensive survey of these earlier approaches. Later on, active contour models (ACM) \cite{kass1988snakes, butenuth2012} were applied to the task of road extraction from aerial imagery \cite{mayer1998multi,laptev2000automatic, rochery2006higher, marikhu2007family}. Here, the authors evolve a snake that captures a road network by minimizing an energy function that specifies geometric and appearance constraints of the roads. The authors in \cite{marcos2018learning} use a deep learning model to define the energy function of an ACM to generate building polygons from aerial imagery but not the road network.  
 \cite{mattyus2015enhancing, mattyus2016hd} apply graphical models on top of deep features  in order to enhance Open Street Maps with semantic information such as the location of sidewalks, parking spots and the number and location of lanes.  In other work \cite{wegner2013higher, wegner2015road, montoya2014mind} extract the road network from aerial images using a conditional random field,  while the works of \cite{mattyus2017deeproadmapper, ventura2018iterative} perform this task by first segmenting the image to road/non-road pixels using a deep network and then performing post process graph optimization. In \cite{bastani2018roadtracer} the authors iteratively grow the road network topology by mixing neural networks and graph search procedures. %
These approaches extract the road network at a coarser scale and are useful for routing applications, however they lack the fine detail of the surroundings  required for the safe navigation of a self driving vehicle.

Predicting the drivable surface is very important for safe navigation of an autonomous vehicle. \cite{mohan2014deep, levi2015stixelnet, yao2015estimating} use graphical models to predict the free space and the road while \cite{tan2006color, lieb2005adaptive, paz2015variational, alvarez2011road,kong2010general,cheng2006lane,wedel2009b, alvarez2012road, kuhnl2012spatial} detect the road using appearance and geometric priors in unsupervised and self-supervised settings.

More recent line of research and industrial work leverage sensors such as camera and LiDAR \cite{moravec1985high, kammel2008lidar} mounted on cars to create HD maps of the environment. In \cite{pollefeys2008detailed, pire2018real, geiger2011stereoscan, barsan2018robust}, multiview and fisheye cameras are used for dense mapping of the static  environment using stereo reconstruction and structure from motion techniques.  In other work \cite{Homayounfar_2018_CVPR,liang2018end} extracted semantic information of the scene such as the precise location and number of the lane boundaries and the crosswalks. These semantics aid the autonomous agent in precise localization and safe navigation. In \cite{Homayounfar_2018_CVPR}, the authors predict a structured representation of lane boundaries in the form of polylines directly from LiDAR point clouds using a recurrent hierarchical network and in \cite{liang2018end}, crosswalks are detected using a deep structured model from top down camera and LiDAR imagery.  \cite{bai2018} fuses LiDAR and camera to perform dense online lane detection. 

In contrast to the aforementioned approaches, in this work we extract road boundaries from LiDAR and camera imagery to create HD maps. Similar to \cite{CastrejonCVPR17, acuna2018efficient}, we use a structured output representation  in the form of polylines. However, unlike them we propose a fully automatic approach and we tackle a very different setting with different sensors and visual cues. %

\vspace{-1mm}

\section{Convolutional Recurrent Road  Extraction}

\begin{table*}[t!]
\vspace{-0.5cm}
\centering
  \begin{tabular}{|l|*{17}{c|}}
  \cline{4-16}
  \multicolumn{3}{c|}{}& \multicolumn{4}{c}{Precision at (px)} & \multicolumn{4}{|c}{Recall at (px)} & \multicolumn{4}{|c|}{F1 Score at (px)} & Conn \\ 
  \cline{1-16}
  L & E & C & 2 & 3 & 5 & 10& 2 & 3 & 5 & 10 & 2 & 3 & 5 & 10 & \\ 
  \hline

  {-} & {-} & {\checkmark} 
  &$42.9$ &$57.6$ &$74.3$  &$86.7$
    &$42.8$ &$57.4$ &$74.0$  &$86.4$
    &$42.9$ &$57.5$ &$74.3$  &$86.6$    
     &$98.8$ \\
  {\checkmark} & {-} & {-} 
  &$44.0$ &$62.2$ &$82.8$  &$93.4$
    &$43.9$ &$62.0$ &$82.7$  &$93.3$
    &$44.0$ &$62.1$ &$82.8$  &$93.3$    
     &$99.2$ \\
  {\checkmark} & {\checkmark} & {-}
  &$51.4$	 &$69.1$ &$86.4$  &$\bf{94.8}$ 
    &$51.3$ &$68.9$ &$86.2$  &$\bf{94.6}$
    &$51.3$ &$69.0$ &$86.2$  &$\bf{94.7}$
     &$99.2$ \\
  {\checkmark} & {\checkmark} & {\checkmark}
  &$\bf{57.3}$ &$\bf{72.9}$ &$\bf{87.3}$  &$94.5$ 
    &$\bf{57.1}$ &$\bf{72.6}$ &$\bf{87.1}$ &$94.3$ 
    &$\bf{57.2}$ &$\bf{72.7}$ &$\bf{87.2}$  &$94.4$ 
     &$\bf{99.2}$ \\
  \hline 
    
  \end{tabular}
  \vspace{1mm}
  \caption{The abbreviated columns are: L (lidar input), E (elevation input),  C (camera input). We show the results for all the models at precision, recall and F1 score thresholded at 2, 3, 5, 10px (4cm/px).}
  \label{tab:compare_inputs}
  \vspace{-5mm}
\end{table*}

High definition maps (HD maps) contain useful information encoding the semantics of the static scene. These maps are typically created by having hundreds of annotators manually label the elements on bird's eye view (BEV) representations of the world. Automating this process is key for achieving self driving cars at scale. 

In this paper we go one step further in this direction, and tackle the problem of estimating drivable regions from both LiDAR and camera data. We encode these drivable regions with  polylines delimiting the road boundaries, as  this is the  typical representations utilized by commercial HD maps. 

Towards this goal, we employ convolutional neural networks to predict a set of visual cues that are employed by a convolutional recurrent network to output a variable number of road boundaries of variable size. 
In particular, our recurrent network attends  to rotated regions of interest in the feature maps and outputs a structured polyline capturing the global topology as well as the fine details of each road boundary.
Next, we first describe the specifics of the feature maps, followed by our convolutional recurrent network.

\subsection{Deep Visual Features}

We now describe how we obtain  deep features that are useful for extracting a globally precise structured polyline representation of the road boundary. 
As input to our system, we take advantage of different sensors such as camera and LiDAR to create a BEV representation of the area of interest. Note that this can contain intersections or straight portions of the road. We also input as an extra channel the gradient of the LiDAR's height value. This input channel is very informative since the drivable and non-drivable regions of the road in a city are mostly flat surfaces at different heights that are separated by a curb.
As shown in our experimental section, these  sources of data are complementary and help the road boundary prediction problem. 
This results in a 5-channel input tensor of size $I \in R^{5 \times H \times W}$ that is fed to the multi-task CNN that predicts three types of feature maps: 
 the location  of the road boundaries encoded as a distance transform, a heatmap encoding the possible location of the endpoints as well as a direction map encoding the direction pointing towards the closest road boundary. 
We refer the reader to Fig. \ref{fig:model} for an illustration of these visual cues, which are explained  in detail below.  %

\paragraph{Road Boundary Dense Detection Map:} To obtain a dense representation of the location of the road boundaries in $I$, we output an inverse truncated distance transform image $S \in \R^{1 \times H \times W}$ that encodes the relative distance of each pixel in $I$ to the closest road boundary \cite{bai2018,liang2018end}, with the road boundary pixels having the highest value and decreasing as we move away.
  In contrast to predicting binary outputs at the road boundary pixels which are very sparse, the truncated inverse distance transforms encodes more information about the locations of the road boundaries.  

\paragraph{Endpoints Heatmap:} We output a heatmap image $E \in \R^{1 \times H \times W}$ encoding the probability of the location of the endpoints of the road boundaries. 
Note that this typically happens at the edges of the image. 

\paragraph{Road Boundary Direction Map:} Finally, we also predict a vector field $D \in \R^{2 \times H \times W}$ of normal directions to the road boundaries. We obtain the ground truth by taking the Sobel derivative of the road boundaries' distance transform image followed by a normalization step. 
This feature map specifies at each pixel the normal direction towards the closest road boundary. The normalization step relieves the network from predicting vectors of arbitrary magnitude. We utilize the direction map as an input to our convolutional recurrent network, as it encourages the polyline vertices to be pulled towards the road boundaries. The direction map is also used in providing the direction of next rotated ROI when evolving the road boundary polyline as we shall explain in section \ref{sec:csnake}.

\paragraph{Network Architecture:} We use an encoder decoder architecture similar to the feature pyramid networks in \cite{lin2016feature, linknet}. This network was chosen for its efficiency and ability to keep spatial information. In particular, there are skip connections between the encoder and decoder that allows for the recovery of lost spatial information which is useful as we use large images in our application. In the encoder stage, each encoder block contains two residual blocks and each residual block contains three dilated convolutional layers. This effectively increases the receptive field to help the network deal with large imagery. In the decoder stage, we have four convolutional layers and a nearest neighbor upsampling of 2x. Prior to each convolutional layer we perform instance normalization followed by a ReLU non-linearity . Our network has three output branches performing pixel wise prediction to output our distance transform, endpoints and direction features. These features all have the same spatial resolution as the input image $I$.

\begin{figure*}[t]
\centering
\includegraphics[width=.3\textwidth]{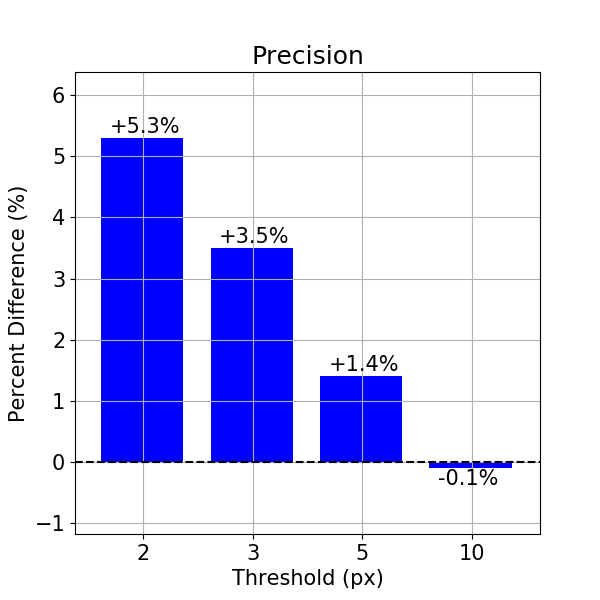} \hskip -15pt
\includegraphics[width=.3\textwidth]{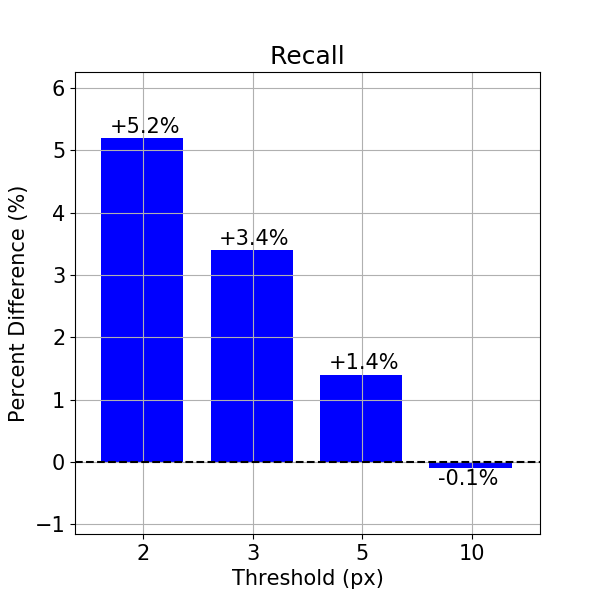} \hskip -15pt
\includegraphics[width=.3\textwidth]{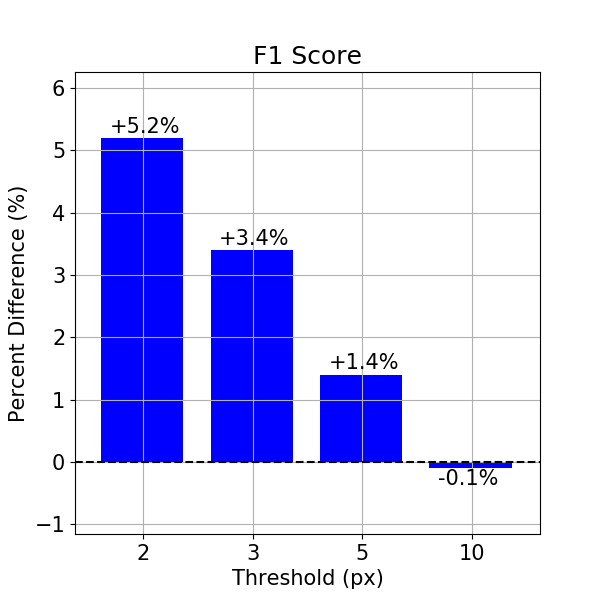} \hskip -15pt
\caption{{\bf Amortized learning:} In this figure we show the percent difference between our models for precision, recall and F1 score trained with and without amortized learning. We see amortized learning significantly improves the result at all thresholds of our metric. Connectivity is the same for both models.}
\label{fig:amortized_learning}
\end{figure*}

\begin{figure*}[t!]
\centering
\includegraphics[width=.3\textwidth]{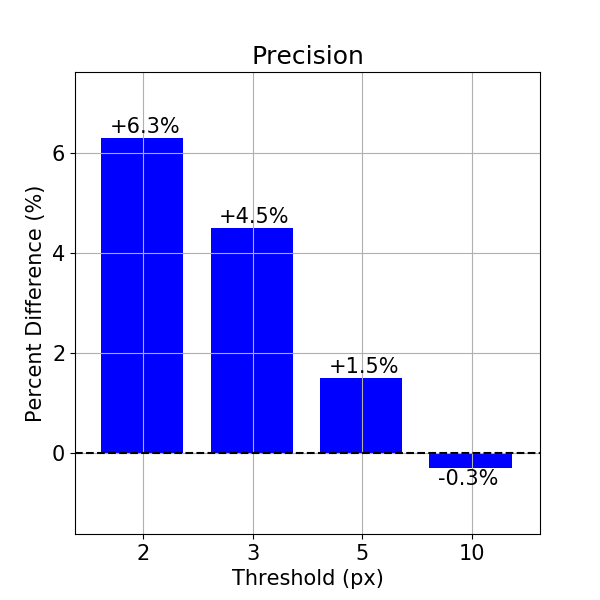} \hskip -15pt
\includegraphics[width=.3\textwidth]{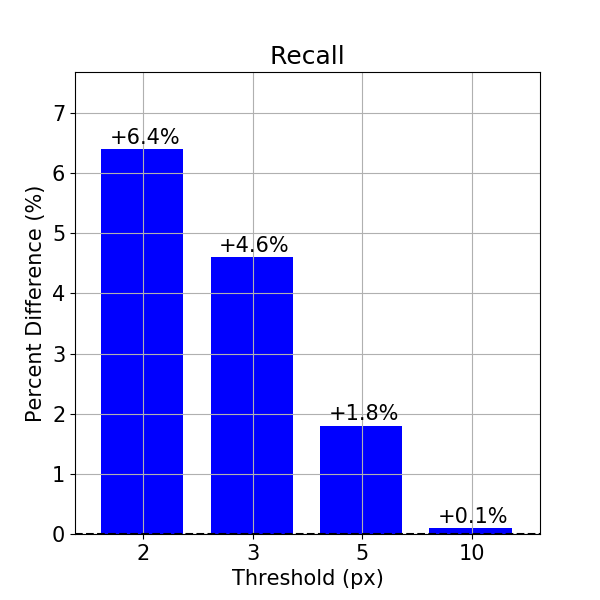} \hskip -15pt
\includegraphics[width=.3\textwidth]{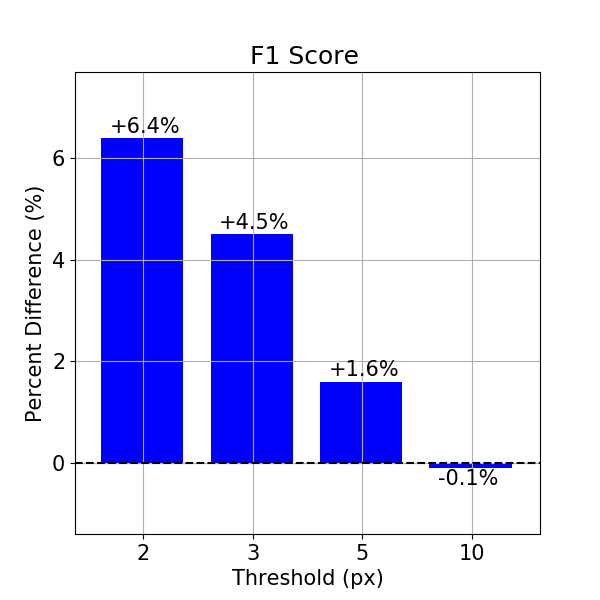} \hskip -15pt
\caption{In this figure we show the percent difference between our models trained to predict the direction map versus predicting the dilated normals. We report the difference for precision, recall and F1 score. We see using the direction map significantly improves the result at all thresholds of our metric. Furthermore, the direction map method improves the connectivity by 1\%.}
\label{fig:direction_methods}
\end{figure*}

\subsection{Convolutional Snake (cSnake)}
\label{sec:csnake}
In this section, we describe the mechanics of our  module that captures the precise topology of a road boundary. In the following we refer to this module as {\it cSnake}. Note that our module is fully automatic and does not require any user in the loop. 
At a high level, drawing on the deep detection and directional features obtained from the input image and the location of the endpoints, the cSnake iteratively attends to rotated regions of interest in the image and outputs the vertices of a polyline corresponding to a road boundary. The direction of travel of the cSnake is obtained from the direction map $D$.

In particular, we first compute the local maxima of the endpoints heatmap $E$ to find the initial vertices of the road boundaries.Then for each endpoint we draw a separate polyline as follows: 
Given an initial vertex $x_0$ of the endpoint and a direction vector $v_0$, we use the Spatial Transformer Network \cite{jaderberg2015spatial} to crop a rotated ROI from the concatenation of the detection and direction maps $S$ and $D$. Intuitively, the detection distance map $S$ and the direction map $D$ encourage the cSnake module to pull and place a vertex on the road boundaries.  
The direction $v_0$ runs in parallel to the road boundary at position $x_0$ and is obtained by first looking up the vector from the closest pixel in the direction map $D$ and then rotating it by 90 degrees pointing away from the image boundary. This rotated ROI is fed to a CNN that outputs an argmax of the next vertex $x_{1}$ in the image. The next direction vector $v_1$ is obtained similarly by looking up the direction map $D$ and rotating it by 90 degrees to be in the same direction of $v_0$. We repeat this process until the end of the road boundary where we fall outside of the image. At the end we obtain a polyline prediction $x = (x_i)$ with vertices in $\R^2$ that captures the global topology of the road boundary.

Thus for each input $I$, we obtain a set of polylines emanating from the predicted endpoints. Note that we can assign a score to each polyline by taking the average of the detection scores on its vertex pixels.
We use this score for two minimal post-processing steps: i) We remove the low scoring polylines protruding from  potentially false negative endpoints. ii) Two predicted endpoints could correspond to the same road boundary giving rise to two different polylines. Thus, we look at all pairs of polylines and if they overlap by more than $30\%$ in a dilated region around them, we only keep the highest scoring one.

\paragraph{Network Architecture:} For each cropped ROI, we feed it through a CNN. We use the same encoder decoder backbone that we used to predict the deep features but with one less convolutional layer in both the encoder and decoder blocks. The output is a score map that we can take argmax of to obtain the next vertex for cropping.

\subsection{Learning}

\paragraph{Deep visual features:} To learn the deep visual features, we use a multi-task objective, where the regression loss is used for the distance transform feature maps $S$ and $E$ and  the cosine similarity loss  is used for  the direction map $D$. Thus:
\begin{align}
	\ell(S,E, D) = \ell_{det}(S) + \lambda_1 \ell_{end}(E) + \lambda_2 \ell_{dir}(D)
\end{align}
In our experiments we set the loss weighting parameters $\lambda_1$ and $\lambda_2$ to be 10.

\paragraph{Convolutional Snake:} Similar to \cite{Homayounfar_2018_CVPR}, in order to match the edges of a predicted polyline $P$ to its corresponding ground truth road boundary $Q$, we use the Chamfer Distance defined as:
\begin{align}
	L(P, Q) &= \sum_{i}\min_{q \in Q}{ \norm{p_i - q}_2} + \sum_{j} \min_{p \in P}{ \norm{p - q_j}_2} \label{eq:poly_loss}
\end{align}
where $p$ and $q$ are the rasterized edge pixels of the polylines $P$ and $Q$ respectively. This loss function encourages the edges of the predicted polyline to fall completely on its ground truth and vice versa. This is a more suitable loss function for matching two polylines rather than one penalizing the position of vertices. For example a ground truth straight line segment can be redundantly represented by three vertices rather than two and thus misleading the neural network when using a vertex based loss.

\section{Experimental Evaluation}

\subsection{Experimental Details} 

\paragraph{Dataset:} Our dataset consists of BEV projected LiDAR point clouds and camera imagery of intersections and other regions of the road from multiple passes of a self-driving vehicle in a major north American city. In total, 4750km were driven to collect this data from a 50 km$^2$ area with a total of approximately 540 billion LiDAR points. We then tile and split the dataset into 2500, 1000, 1250 train/val/test BEV images that are separated based on 2 longitudinal lines dividing the city. On average, our images are 1927px ($\pm 893$) x 2162px ($\pm 712$) with 4cm/px resolution. 

In more detail, to create our dataset,  we first drive around the areas we want to map several times. We exploit an efficient deep-learning based  pointcloud segmentation algorithm \cite{zhang2018efficient} to remove moving objects and register the pointclouds using ICP. We then generate a very high resolution bird's eye view image given the pointcloud (i.e., 4cm/pixel). This process allows us to have views of the world without occlusion. We also generate panoramas using the camera rig in a similar fashion. To compute the elevation gradient image, we simply apply a sobel filter on the LiDAR elevation for both x and y directions and then take its magnitude.

\begin{figure}[t!]
\centering
\includegraphics[width=.8\linewidth]{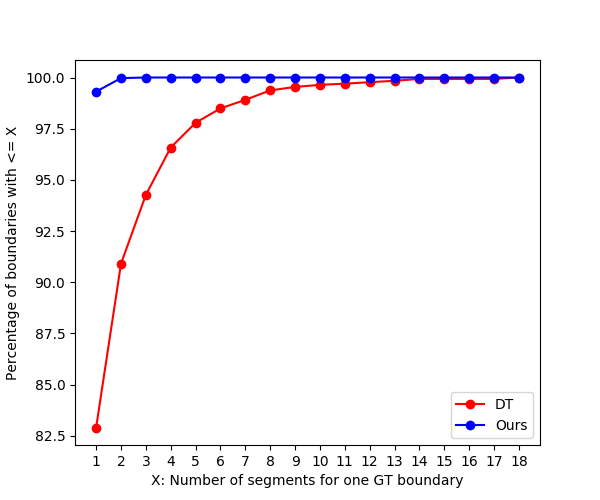} 
 
\caption{In this figure we show the cumulative percentage of GT boundaries with X number of predicted segments. For our model, we see that 99.3\% of the GT boundaries have a single predicted boundary.}
\label{fig:cumul_connectivity}
\end{figure}

\vspace{-5mm}

\paragraph{Baseline:} Since this is a new task, we cannot compare to prior work.  To create a competitive baseline,  we first binarize the distance transform output of our method at a threshold and then skeletonize. For a fair comparison, we use grid search to find the threshold that gives us the best results. Next, we find the connected components and consider each as a predicted polyline.

\vspace{-5mm}

\paragraph{Implementation Details:}
We trained our deep feature prediction model distributed over 16 Titan 1080 Ti GPUs each with a batch size of 1 using ADAM \cite{ADAM} with a learning rate of 1e-4 and a weight decay of 5e-4. %
We perform data augmentation by randomly flipping and rotating the images during training. The model is trained for 250 epochs over the entire dataset and takes 12 hours.
The cSnake is also trained distributed on 16 Titan 1080 Ti GPUs each with a batch size of 1, ADAM, a learning rate of 5e-4 and a weight decay of 1e-4. The model is trained for 60 epochs over the entire dataset. During training, we give the network the ground truth end points and add $\pm$16 pixels of noise. We also use amortized learning and train with ground truth direction and distance transform features 50\% of the time. During training, we  give the network the number of steps based on the length of the ground truth boundary plus 5 extra steps. %

\subsection{Metrics}

\begin{figure}[t]
	\[\arraycolsep=1.0pt
	\begin{array}{c}

	\includegraphics[width=\linewidth]{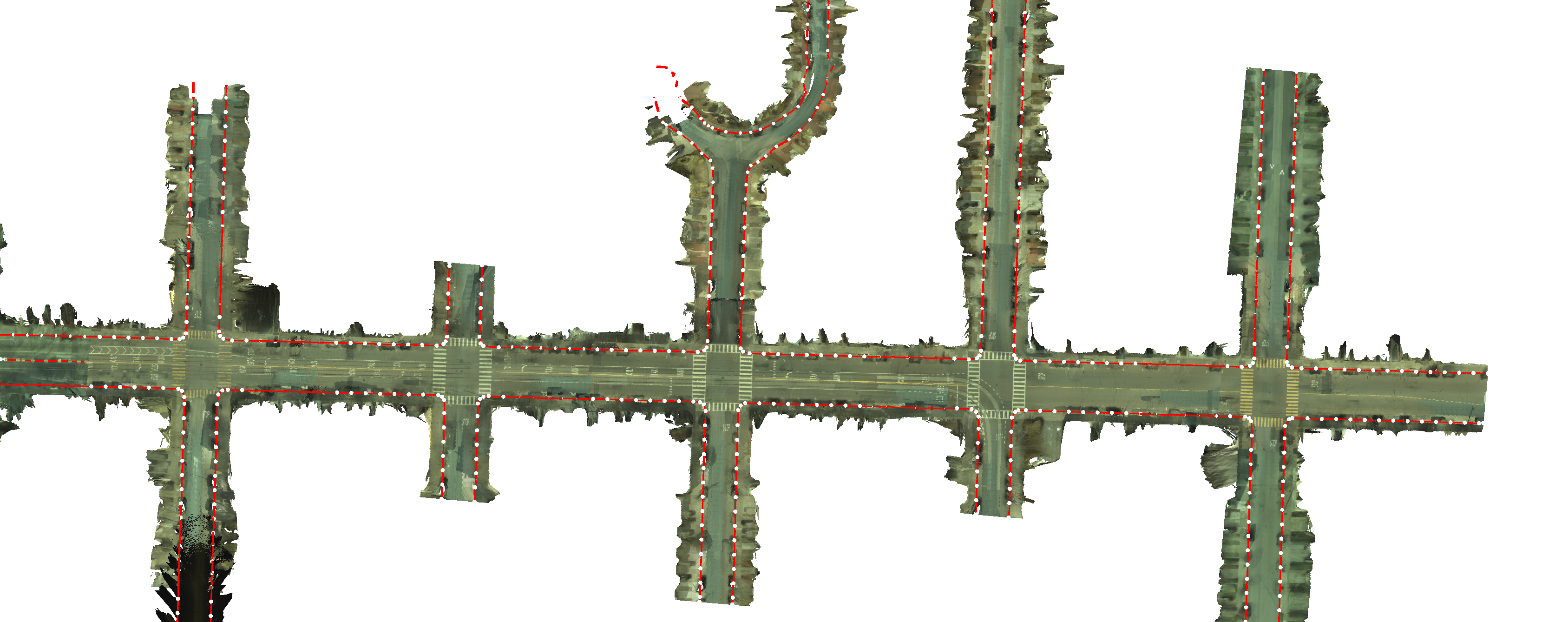}  \\ 
	
	\end{array}
	\]
	\caption{Example of our outputs stitched together (please zoom).}
	\label{fig:stitched}
\end{figure}

\begin{figure*}[t]
	\vspace{-0.5cm}
	\centering
	\setlength{\tabcolsep}{1pt}
	\begin{tabular}{cccccc}

		\raisebox{30px}{\rotatebox{90}{GT}}
		\includegraphics[width=0.19\linewidth]{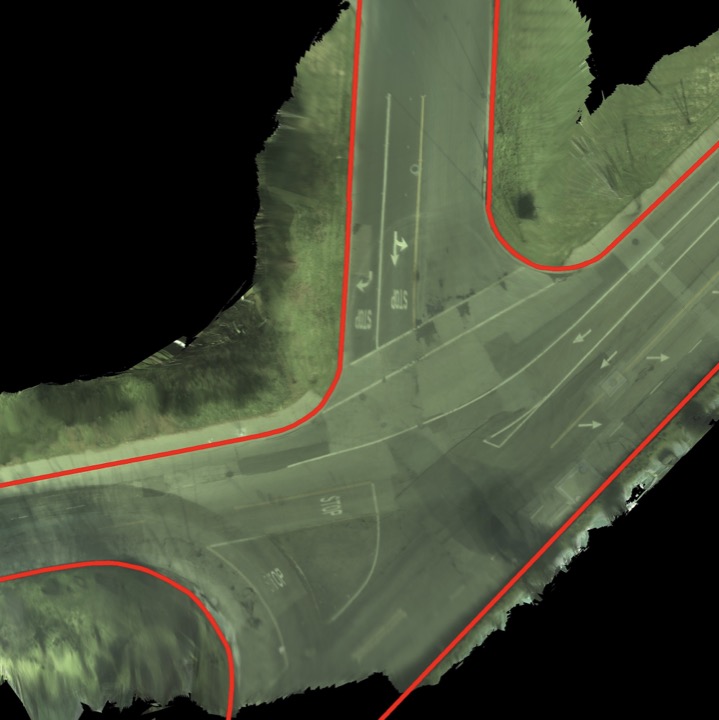}  & 
		\includegraphics[width=0.19\linewidth]{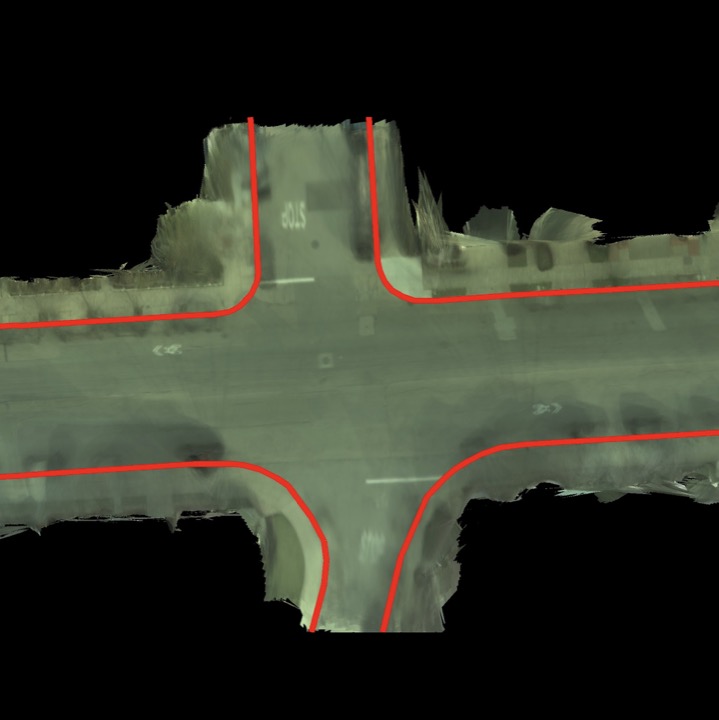}  & 
		\includegraphics[width=0.19\linewidth]{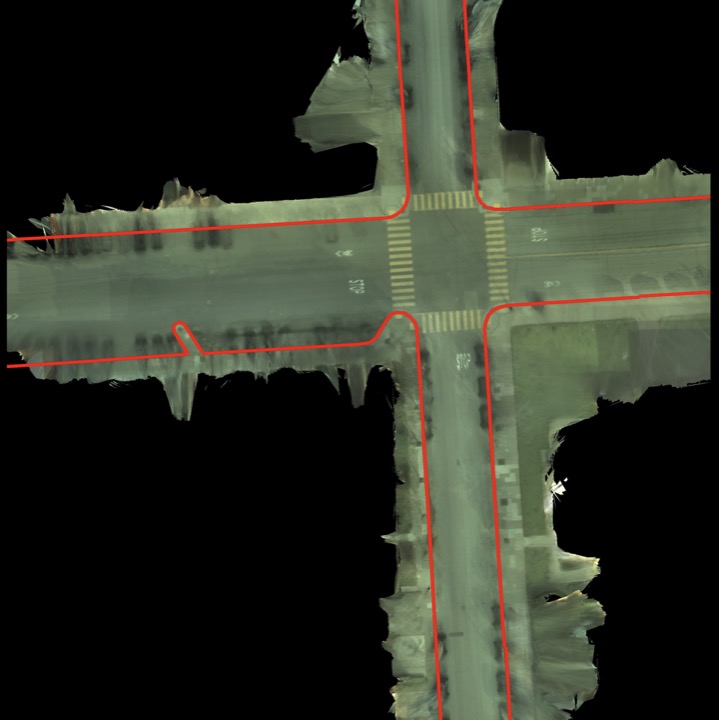}  & 
		\includegraphics[width=0.19\linewidth]{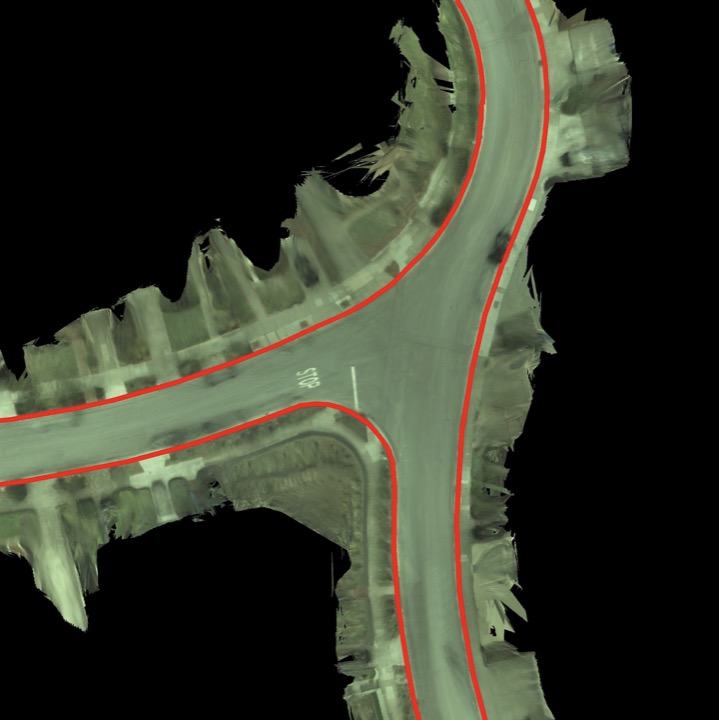}  & 
		\includegraphics[width=0.19\linewidth]{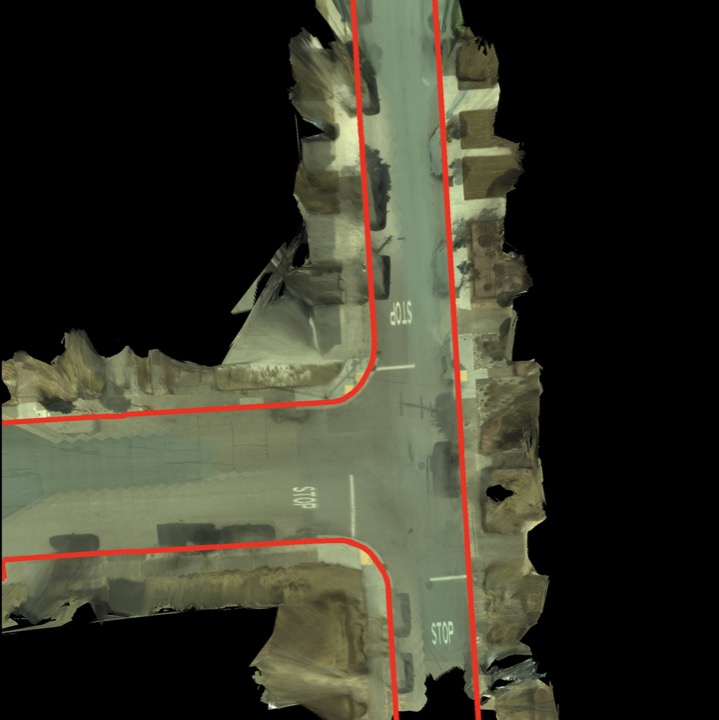}  \\
		
		\raisebox{20px}{\rotatebox{90}{Predictions}}
		\includegraphics[width=0.19\linewidth]{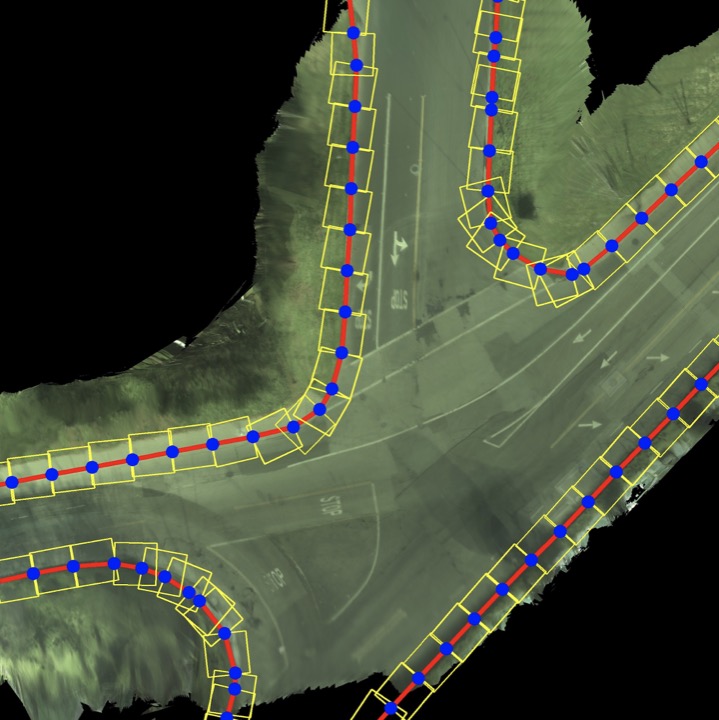}  & 
		\includegraphics[width=0.19\linewidth]{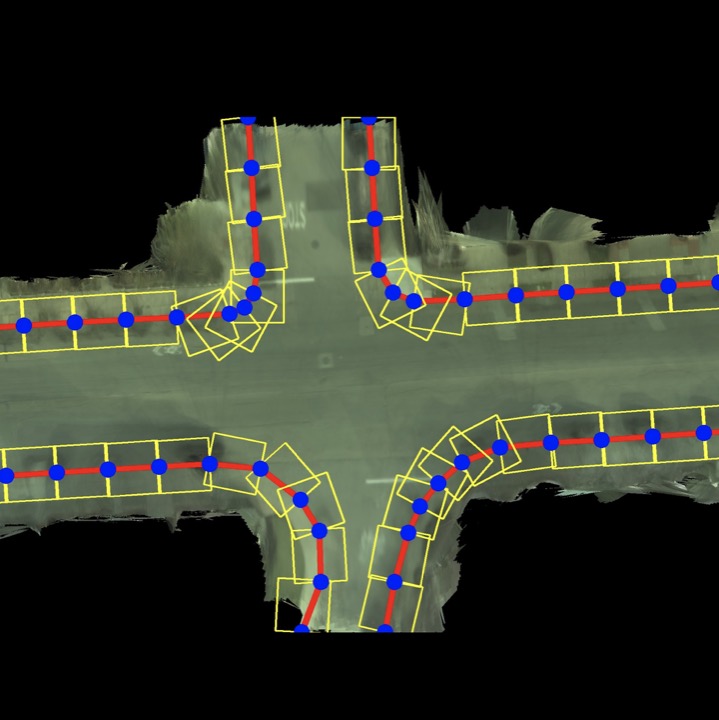}  & 
		\includegraphics[width=0.19\linewidth]{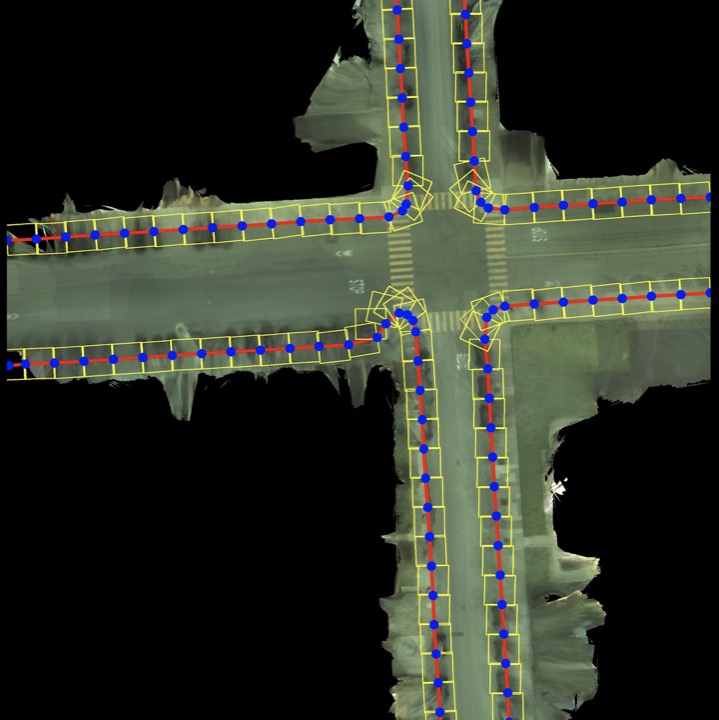}  & 
		\includegraphics[width=0.19\linewidth]{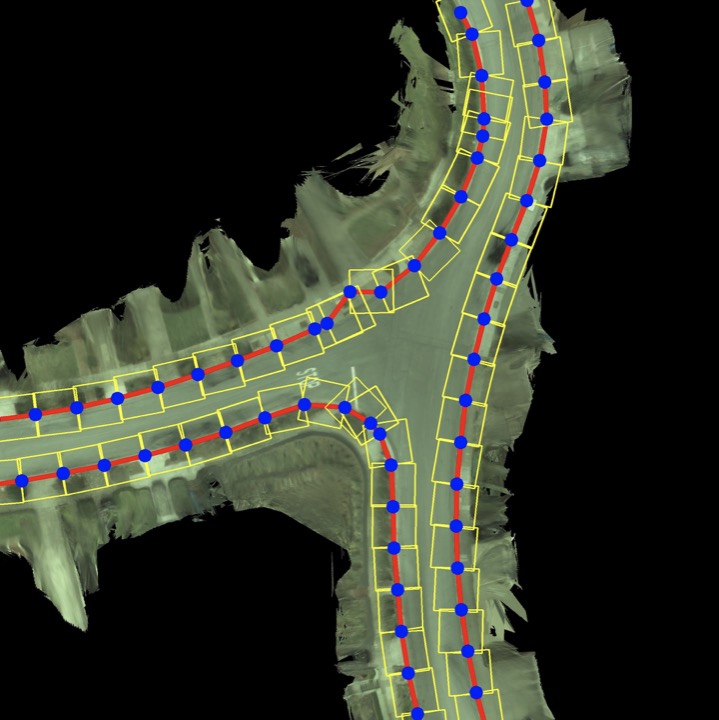}  & 
		\includegraphics[width=0.19\linewidth]{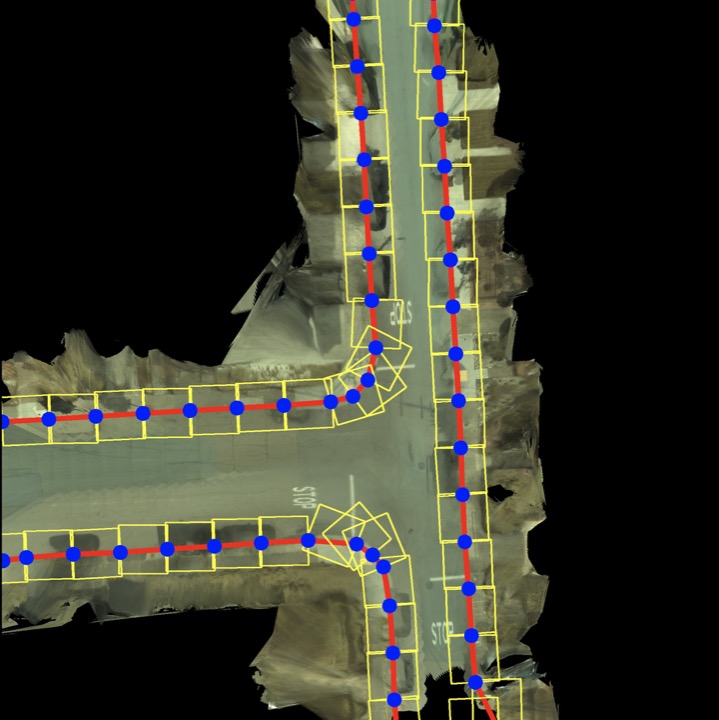}  \\
		
		\raisebox{30px}{\rotatebox{90}{GT}}
		\includegraphics[width=0.19\linewidth]{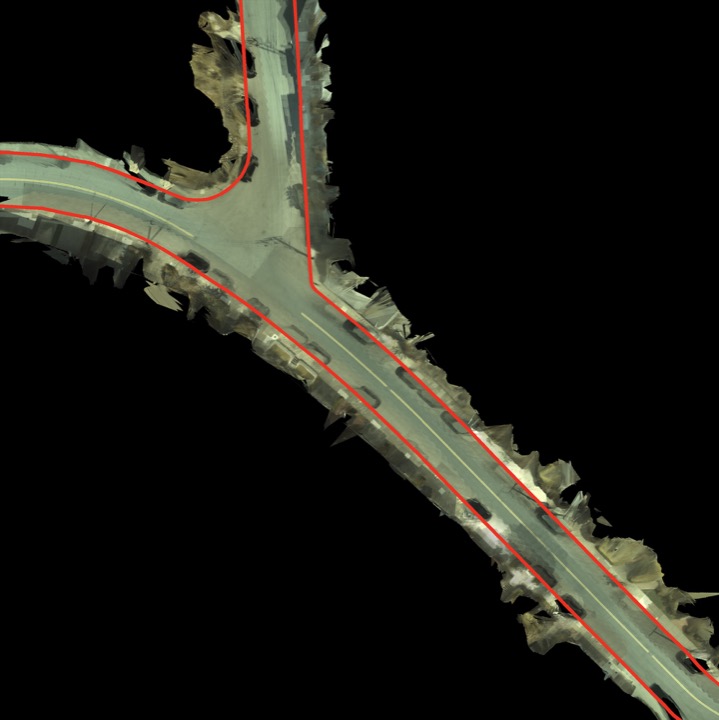}  & 
		\includegraphics[width=0.19\linewidth]{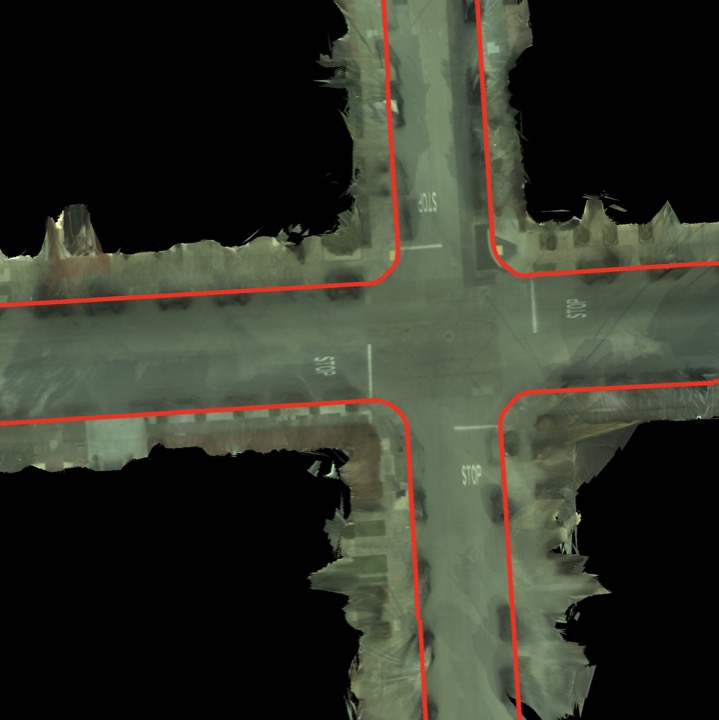}  & 
		\includegraphics[width=0.19\linewidth]{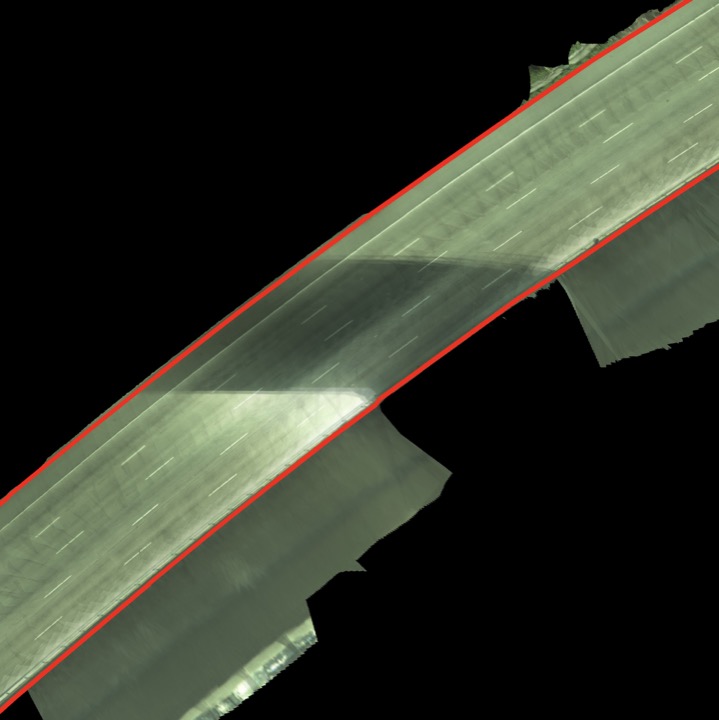}  & 
		\includegraphics[width=0.19\linewidth]{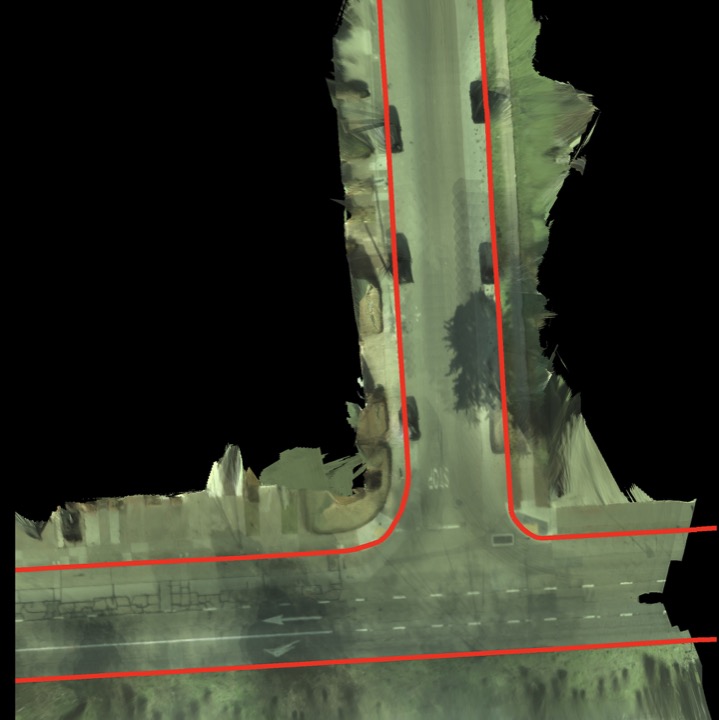}  & 
		\includegraphics[width=0.19\linewidth]{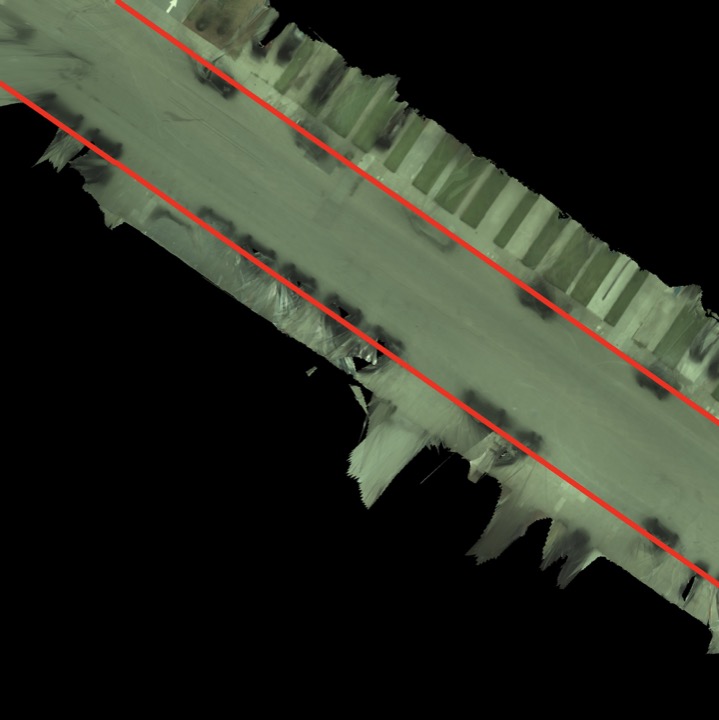}  \\
		
		\raisebox{20px}{\rotatebox{90}{Predictions}}
		\includegraphics[width=0.19\linewidth]{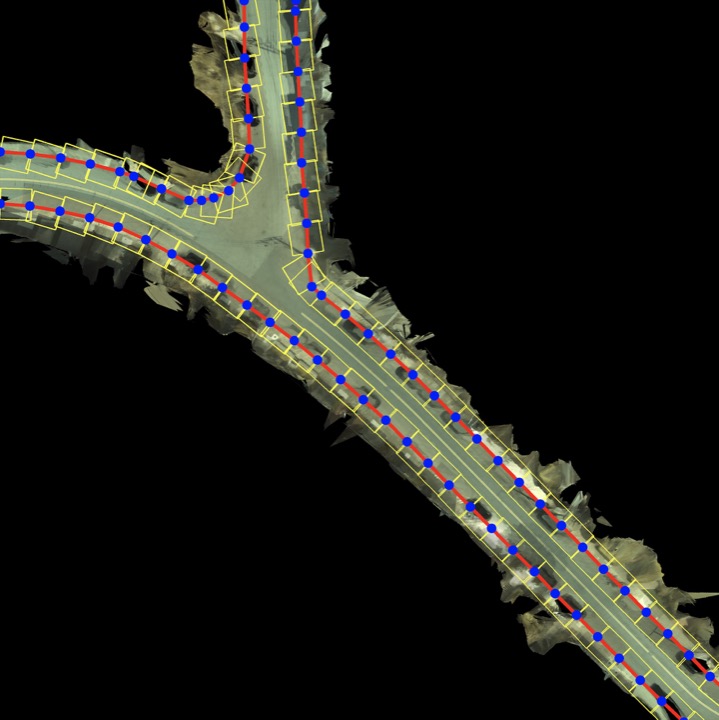}  & 
		\includegraphics[width=0.19\linewidth]{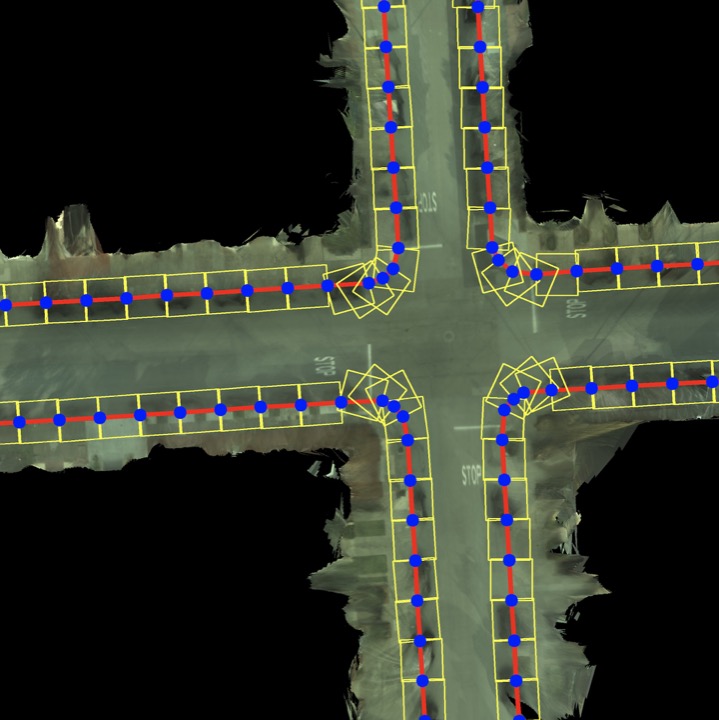}  & 
		\includegraphics[width=0.19\linewidth]{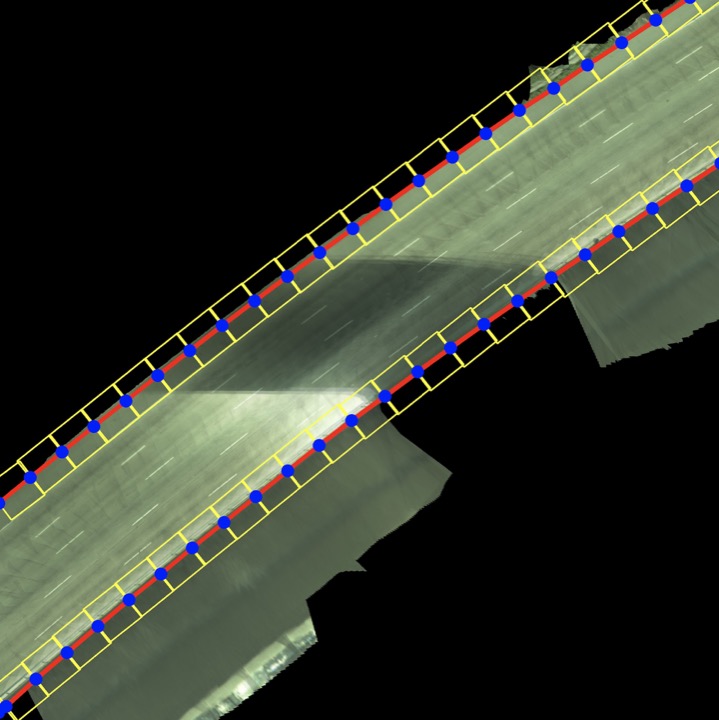}  & 
		\includegraphics[width=0.19\linewidth]{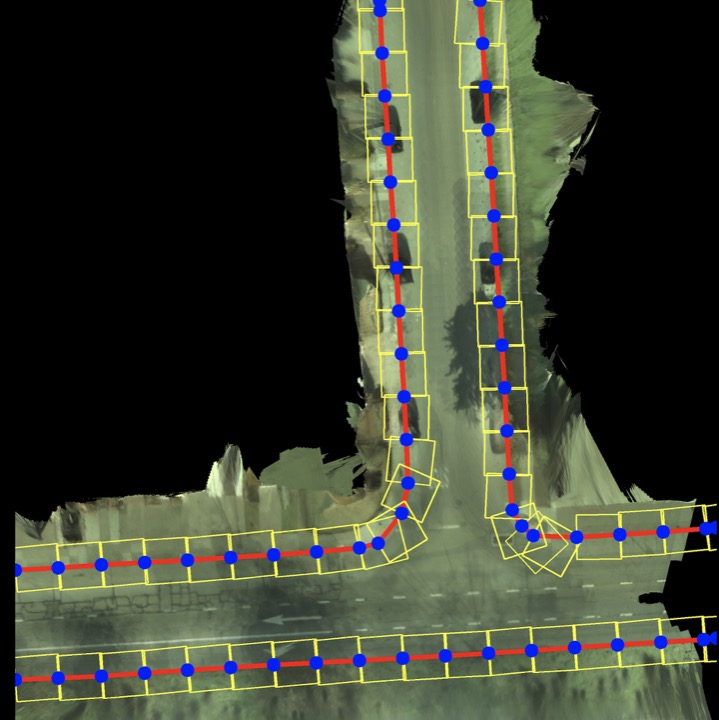}  & 
		\includegraphics[width=0.19\linewidth]{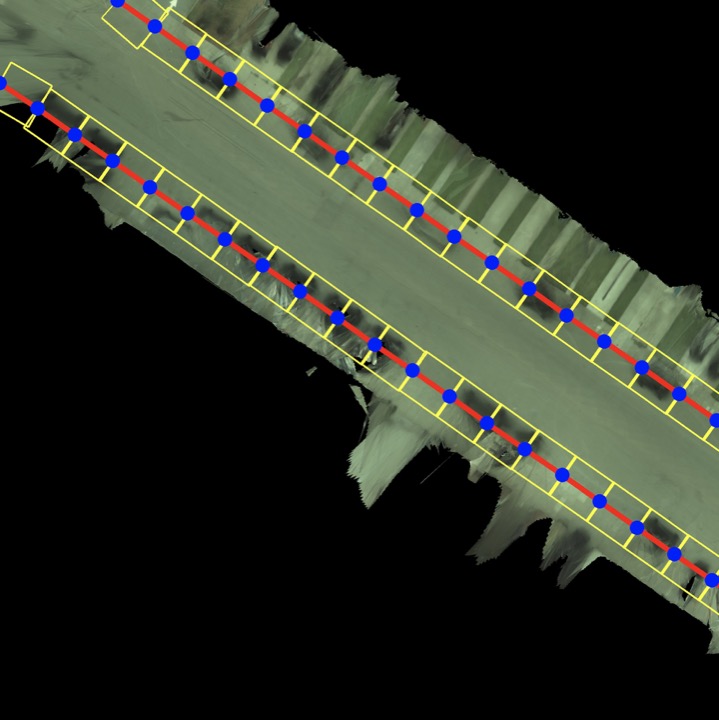}  \\

	\end{tabular}
	\caption{Qualitative results: (\textbf{Rows 1,3}) GT road boundaries displayed in red and overlaid on camera imagery. (\textbf{ Rows 2,4}) Road boundary prediction polylines. The blue dots correspond to the vertex outputs of the cSnake module. The yellow boxes are the rotated ROIs that the cSnake attends to while drawing the road boundary polylines. Note that we crop our imagery for better visualization. Please refer to the supplementary material for more complete visualization.}
	\label{fig:qual-results}
	\vspace{-2mm}
\end{figure*}

Given a set of polyline predictions, we assign each one to a ground truth road boundary that has the smallest Hausdorff distance. Note that multiple predictions could be assigned to the same ground truth boundary but only one ground truth road boundary can be assigned to a prediction polyline.
We now  specify our metrics.  

\vspace{-5mm}

\paragraph{Precision and Recall:} For precision and recall, we use the definition of \cite{TCity2017} and specify points on the predicted polylines as either true positive or false positive if they are within a threshold of the ground truth polyline. False negatives are the points on the ground truth polyline that fall outside that threshold. We also combine the two metrics and report its result for harmonic mean (f1 score).

\paragraph{Connectivity:} For each ground truth boundary, let $M$ be the number of its assigned predicted polylines. We define:
\begin{align}
Connectivity = \frac{1(M>0)}{M}
\end{align}

This metric penalizes the assignment of multiple small predicted segments to a ground truth road boundary.

\paragraph{Aggregate Metrics:} We report the mean taken across the ground truth road boundaries at different thresholds.

\subsection{ Results}
\paragraph{Baseline: } As shown in  Table \ref{tab:baseline}, our method significantly outperforms the baseline in almost all metrics. The baseline is better at in precision at a threshold of 10px, however, this is because the baseline has lots of small connected components. %
However, in practice, these would be thrown out when doing actual annotation of road boundaries as they are too small to be useful.

\vspace{-5mm}

\paragraph{Sensor Modality: } We explore various sensor input combinations for our model. In Table \ref{tab:compare_inputs}, we show in line (4) that using every sensor input from L (lidar intensity), E (lidar elevation gradient) and C (camera) produces the best result. We perform an ablation studies by removing the camera input and then the elevation input, and also an experiment with camera only and show a significant performance drop.

\begin{figure*}[t]
	\vspace{-0.5cm}
	\[\arraycolsep=1.0pt
	\begin{array}{cccccc}

	\raisebox{2px}{{Lidar}} &	
	\raisebox{2px}{{Camera}} &
	\raisebox{2px}{{Elevation Gradient}} &
	\raisebox{2px}{{Detection Map}} &
	\raisebox{2px}{{Endpoints}} &
	\raisebox{2px}{{Direction Map}} \\	
	
	\includegraphics[width=0.16\linewidth]{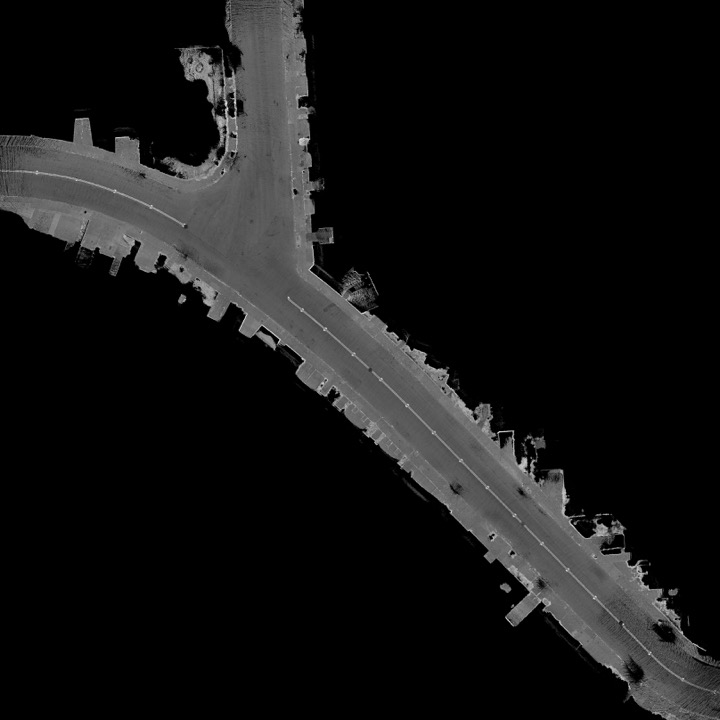}  & 
	\includegraphics[width=0.16\linewidth]{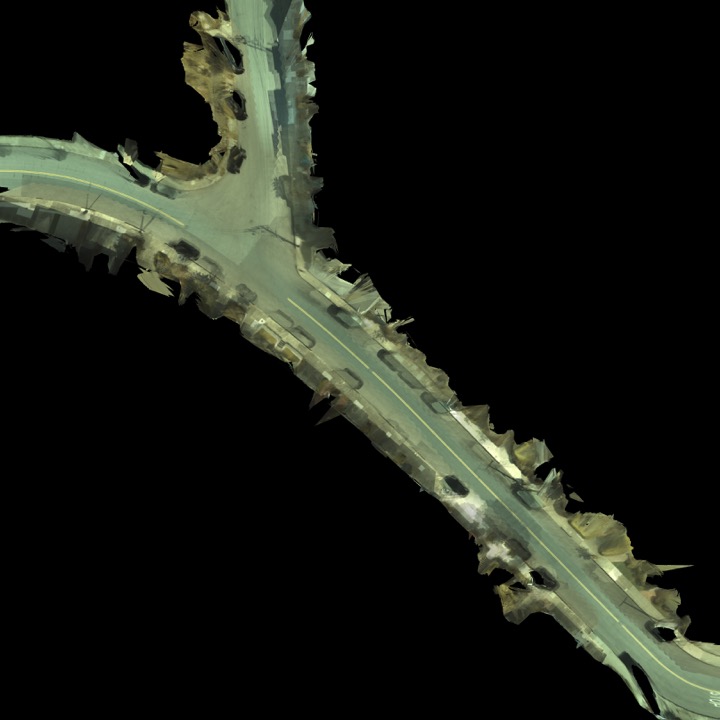}&  
	\includegraphics[width=0.16\linewidth]{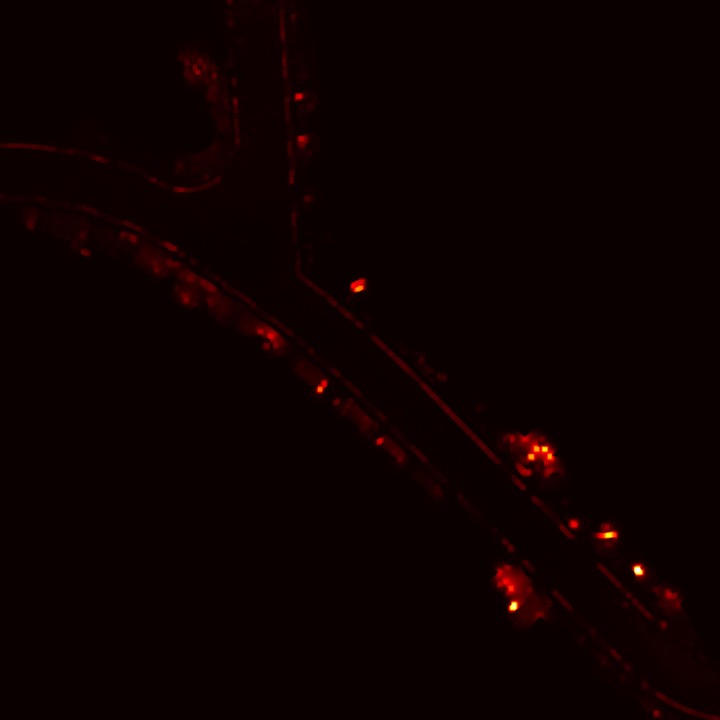} &
	\includegraphics[width=0.16\linewidth]{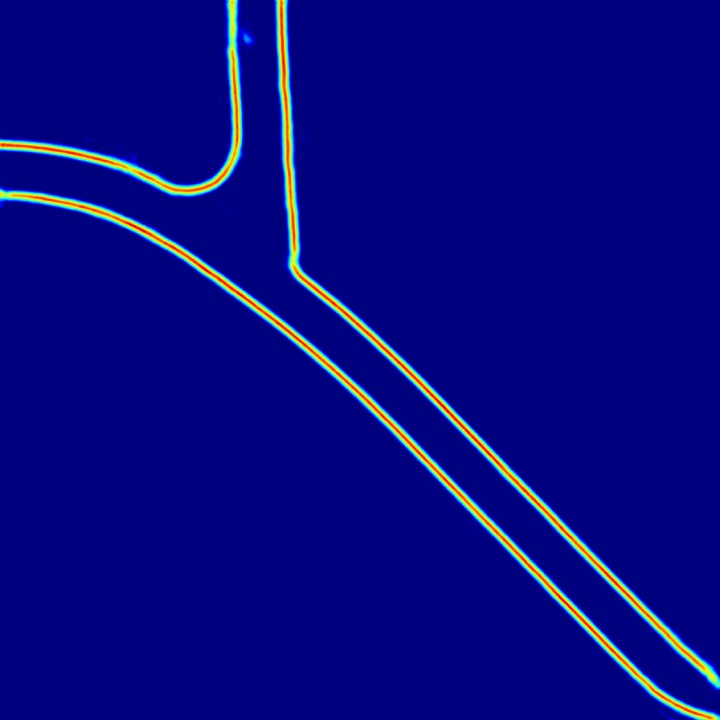}  & 
	\includegraphics[width=0.16\linewidth]{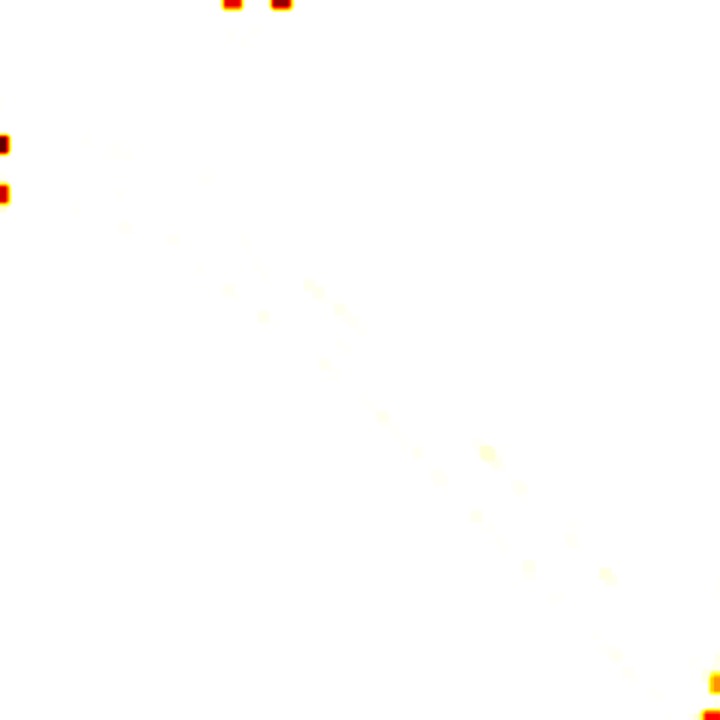}& 
	\includegraphics[width=0.16\linewidth]{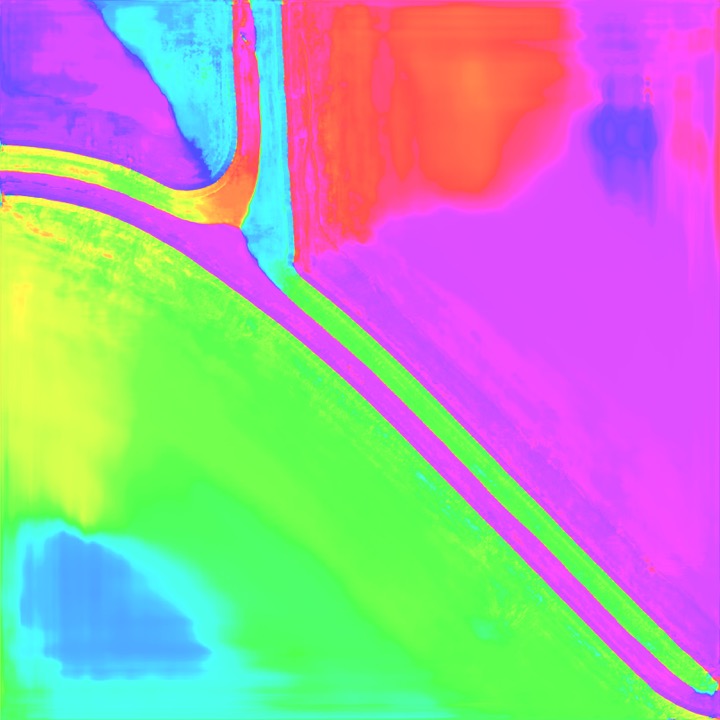} \\

	\includegraphics[width=0.16\linewidth]{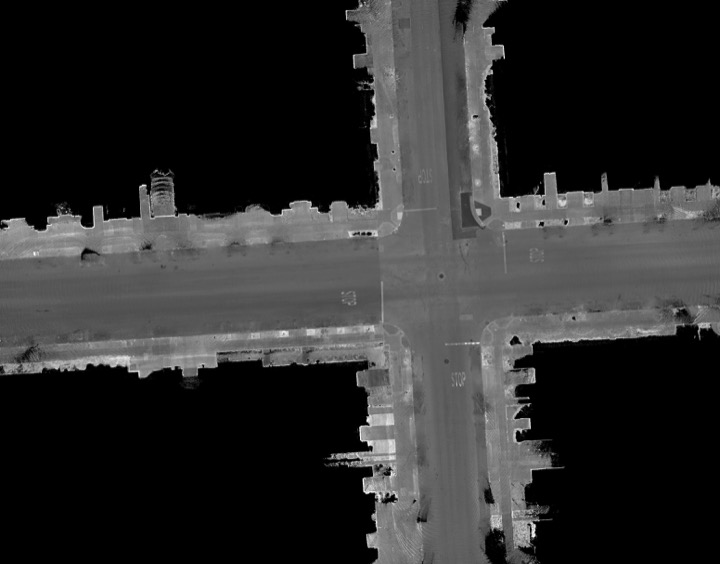}  & 
	\includegraphics[width=0.16\linewidth]{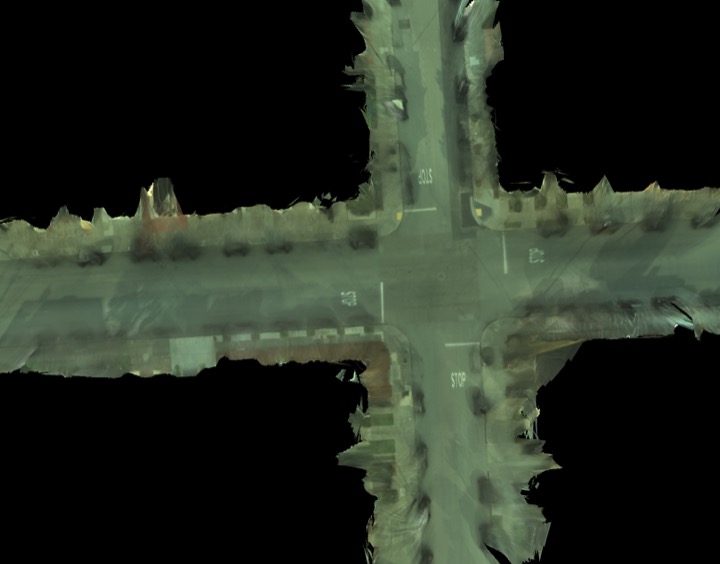}&  
	\includegraphics[width=0.16\linewidth]{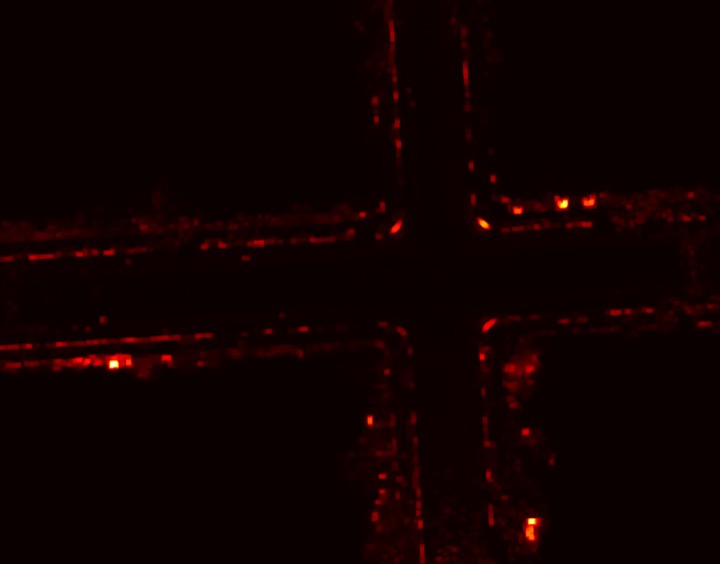} &
	\includegraphics[width=0.16\linewidth]{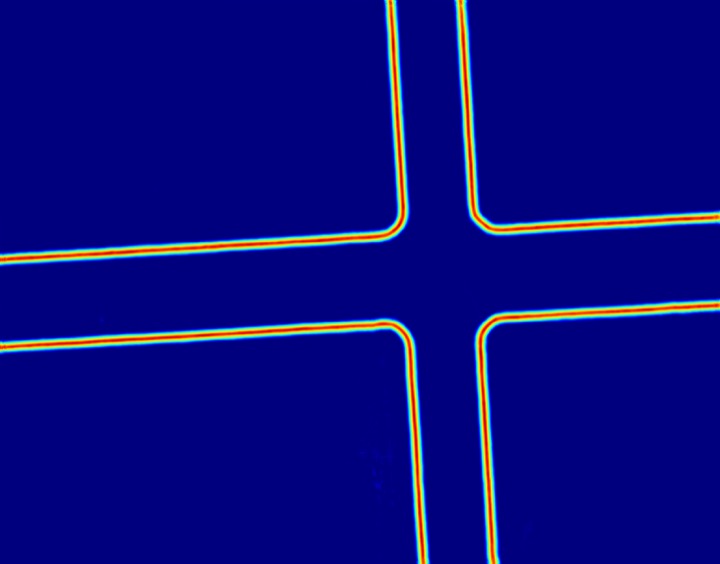}  & 
	\includegraphics[width=0.16\linewidth]{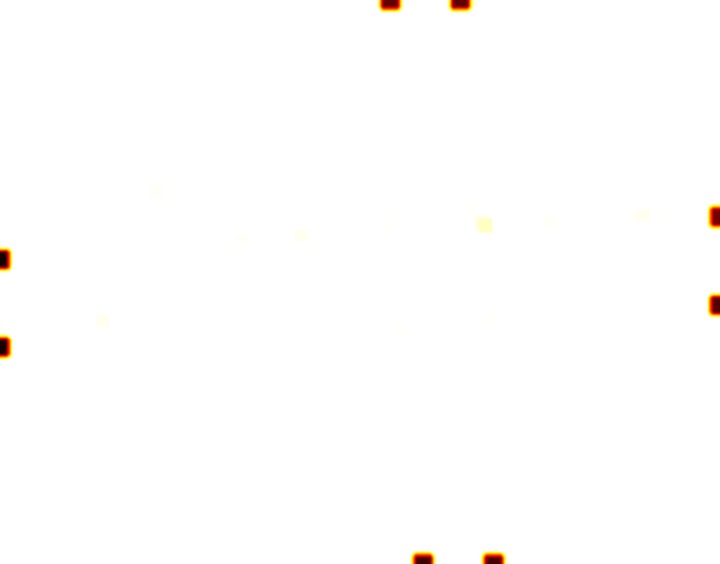}& 
	\includegraphics[width=0.16\linewidth]{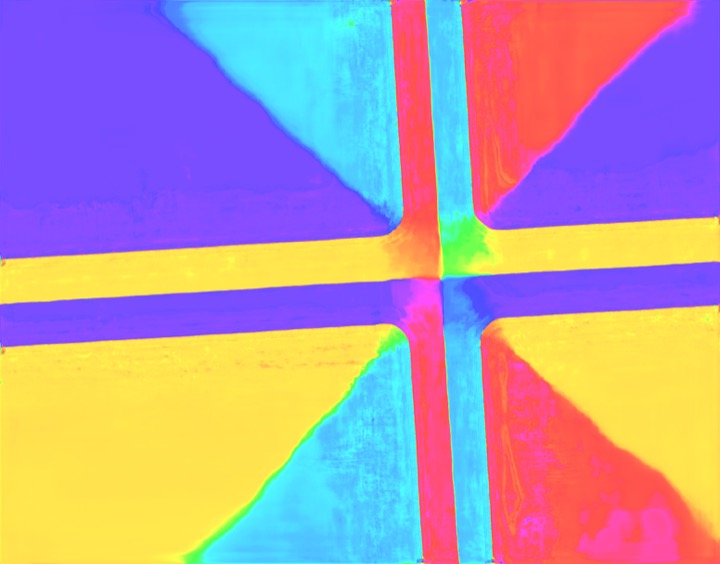} \\

	\includegraphics[width=0.16\linewidth]{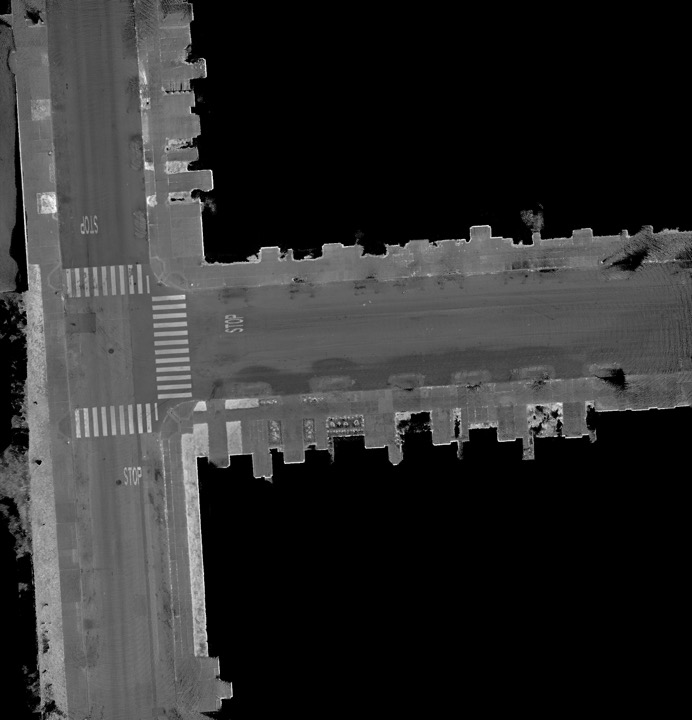}  & 
	\includegraphics[width=0.16\linewidth]{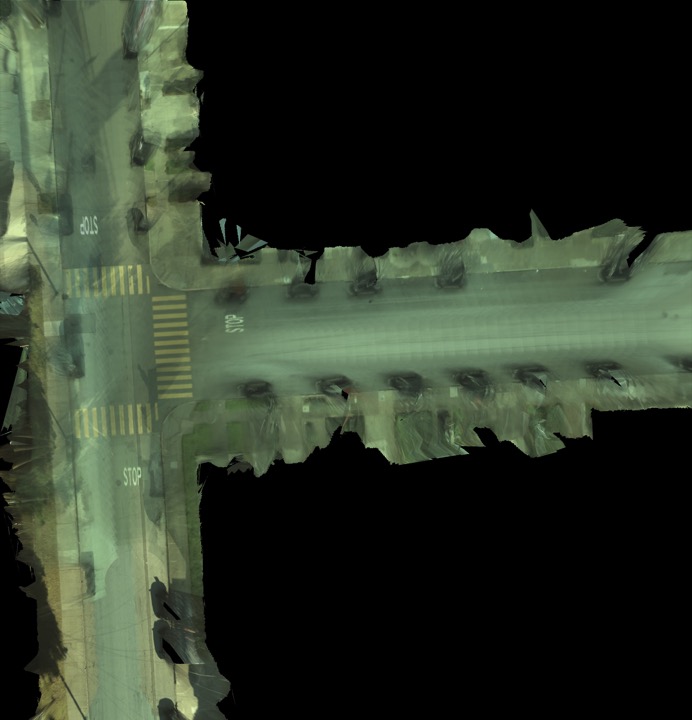}&  
	\includegraphics[width=0.16\linewidth]{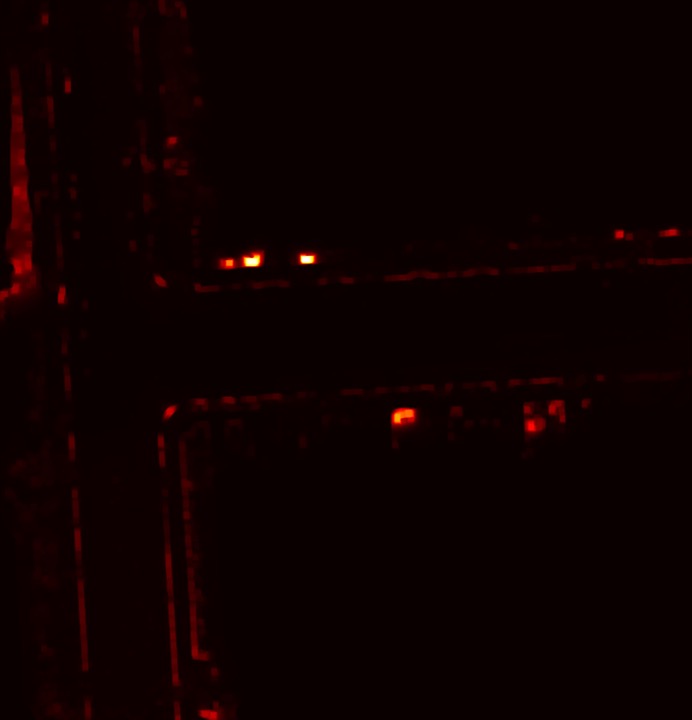} &
	\includegraphics[width=0.16\linewidth]{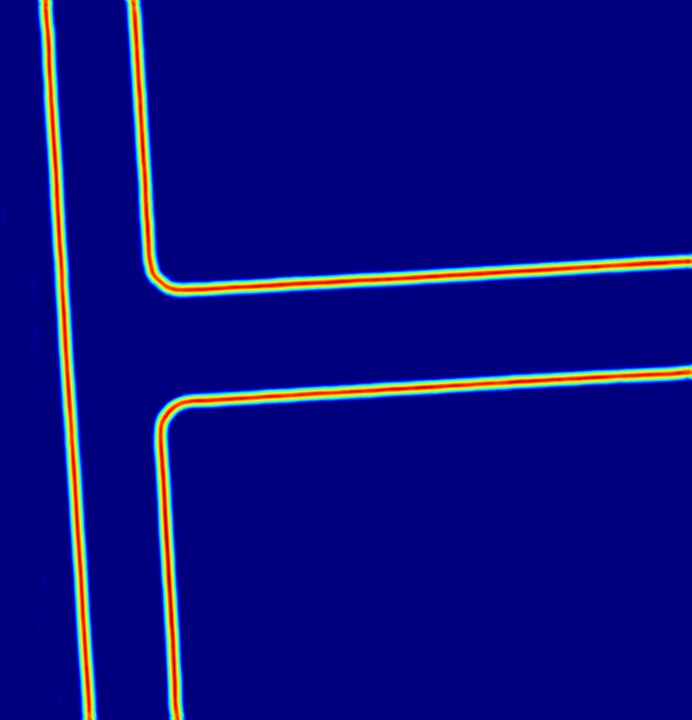}  & 
	\includegraphics[width=0.16\linewidth]{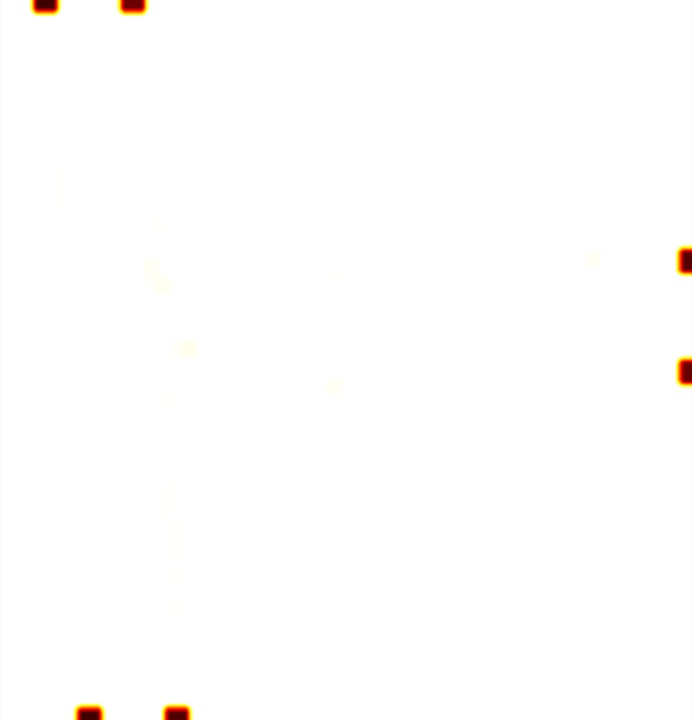}& 
	\includegraphics[width=0.16\linewidth]{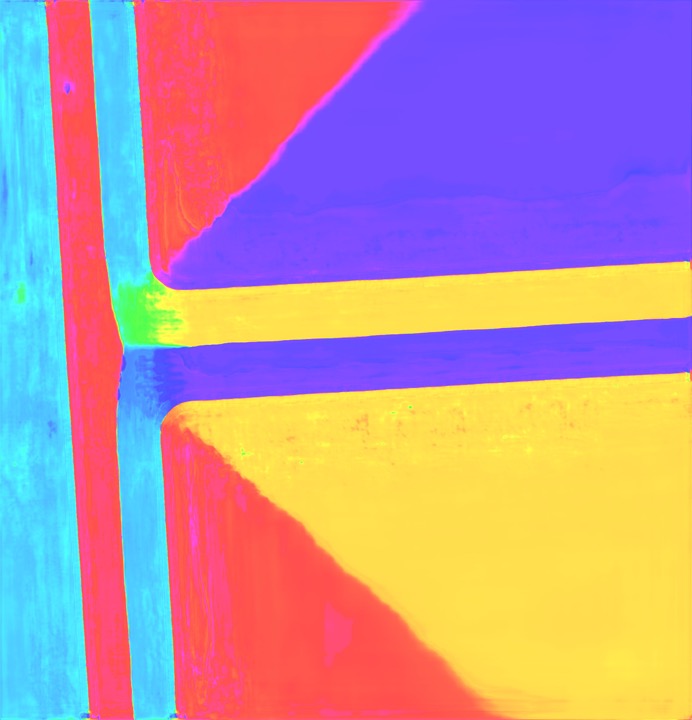} \\
	
	\includegraphics[width=0.16\linewidth]{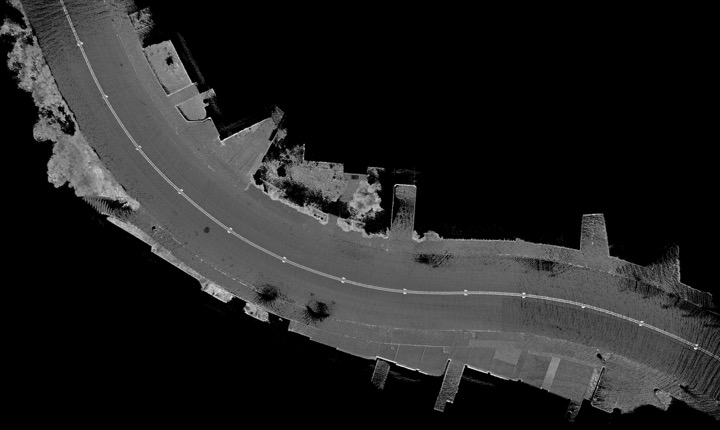}  & 
	\includegraphics[width=0.16\linewidth]{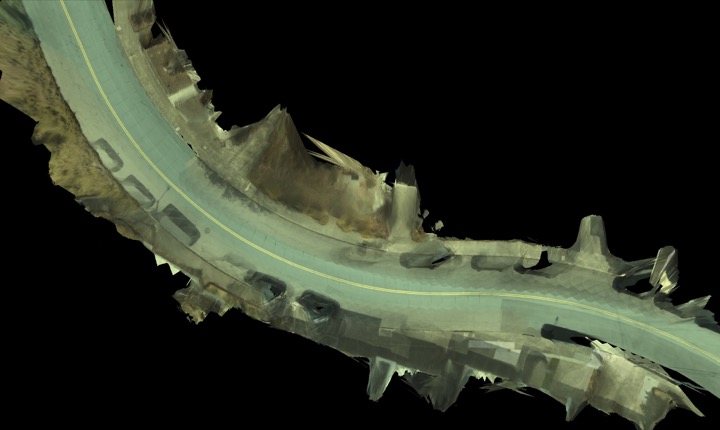}&  
	\includegraphics[width=0.16\linewidth]{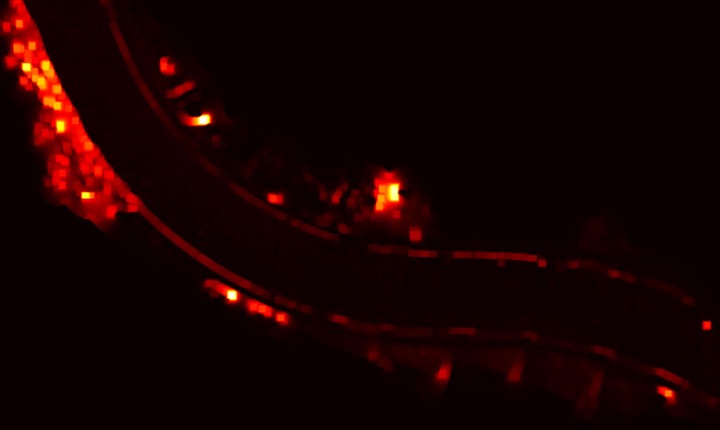} &
	\includegraphics[width=0.16\linewidth]{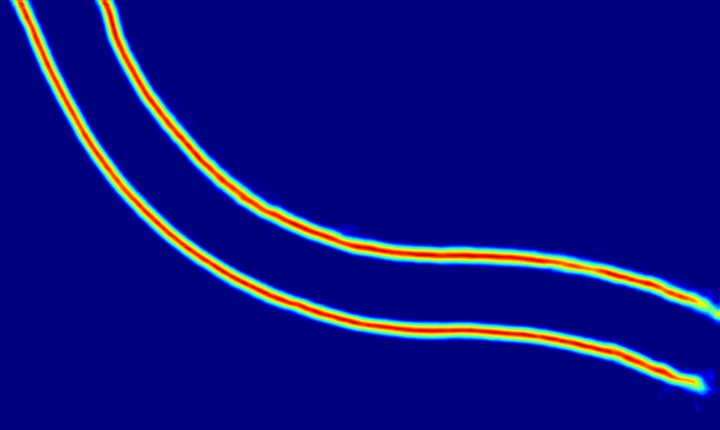}  & 
	\includegraphics[width=0.16\linewidth]{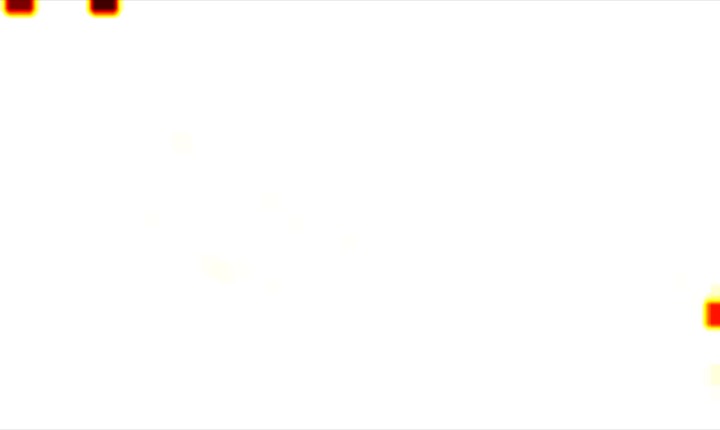}& 
	\includegraphics[width=0.16\linewidth]{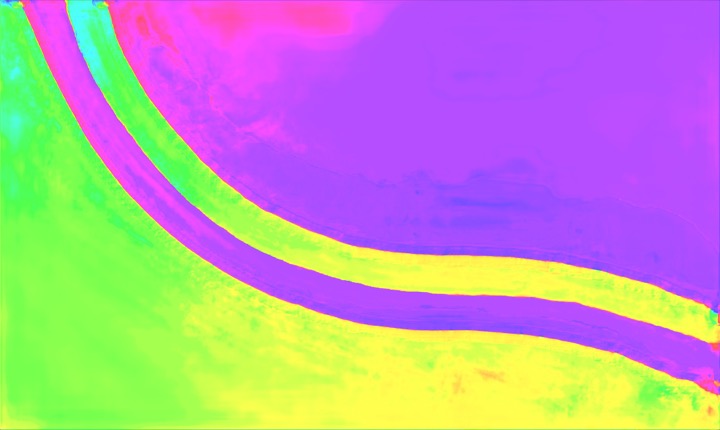} \\

	\end{array}
	\]
	
	\caption{Deep Features: Columns \textbf{(1-3)} correspond to the  inputs and columns \textbf{(4-6)} correspond to the deep feature maps. The direction map shown here as a flow field \cite{color_code}.}
	\label{fig:feat}
	\vspace{-4mm}
\end{figure*}

\begin{figure}[t]
\vspace{-0.5cm}
	\[\arraycolsep=1.0pt
	\begin{array}{cc}

	\raisebox{2px}{{Predictions}} &	
	\raisebox{2px}{{GT}} \\	
	
	\includegraphics[width=0.49\linewidth]{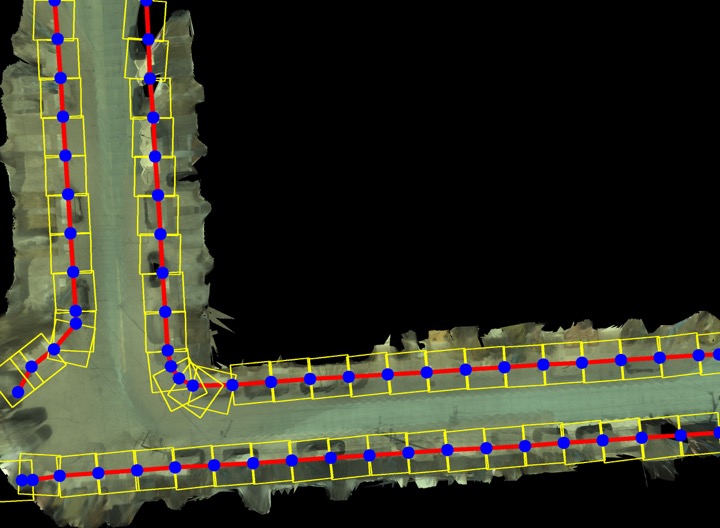}  & 
	\includegraphics[width=0.49\linewidth]{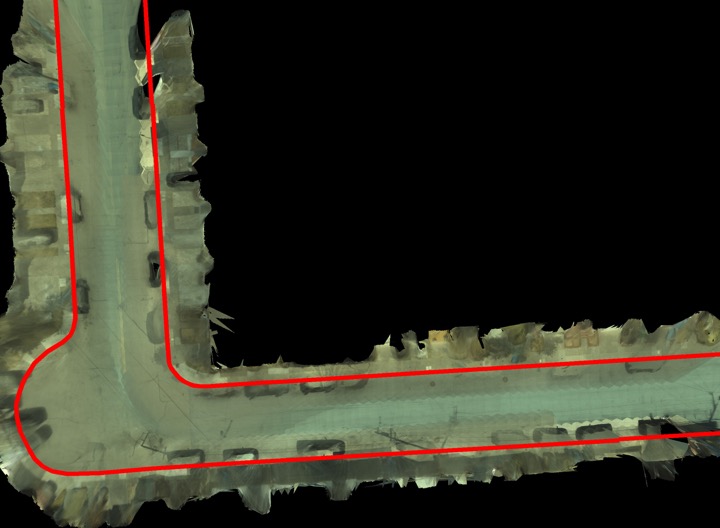}  \\

	\includegraphics[width=0.49\linewidth]{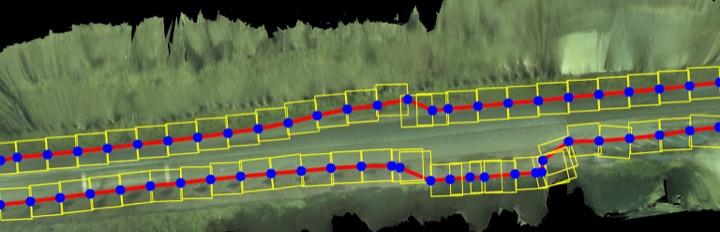}  & 
	\includegraphics[width=0.49\linewidth]{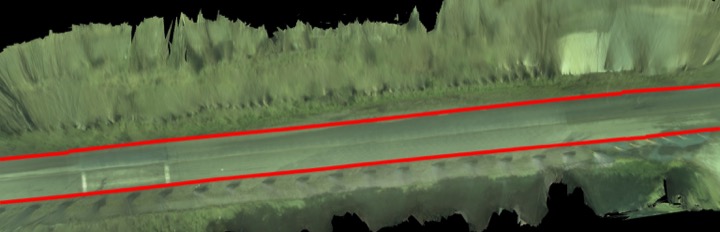}  \\

	\includegraphics[angle=90,origin=c,width=0.49\linewidth]{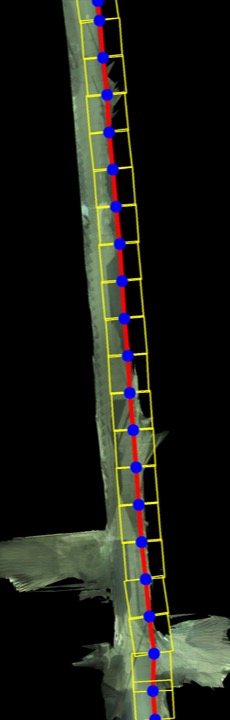}  & 
	\includegraphics[angle=90,origin=c,width=0.49\linewidth]{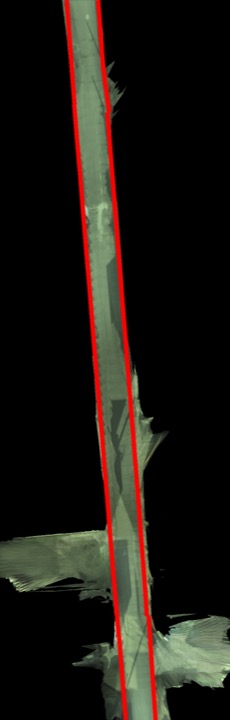}  \\

	\end{array}
	\]
	\vspace{-18mm}
	\caption{Failure Modes: \textbf{(Left)} predictions \textbf{(right)} GT.}
	\label{fig:fail}
\end{figure}

\vspace{-5mm}

\paragraph{Amortized Learning: } The convolutional snake can be trained using either the ground truth or predicted deep features. For our best model, we train using half ground truth and half predicted deep features. We also train a model using only predicted deep features and show the difference in results of the two models in Figure \ref{fig:amortized_learning}. This figure shows the percentage difference between our model trained with amortized learning and the model trained with only predicted features. At a threshold of 2px (8cm), the model trained with amortized learning is 5\% better across all our metrics. In terms of connectivity, both methods of training achieve around the same result.

\vspace{-5mm}

\paragraph{Exploring Direction Prediction Alternatives: } We explore another method to predict the direction feature used by the convolutional snake. Here, we predict a pixel-wise direction at the boundary of the road that is the normal pointing into the road. Since this notion only exists at the road boundaries, we dilate the road boundary by 16 pixels and each pixel's direction will be equal to the normal of the closest road boundary pixel. However, the problem with this is that outside of this dilation, there will be no direction. We show in Figure \ref{fig:direction_methods} that our predicted direction map performs much better than these dilated normals. Here we show the percentage difference between the two models for all our metrics. Not shown in these figures is that our direction map method also produces a connectivity score that is 1\% higher.

\vspace{-5mm}

\paragraph{Cumulative Distribution of Connectivity:}
In Figure \ref{fig:cumul_connectivity} we compare the number of predicted connected boundaries for each ground truth boundary for our model and the baselines. In this figure, we plot the cumulative distribution of the percentage of ground truth boundaries with X number of predicted segments. We show that our model significantly outperforms the baselines. For our model, 99.3\% of the ground truth boundaries have a single predicted boundary whereas for the baselines this number is around 80\%.

\vspace{-5mm}

\paragraph{Qualitative Results:}
In Figure \ref{fig:feat}, we visualize the learnt features of our network given the input modalities. In particular, given the camera, LiDAR and the elevation gradient, our model outputs the dense location of the road boundaries, their endpoints as well as the vector field of the normalized normals to the road boundaries. These features aid the cSnake to output structured polylines for each road boundary as demonstrated in Figure \ref{fig:qual-results}.

\vspace{-5mm}

\paragraph{Stitched Results:} Our model is fully convolutional and independent of the image dimensions. Fig. \ref{fig:stitched} shows stitched
examples of different crops on larger areas.

\vspace{-5mm}

\paragraph{Speed:}
On a Titan 1080 Ti GPU, the average compute per image is 196ms for the deep features and
1.76s for the cSnake module.

\vspace{-5mm}

\paragraph{Failure Cases:}
In Figure \ref{fig:fail}, we demonstrate a few failure cases of our model. On the first row, we can see that two disconnected polylines have been assigned to the same road boundary. In the second row, lower and bottom predictions deviate from the road boundaries before coming back. In the last row, one road boundary has not been captured. %

\section{Conclusion}

In this paper, we proposed a deep fully convolutional model that extracts road boundaries from LiDAR and camera imagery. In particular, a CNN first outputs deep features corresponding to the road boundaries such as their location and directional clues. Then a our cSnake module, which is a convolutional recurrent network, outputs polylines corresponding to each road boundary. We demonstrated the effectiveness of our approach through extensive ablation studies and comparison with a strong baseline. In particular, we achieve F1 score of $87.2\%$ at 5 pixels away from the ground truth road boundaries with a connectivity of $99.2\%$. In the future, we plan to leverage aerial imagery alongside our LiDAR and camera sensors as well as extend our approach to other static elements of the scene.

{\small
\bibliographystyle{ieee_fullname}
\bibliography{ref}

\begin{thebibliography}{10}\itemsep=-1pt

\bibitem{acuna2018efficient}
David Acuna, Huan Ling, Amlan Kar, and Sanja Fidler.
\newblock Efficient interactive annotation of segmentation datasets with
  polygon-rnn++.
\newblock 2018.

\bibitem{alvarez2012road}
Jose~M Alvarez, Theo Gevers, Yann LeCun, and Antonio~M Lopez.
\newblock Road scene segmentation from a single image.
\newblock In {\em ECCV}, 2012.

\bibitem{alvarez2011road}
Jos{\'e} M~{\'A}lvarez Alvarez and Antonio~M Lopez.
\newblock Road detection based on illuminant invariance.
\newblock {\em IEEE Transactions on Intelligent Transportation Systems}, 2011.

\bibitem{fortier1999survey}
Marie-Flavie Auclair-Fortier, Djemel Ziou, and Costas Armenakis.
\newblock Survey of work on road extraction in aerial and satellite images.
\newblock 2002.

\bibitem{bai2018}
Min Bai, Gellert Mattyus, Namdar Homayounfar, Shenlong Wang, Kowshika
  Lakshmikanth, Shrinidhi, and Raquel Urtasun.
\newblock Deep multi-sensor lane detection.
\newblock In {\em IROS}, 2018.

\bibitem{Bajcsy1976ComputerRO}
Ruzena Bajcsy and Mohamad Tavakoli.
\newblock Computer recognition of roads from satellite pictures.
\newblock {\em IEEE Transactions on Systems, Man, and Cybernetics}, 1976.

\bibitem{color_code}
Simon Baker, Daniel Scharstein, J.~P. Lewis, Stefan Roth, Michael~J. Black, and
  Richard Szeliski.
\newblock A database and evaluation methodology for optical flow.
\newblock {\em International Journal of Computer Vision}, 2011.

\bibitem{barsan2018robust}
Ioan~Andrei B{\^a}rsan, Peidong Liu, Marc Pollefeys, and Andreas Geiger.
\newblock Robust dense mapping for large-scale dynamic environments.
\newblock In {\em ICRA}, 2018.

\bibitem{deep-gill}
Ioan~Andrei Barsan, Shenlong Wang, Andrei Pokrovsky, and Raquel Urtasun.
\newblock Learning to localize using a lidar intensity map.
\newblock In {\em Proceedings of The 2nd Conference on Robot Learning}, 2018.

\bibitem{bastani2018roadtracer}
Favyen Bastani, Songtao He, Sofiane Abbar, Mohammad Alizadeh, Hari
  Balakrishnan, Sanjay Chawla, Sam Madden, and David DeWitt.
\newblock Roadtracer: Automatic extraction of road networks from aerial images.
\newblock In {\em CVPR}, 2018.

\bibitem{butenuth2012}
Matthias Butenuth and Christian Heipke.
\newblock Network snakes: Graph-based object delineation with active contour
  models.
\newblock {\em Mach. Vis. Appl.}, 2012.

\bibitem{casas2018intentnet}
Sergio Casas, Wenjie Luo, and Raquel Urtasun.
\newblock Intentnet: Learning to predict intention from raw sensor data.
\newblock In {\em Conference on Robot Learning}, 2018.

\bibitem{CastrejonCVPR17}
Lluis Castrejon, Kaustav Kundu, Raquel Urtasun, and Sanja Fidler.
\newblock Annotating object instances with a polygon-rnn.
\newblock In {\em CVPR}, 2017.

\bibitem{linknet}
Abhishek Chaurasia and Eugenio Culurciello.
\newblock Linknet: Exploiting encoder representations for efficient semantic
  segmentation.
\newblock {\em CoRR}, 2017.

\bibitem{cheng2006lane}
Hsu-Yung Cheng, Bor-Shenn Jeng, Pei-Ting Tseng, and Kuo-Chin Fan.
\newblock Lane detection with moving vehicles in the traffic scenes.
\newblock {\em IEEE Transactions on intelligent transportation systems}, 2006.

\bibitem{geiger2011stereoscan}
Andreas Geiger, Julius Ziegler, and Christoph Stiller.
\newblock Stereoscan: Dense 3d reconstruction in real-time.
\newblock In {\em Intelligent Vehicles Symposium (IV), 2011 IEEE}, 2011.

\bibitem{Homayounfar_2018_CVPR}
Namdar Homayounfar, Wei-Chiu Ma, Shrinidhi Kowshika~Lakshmikanth, and Raquel
  Urtasun.
\newblock Hierarchical recurrent attention networks for structured online maps.
\newblock In {\em CVPR}, 2018.

\bibitem{jaderberg2015spatial}
Max Jaderberg, Karen Simonyan, Andrew Zisserman, et~al.
\newblock Spatial transformer networks.
\newblock In {\em NIPS}, 2015.

\bibitem{kammel2008lidar}
Soren Kammel and Benjamin Pitzer.
\newblock Lidar-based lane marker detection and mapping.
\newblock In {\em Intelligent Vehicles Symposium, 2008 IEEE}, 2008.

\bibitem{kass1988snakes}
Michael Kass, Andrew Witkin, and Demetri Terzopoulos.
\newblock Snakes: Active contour models.
\newblock {\em International journal of computer vision}, 1988.

\bibitem{ADAM}
Diederik Kingma and Jimmy Ba.
\newblock Adam: A method for stochastic optimization.
\newblock {\em ICLR}, 2015.

\bibitem{kong2010general}
Hui Kong, Jean-Yves Audibert, and Jean Ponce.
\newblock General road detection from a single image.
\newblock {\em TIP}, 2010.

\bibitem{kuhnl2012spatial}
Tobias K{\"u}hnl, Franz Kummert, and Jannik Fritsch.
\newblock Spatial ray features for real-time ego-lane extraction.
\newblock In {\em Intelligent Transportation Systems (ITSC), 2012 15th
  International IEEE Conference on}, 2012.

\bibitem{laptev2000automatic}
Ivan Laptev, Helmut Mayer, Tony Lindeberg, Wolfgang Eckstein, Carsten Steger,
  and Albert Baumgartner.
\newblock Automatic extraction of roads from aerial images based on scale space
  and snakes.
\newblock {\em Machine Vision and Applications}, 2000.

\bibitem{levi2015stixelnet}
Dan Levi, Noa Garnett, Ethan Fetaya, and Israel Herzlyia.
\newblock Stixelnet: A deep convolutional network for obstacle detection and
  road segmentation.
\newblock In {\em BMVC}, 2015.

\bibitem{liang2018end}
Justin Liang and Raquel Urtasun.
\newblock End-to-end deep structured models for drawing crosswalks.
\newblock In {\em ECCV}, 2018.

\bibitem{liang2018deep}
Ming Liang, Bin Yang, Shenlong Wang, and Raquel Urtasun.
\newblock Deep continuous fusion for multi-sensor 3d object detection.
\newblock In {\em ECCV}, 2018.

\bibitem{lieb2005adaptive}
David Lieb, Andrew Lookingbill, and Sebastian Thrun.
\newblock Adaptive road following using self-supervised learning and reverse
  optical flow.
\newblock In {\em RSS}, 2005.

\bibitem{lin2016feature}
Tsung-Yi Lin, Piotr Doll{\'a}r, Ross Girshick, Kaiming He, Bharath Hariharan,
  and Serge Belongie.
\newblock Feature pyramid networks for object detection.
\newblock {\em arXiv}, 2016.

\bibitem{ma2017find}
Wei-Chiu Ma, Shenlong Wang, Marcus~A Brubaker, Sanja Fidler, and Raquel
  Urtasun.
\newblock Find your way by observing the sun and other semantic cues.
\newblock In {\em ICRA}, 2017.

\bibitem{marcos2018learning}
Diego Marcos, Devis Tuia, Benjamin Kellenberger, Lisa Zhang, Min Bai, Renjie
  Liao, and Raquel Urtasun.
\newblock Learning deep structured active contours end-to-end.
\newblock In {\em CVPR}, 2018.

\bibitem{marikhu2007family}
Ramesh Marikhu, Matthew~N Dailey, Stanislav Makhanov, and Kiyoshi Honda.
\newblock A family of quadratic snakes for road extraction.
\newblock In {\em Asian Conference on Computer Vision}, 2007.

\bibitem{marmanis2018classification}
Dimitrios Marmanis, Konrad Schindler, Jan~Dirk Wegner, Silvano Galliani, Mihai
  Datcu, and Uwe Stilla.
\newblock Classification with an edge: Improving semantic image segmentation
  with boundary detection.
\newblock {\em ISPRS Journal of Photogrammetry and Remote Sensing}, 2018.

\bibitem{marmanis2016semantic}
Dimitrios Marmanis, Jan~D Wegner, Silvano Galliani, Konrad Schindler, Mihai
  Datcu, and Uwe Stilla.
\newblock Semantic segmentation of aerial images with an ensemble of cnss.
\newblock {\em ISPRS Annals of the Photogrammetry, Remote Sensing and Spatial
  Information Sciences, 2016}, 2016.

\bibitem{mattyus2017deeproadmapper}
Gell{\'e}rt M{\'a}ttyus, Wenjie Luo, and Raquel Urtasun.
\newblock Deeproadmapper: Extracting road topology from aerial images.
\newblock In {\em ICCV}, 2017.

\bibitem{mattyus2015enhancing}
Gellert Mattyus, Shenlong Wang, Sanja Fidler, and Raquel Urtasun.
\newblock Enhancing road maps by parsing aerial images around the world.
\newblock In {\em ICCV}, 2015.

\bibitem{mattyus2016hd}
Gell{\'e}rt M{\'a}ttyus, Shenlong Wang, Sanja Fidler, and Raquel Urtasun.
\newblock Hd maps: Fine-grained road segmentation by parsing ground and aerial
  images.
\newblock In {\em CVPR}, 2016.

\bibitem{mayer1998multi}
Helmut Mayer, Ivan Laptev, and Albert Baumgartner.
\newblock Multi-scale and snakes for automatic road extraction.
\newblock In {\em ECCV}, 1998.

\bibitem{mena2005automatic}
Juan~B Mena and Jos{\'e}~A Malpica.
\newblock An automatic method for road extraction in rural and semi-urban areas
  starting from high resolution satellite imagery.
\newblock {\em Pattern recognition letters}, 2005.

\bibitem{mnih2010learning}
Volodymyr Mnih and Geoffrey~E Hinton.
\newblock Learning to detect roads in high-resolution aerial images.
\newblock In {\em ECCV}, 2010.

\bibitem{mnih2012learning}
Volodymyr Mnih and Geoffrey~E Hinton.
\newblock Learning to label aerial images from noisy data.
\newblock In {\em ICML}, 2012.

\bibitem{mohan2014deep}
Rahul Mohan.
\newblock Deep deconvolutional networks for scene parsing.
\newblock {\em arXiv}, 2014.

\bibitem{montoya2014mind}
Javier~A Montoya-Zegarra, Jan~D Wegner, L'ubor Ladick{\`y}, and Konrad
  Schindler.
\newblock Mind the gap: modeling local and global context in (road) networks.
\newblock In {\em German Conference on Pattern Recognition}, 2014.

\bibitem{moravec1985high}
Hans~P Moravec and Alberto Elfes.
\newblock High resolution maps from wide angle sonar.
\newblock In {\em ICRA}, 1985.

\bibitem{paz2015variational}
Lina~Maria Paz, Pedro Pini{\'e}s, and Paul Newman.
\newblock A variational approach to online road and path segmentation with
  monocular vision.
\newblock In {\em ICRA}, 2015.

\bibitem{pire2018real}
Taih{\'u} Pire, Rodrigo Baravalle, Ariel D'Alessandro, and Javier Civera.
\newblock Real-time dense map fusion for stereo slam.
\newblock {\em Robotica}, 2018.

\bibitem{pollefeys2008detailed}
Marc Pollefeys, David Nist{\'e}r, J-M Frahm, Amir Akbarzadeh, Philippos
  Mordohai, Brian Clipp, Chris Engels, David Gallup, S-J Kim, Paul Merrell,
  et~al.
\newblock Detailed real-time urban 3d reconstruction from video.
\newblock {\em International Journal of Computer Vision}, 2008.

\bibitem{richards_2013}
John~A. Richards and Xiuping Jia.
\newblock {\em Remote Sensing Digital Image Analysis}.
\newblock 2013.

\bibitem{rochery2006higher}
Marie Rochery, Ian~H Jermyn, and Josiane Zerubia.
\newblock Higher order active contours.
\newblock {\em International Journal of Computer Vision}, 2006.

\bibitem{simonett1970use}
David~S Simonett, Floyd~M Henderson, and Dwight~D Egbert.
\newblock On the use of space photography for identifying transportation
  routes: A summary of problems.
\newblock 1970.

\bibitem{tan2006color}
Ceryen Tan, Tsai Hong, Tommy Chang, and Michael Shneier.
\newblock Color model-based real-time learning for road following.
\newblock In {\em Intelligent Transportation Systems Conference, 2006. ITSC'06.
  IEEE}, 2006.

\bibitem{ventura2018iterative}
Carles Ventura, Jordi Pont-Tuset, Sergi Caelles, Kevis-Kokitsi Maninis, and Luc
  Van~Gool.
\newblock Iterative deep learning for road topology extraction.
\newblock {\em arXiv}, 2018.

\bibitem{TCity2017}
Shenlong Wang, Min Bai, Gellert Mattyus, Hang Chu, Wenjie Luo, Bin Yang, Justin
  Liang, Joel Cheverie, Sanja Fidler, and Raquel Urtasun.
\newblock Torontocity: Seeing the world with a million eyes.
\newblock In {\em ICCV}, 2017.

\bibitem{wedel2009b}
Andreas Wedel, Hern{\'a}n Badino, Clemens Rabe, Heidi Loose, Uwe Franke, and
  Daniel Cremers.
\newblock B-spline modeling of road surfaces with an application to free-space
  estimation.
\newblock {\em IEEE Transactions on Intelligent Transportation Systems}, 2009.

\bibitem{wegner2013higher}
Jan~D Wegner, Javier~A Montoya-Zegarra, and Konrad Schindler.
\newblock A higher-order crf model for road network extraction.
\newblock In {\em CVPR}, 2013.

\bibitem{wegner2015road}
Jan~Dirk Wegner, Javier~Alexander Montoya-Zegarra, and Konrad Schindler.
\newblock Road networks as collections of minimum cost paths.
\newblock {\em ISPRS Journal of Photogrammetry and Remote Sensing}, 2015.

\bibitem{yao2015estimating}
Jian Yao, Srikumar Ramalingam, Yuichi Taguchi, Yohei Miki, and Raquel Urtasun.
\newblock Estimating drivable collision-free space from monocular video.
\newblock In {\em Applications of Computer Vision (WACV), 2015 IEEE Winter
  Conference on}, 2015.

\bibitem{zhang2018efficient}
Chris Zhang, Wenjie Luo, and Raquel Urtasun.
\newblock Efficient convolutions for real-time semantic segmentation of 3d
  point clouds.
\newblock In {\em 2018 International Conference on 3D Vision (3DV)}, 2018.

\end{thebibliography}
}

\onecolumn
\begin{appendices}
\vspace{-10.0mm}
\section{Outline}
\begin{figure*}[!h]
    \centering
	\[\arraycolsep=1.0pt
	\begin{array}{cccccc}

	\includegraphics[width=0.155\linewidth]{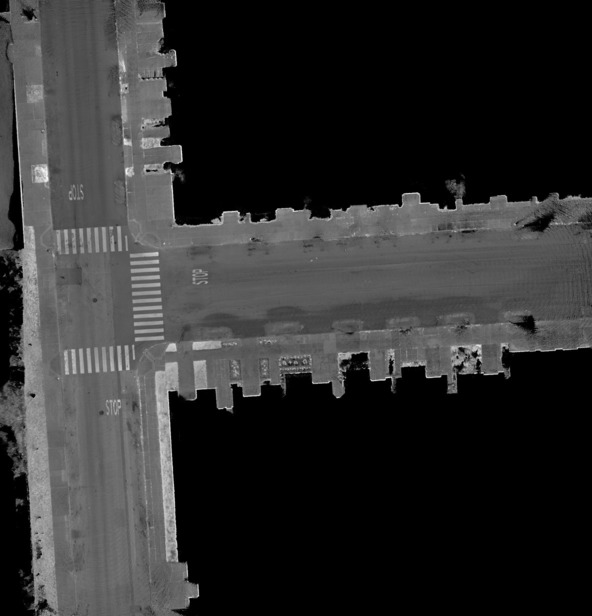}  & 
	\includegraphics[width=0.155\linewidth]{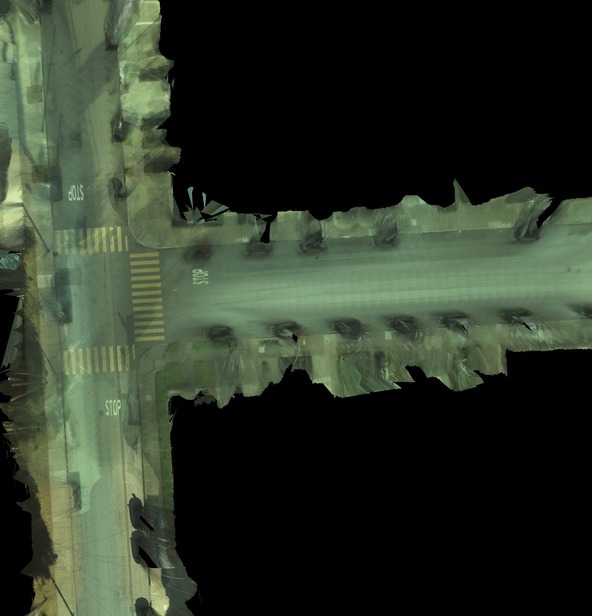}&  
	\includegraphics[width=0.155\linewidth]{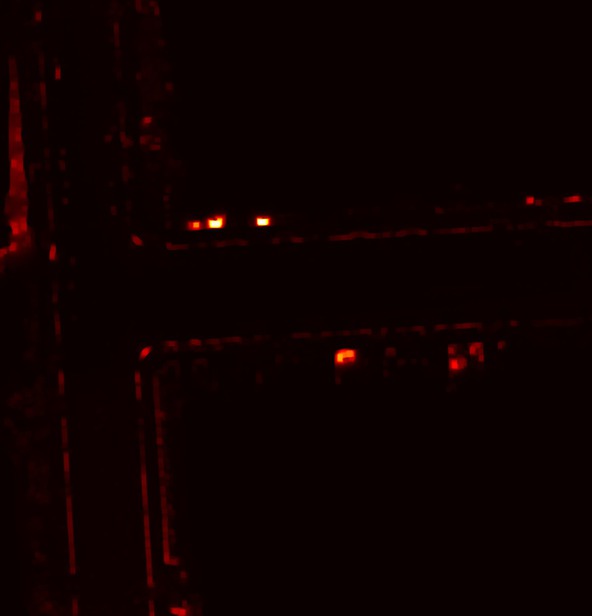} &
	\includegraphics[width=0.155\linewidth]{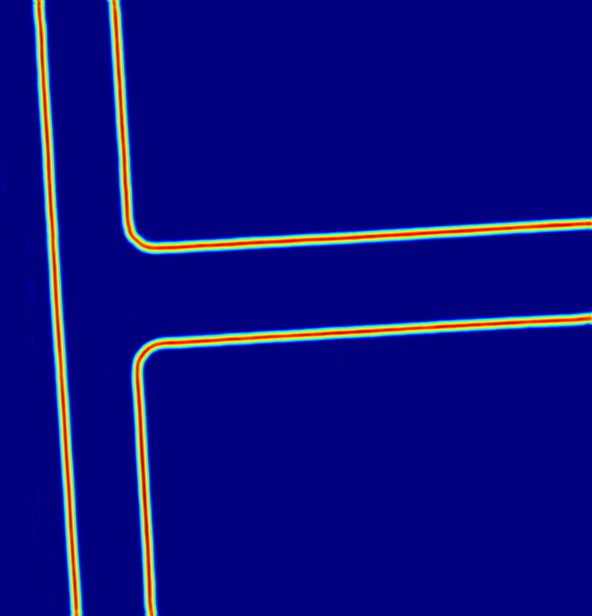}  & 
	\includegraphics[width=0.155\linewidth]{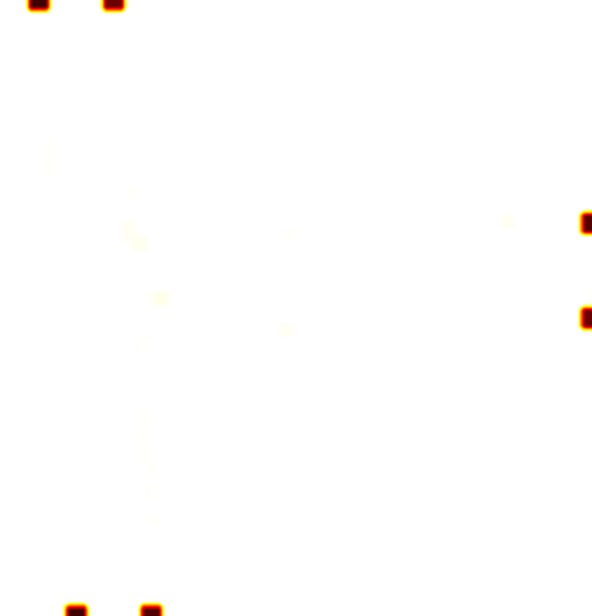}& 
	\includegraphics[width=0.155\linewidth]{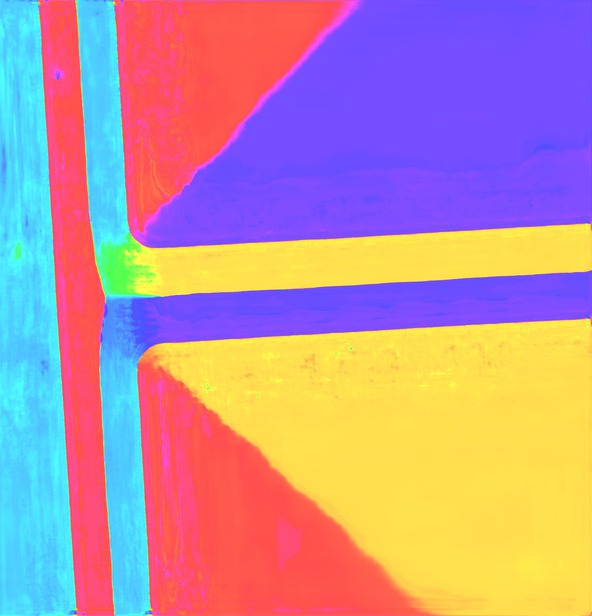} \\

	\includegraphics[width=0.155\linewidth]{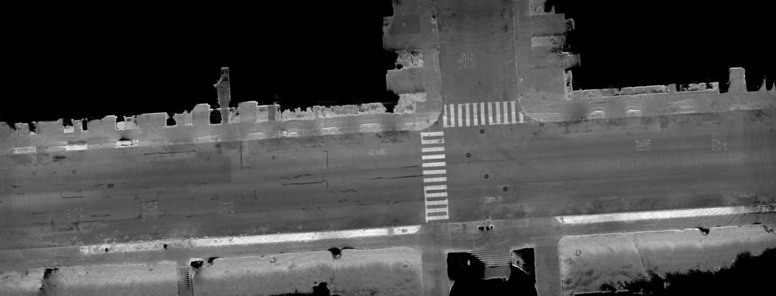}  & 
	\includegraphics[width=0.155\linewidth]{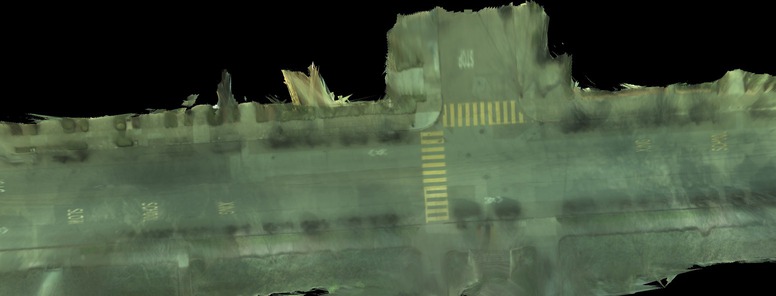}&  
	\includegraphics[width=0.155\linewidth]{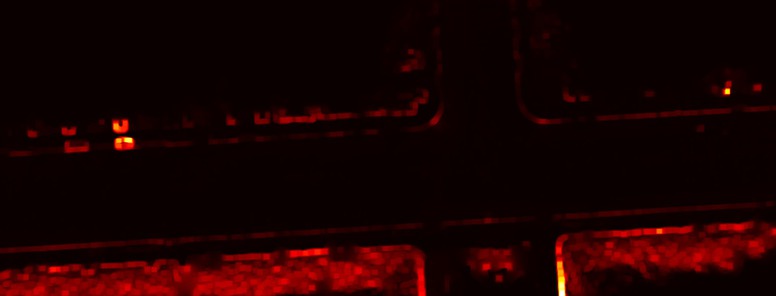} &
	\includegraphics[width=0.155\linewidth]{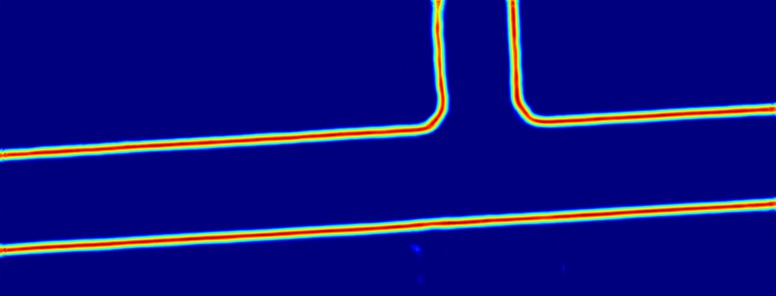}  & 
	\includegraphics[width=0.155\linewidth]{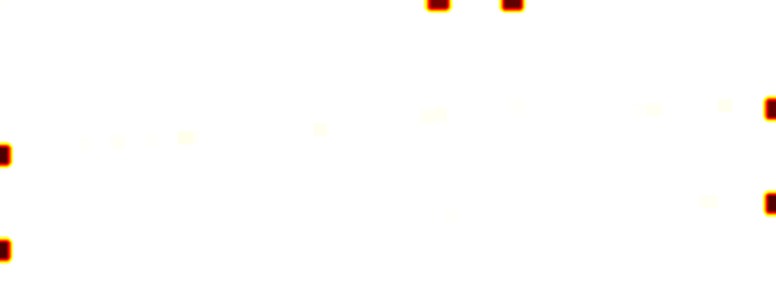}& 
	\includegraphics[width=0.155\linewidth]{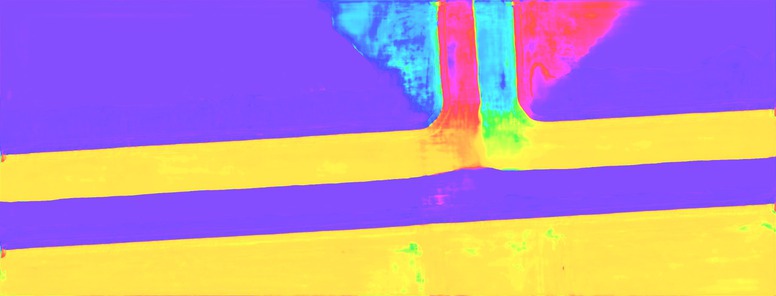} \\

	\includegraphics[width=0.155\linewidth]{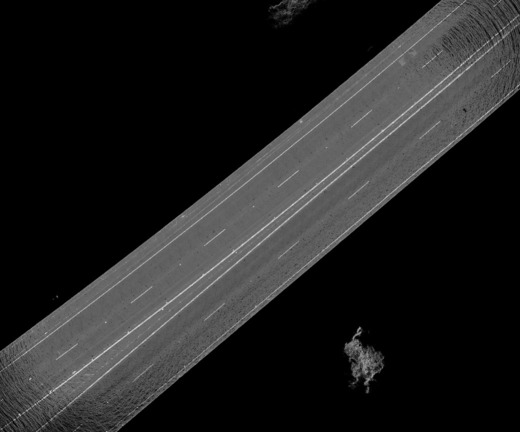}  & 
	\includegraphics[width=0.155\linewidth]{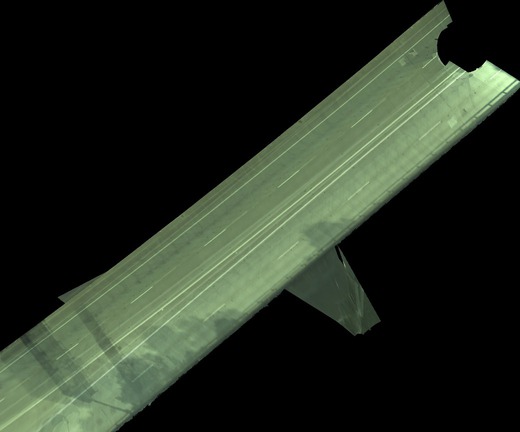}&  
	\includegraphics[width=0.155\linewidth]{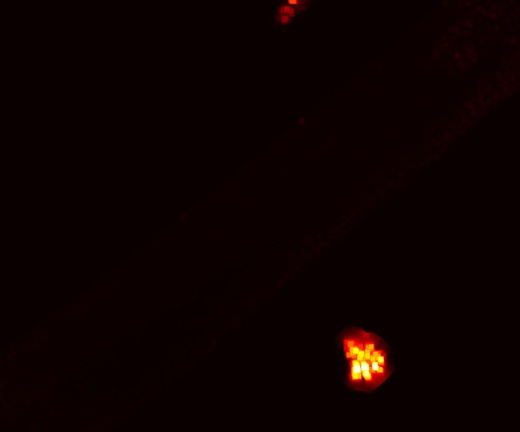} &
	\includegraphics[width=0.155\linewidth]{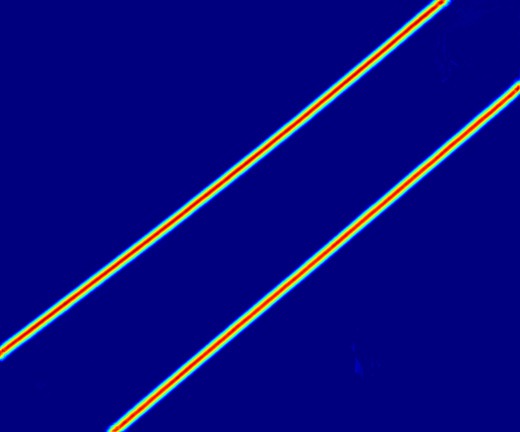}  & 
	\includegraphics[width=0.155\linewidth]{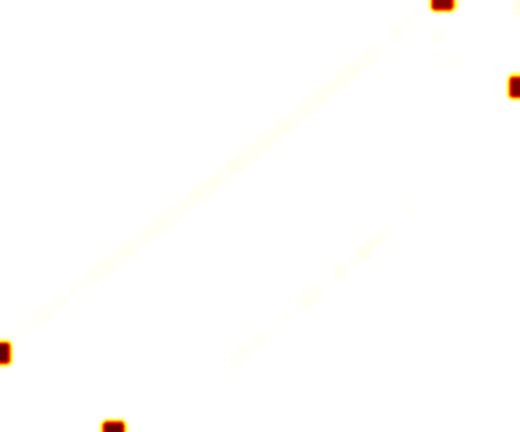}& 
	\includegraphics[width=0.155\linewidth]{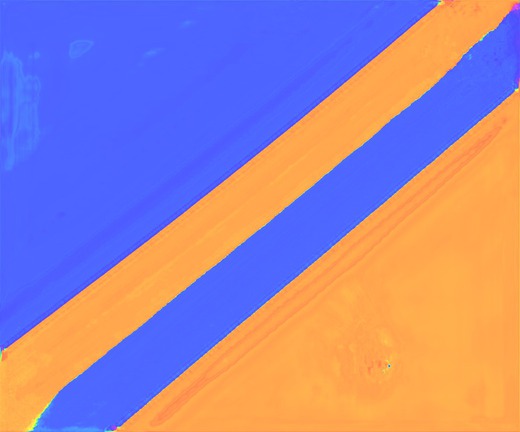} \\

	\includegraphics[width=0.155\linewidth]{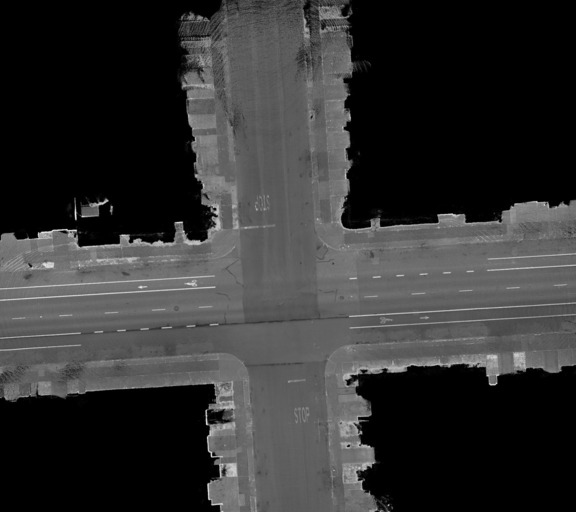}  & 
	\includegraphics[width=0.155\linewidth]{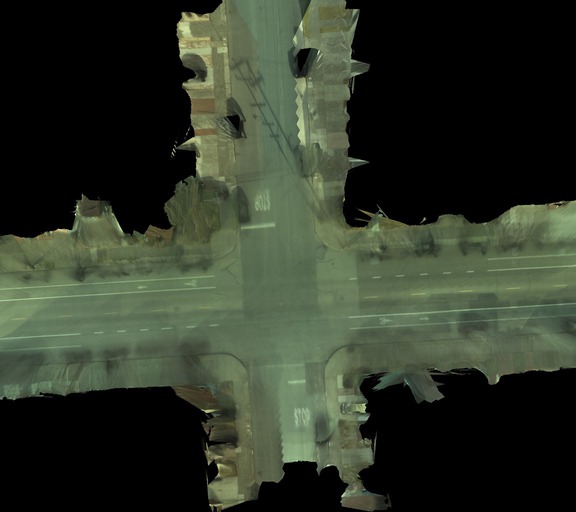}&  
	\includegraphics[width=0.155\linewidth]{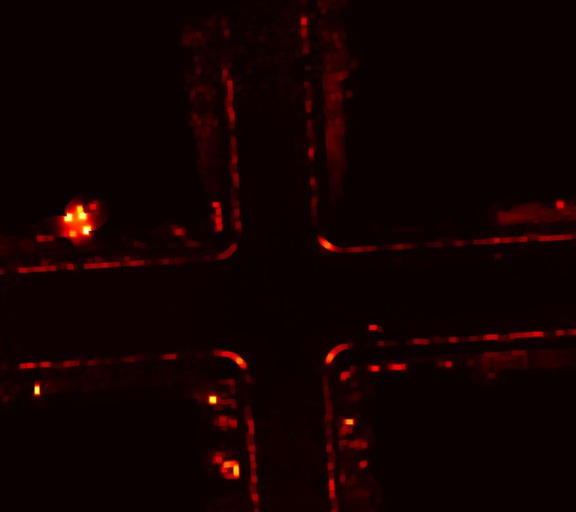} &
	\includegraphics[width=0.155\linewidth]{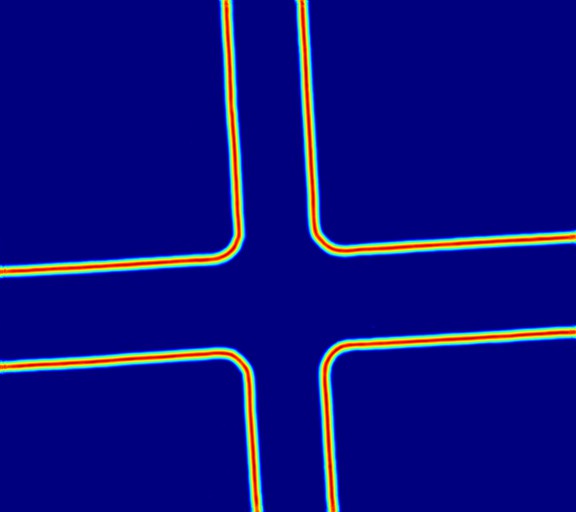}  & 
	\includegraphics[width=0.155\linewidth]{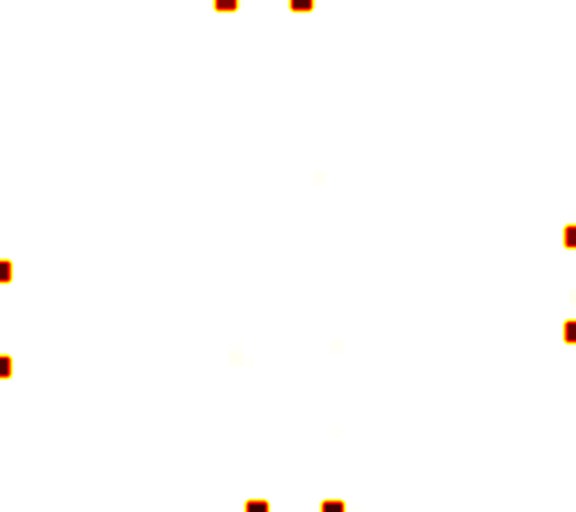}& 
	\includegraphics[width=0.155\linewidth]{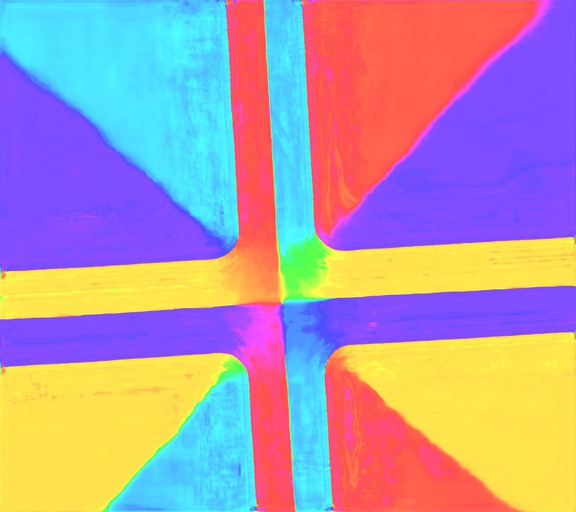} \\

	\includegraphics[width=0.155\linewidth]{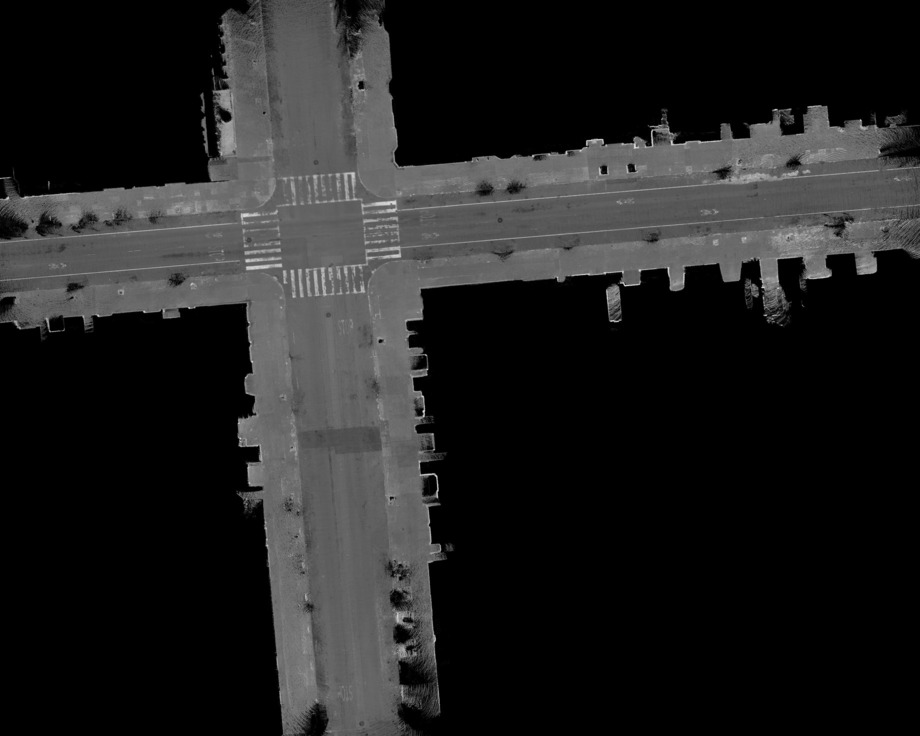}  & 
	\includegraphics[width=0.155\linewidth]{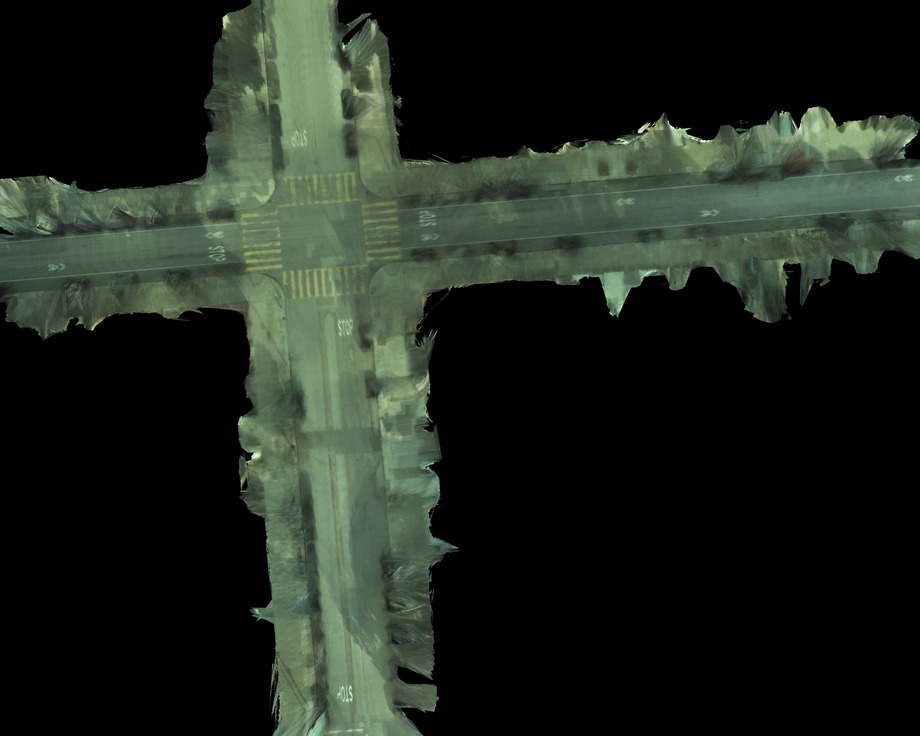}&  
	\includegraphics[width=0.155\linewidth]{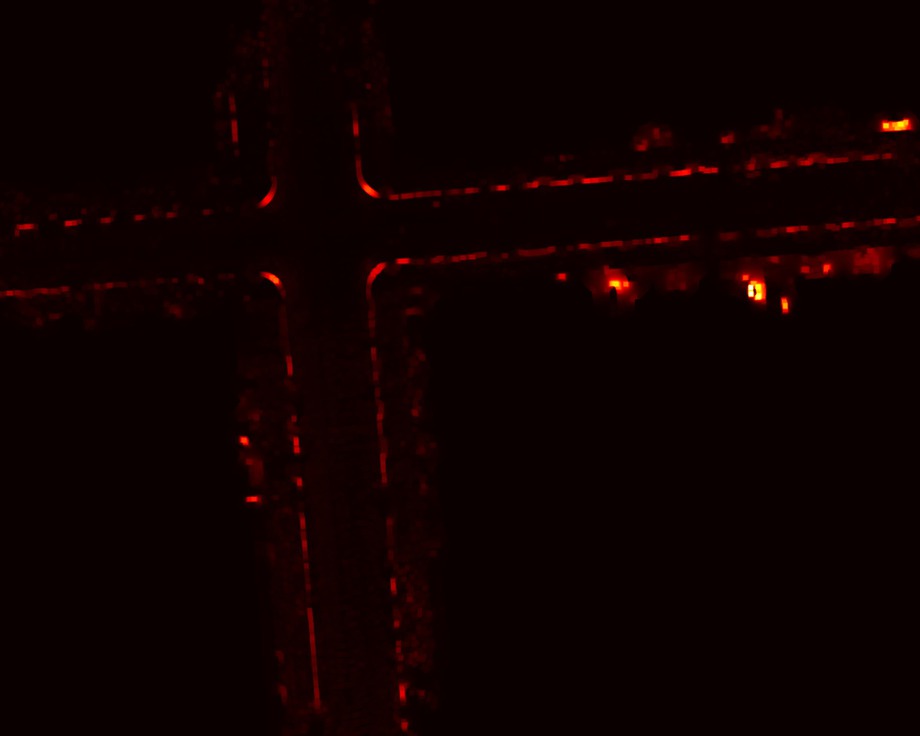} &
	\includegraphics[width=0.155\linewidth]{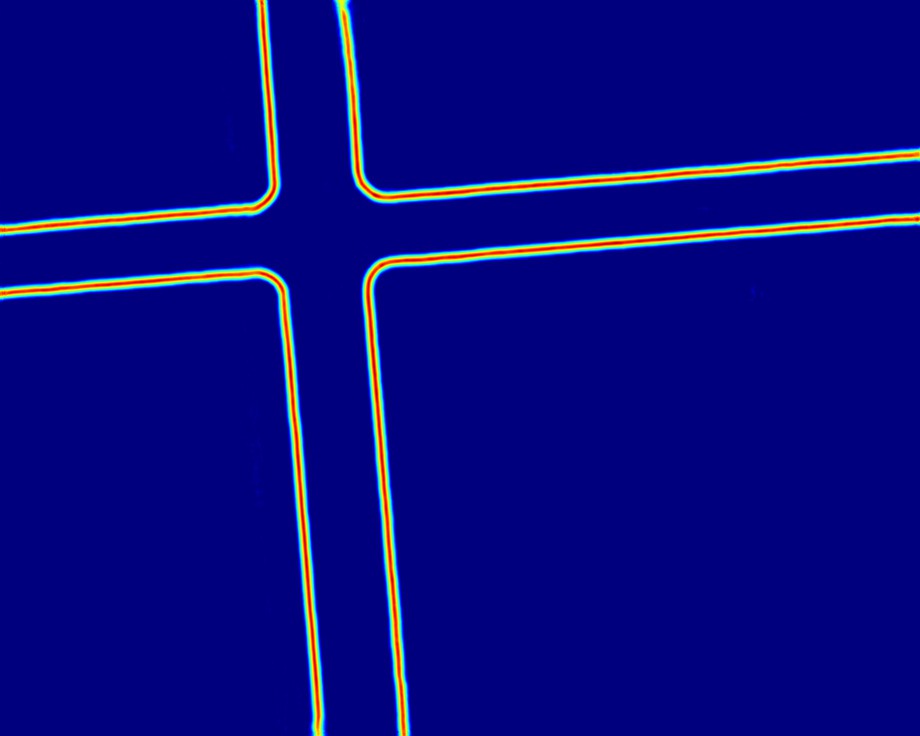}  & 
	\includegraphics[width=0.155\linewidth]{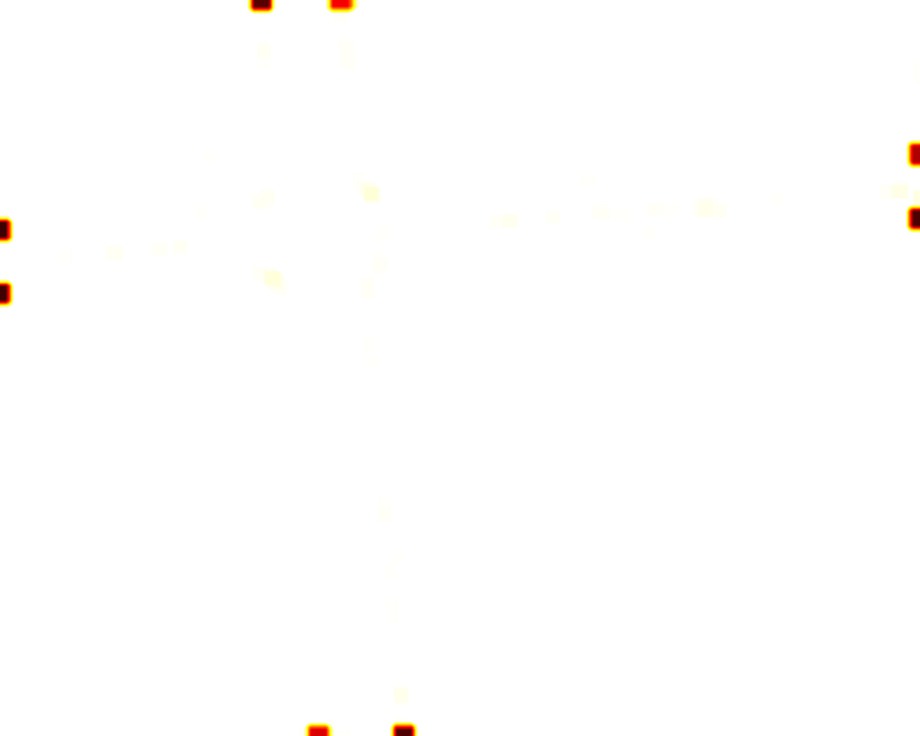}& 
	\includegraphics[width=0.155\linewidth]{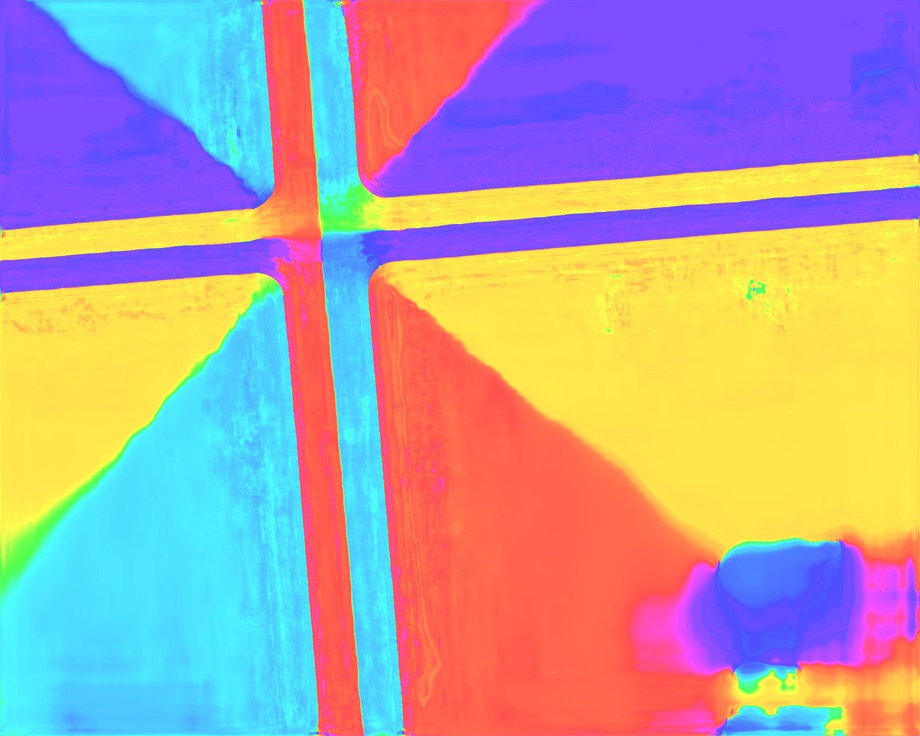} \\

	\raisebox{2px}{{(Lidar)}} &	
	\raisebox{2px}{{(Camera)}} &
	\raisebox{2px}{{(Elevation Gradient)}} &
	\raisebox{2px}{{(Detection Map)}} &
	\raisebox{2px}{{(Endpoints)}} &
	\raisebox{2px}{{(Direction Map)}}

	\end{array}
	\]

	\caption{Deep Features: Columns \textbf{(1-3)} correspond to the  inputs and columns \textbf{(4-6)} correspond to the deep feature maps. The direction map shown here as a flow field \cite{color_code}.}
	\label{fig:feat1}
\end{figure*}

\begin{figure*}[h]
    \centering
	\[\arraycolsep=1.0pt
	\begin{array}{cccccc}

	\includegraphics[width=0.155\linewidth]{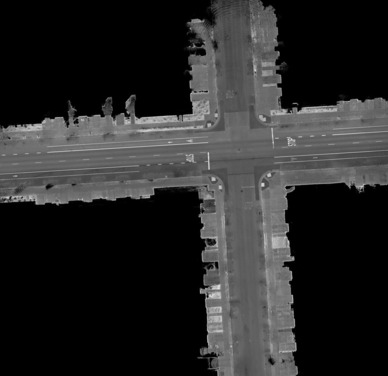}  & 
	\includegraphics[width=0.155\linewidth]{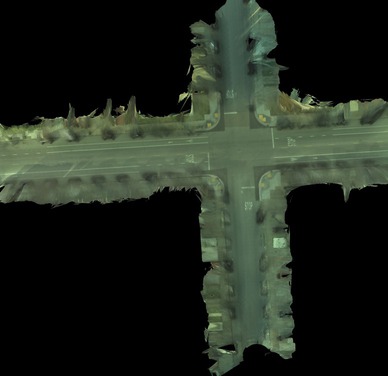}&  
	\includegraphics[width=0.155\linewidth]{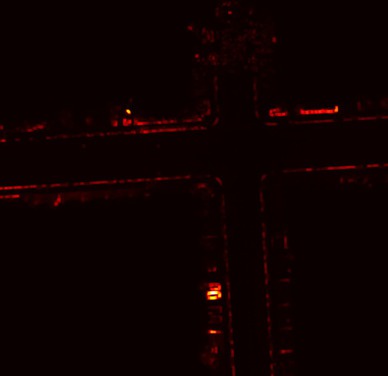} &
	\includegraphics[width=0.155\linewidth]{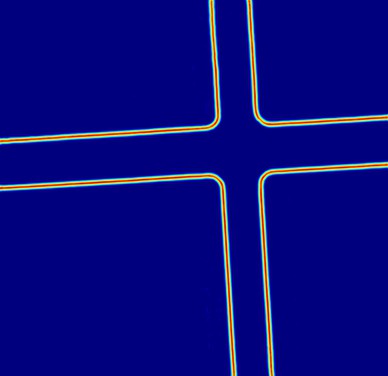}  & 
	\includegraphics[width=0.155\linewidth]{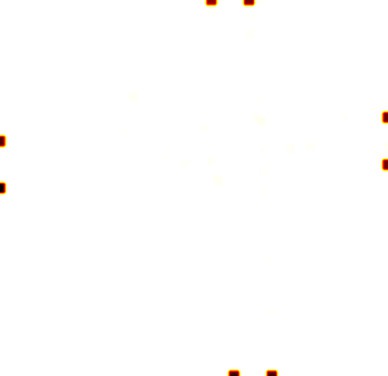}& 
	\includegraphics[width=0.155\linewidth]{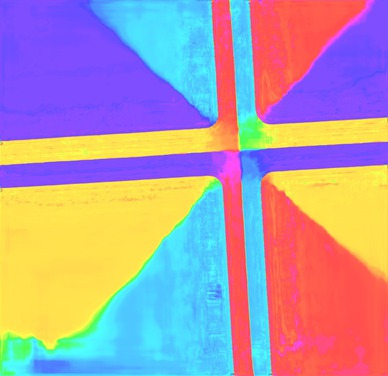} \\

	\includegraphics[width=0.155\linewidth]{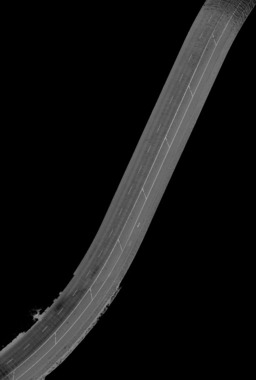}  & 
	\includegraphics[width=0.155\linewidth]{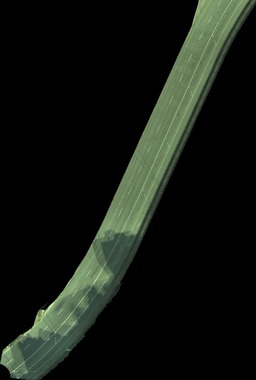}&  
	\includegraphics[width=0.155\linewidth]{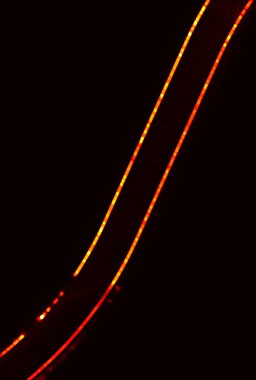} &
	\includegraphics[width=0.155\linewidth]{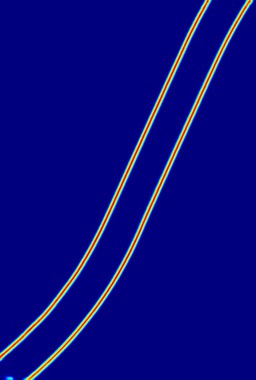}  & 
	\includegraphics[width=0.155\linewidth]{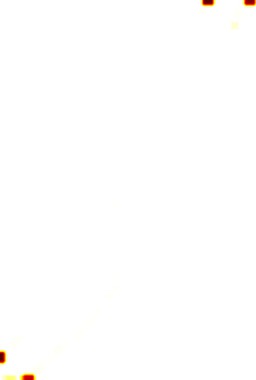}& 
	\includegraphics[width=0.155\linewidth]{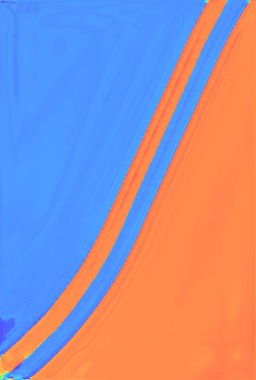} \\
		
	\includegraphics[width=0.155\linewidth]{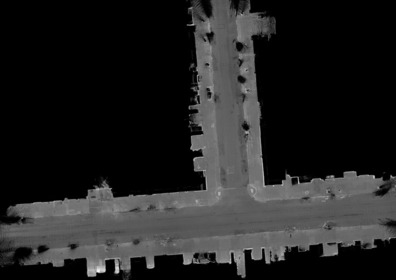}  & 
	\includegraphics[width=0.155\linewidth]{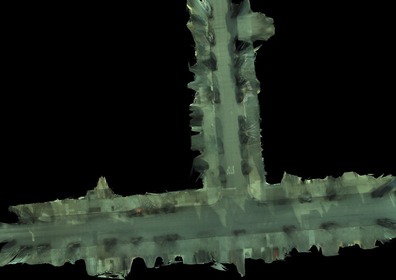}&  
	\includegraphics[width=0.155\linewidth]{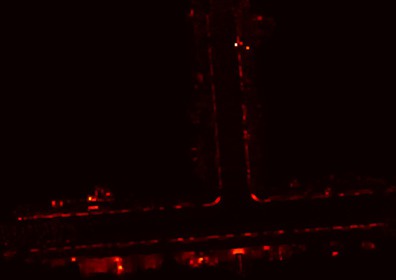} &
	\includegraphics[width=0.155\linewidth]{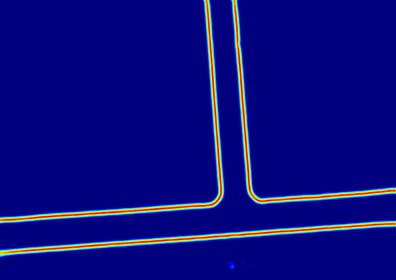}  & 
	\includegraphics[width=0.155\linewidth]{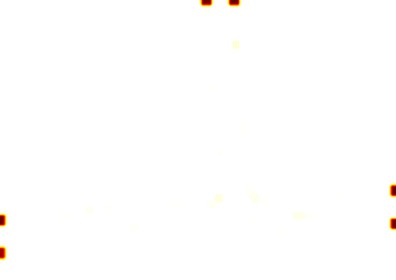}& 
	\includegraphics[width=0.155\linewidth]{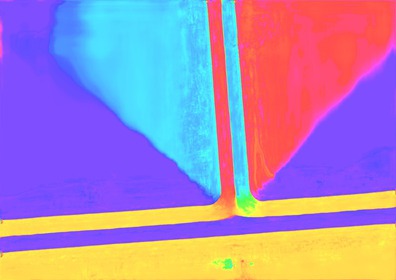} \\

	\includegraphics[width=0.155\linewidth]{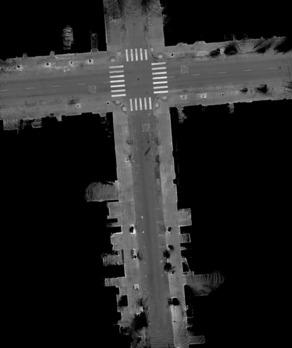}  & 
	\includegraphics[width=0.155\linewidth]{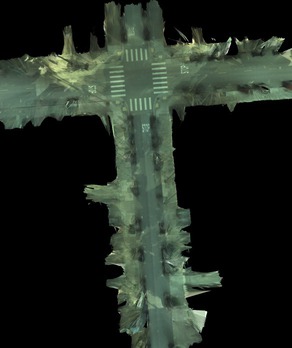}&  
	\includegraphics[width=0.155\linewidth]{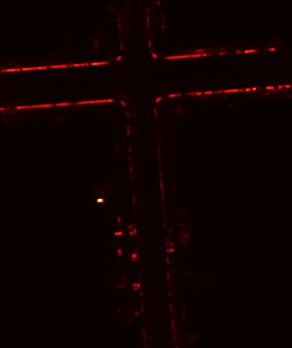} &
	\includegraphics[width=0.155\linewidth]{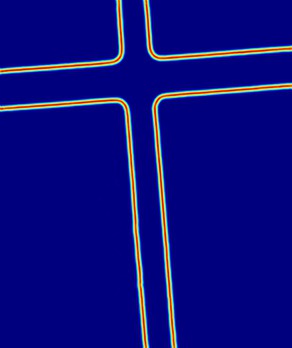}  & 
	\includegraphics[width=0.155\linewidth]{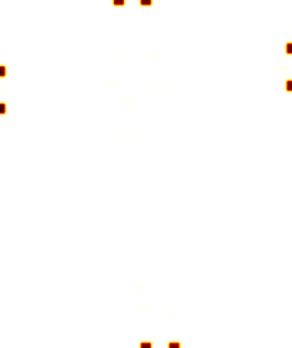}& 
	\includegraphics[width=0.155\linewidth]{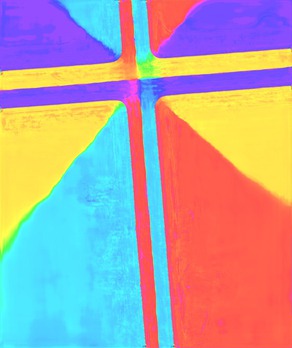} \\

	\includegraphics[width=0.155\linewidth]{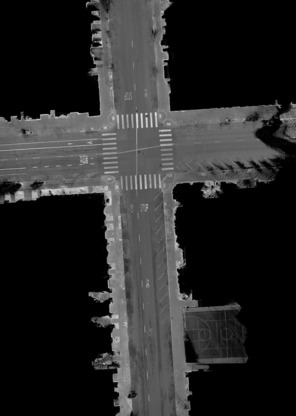}  & 
	\includegraphics[width=0.155\linewidth]{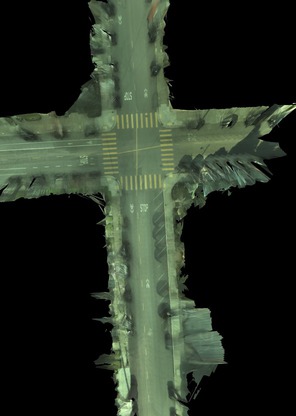}&  
	\includegraphics[width=0.155\linewidth]{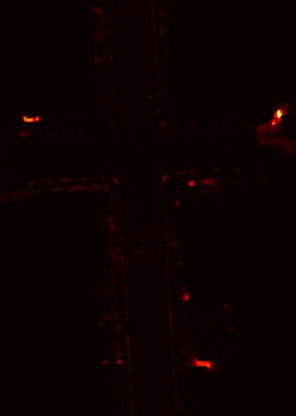} &
	\includegraphics[width=0.155\linewidth]{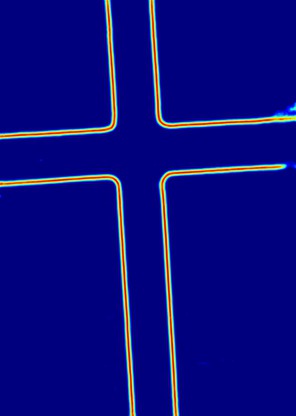}  & 
	\includegraphics[width=0.155\linewidth]{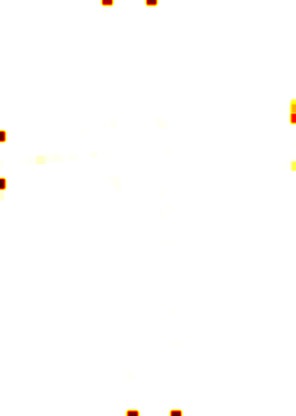}& 
	\includegraphics[width=0.155\linewidth]{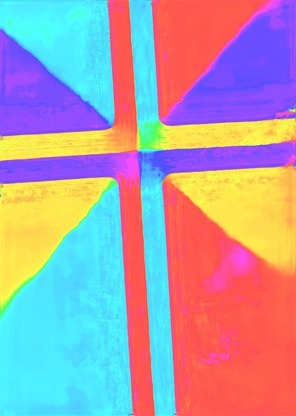} \\

	\includegraphics[width=0.155\linewidth]{supp_figs/feat/sfo_f78e509f-3730-4af2-dc9a-200d09fc54f5_lidar.jpg}  & 
	\includegraphics[width=0.155\linewidth]{supp_figs/feat/sfo_f78e509f-3730-4af2-dc9a-200d09fc54f5_camera.jpg}&  
	\includegraphics[width=0.155\linewidth]{supp_figs/feat/sfo_f78e509f-3730-4af2-dc9a-200d09fc54f5_elevation.jpg} &
	\includegraphics[width=0.155\linewidth]{supp_figs/feat/sfo_f78e509f-3730-4af2-dc9a-200d09fc54f5_pred_dt.jpg}  & 
	\includegraphics[width=0.155\linewidth]{supp_figs/feat/sfo_f78e509f-3730-4af2-dc9a-200d09fc54f5_pred_end.jpg}& 
	\includegraphics[width=0.155\linewidth]{supp_figs/feat/sfo_f78e509f-3730-4af2-dc9a-200d09fc54f5_pred_dir.jpg} \\

	\raisebox{2px}{{(Lidar)}} &	
	\raisebox{2px}{{(Camera)}} &
	\raisebox{2px}{{(Elevation Gradient)}} &
	\raisebox{2px}{{(Detection Map)}} &
	\raisebox{2px}{{(Endpoints)}} &
	\raisebox{2px}{{(Direction Map)}}

	\end{array}
	\]
	
	\caption{Deep Features: Columns \textbf{(1-3)} correspond to the  inputs and columns \textbf{(4-6)} correspond to the deep feature maps. The direction map shown here as a flow field \cite{color_code}.}
	\label{fig:feat2}
\end{figure*}

\begin{figure*}[h]
    \centering
	\[\arraycolsep=1.0pt
	\begin{array}{cccccc}

	\includegraphics[width=0.155\linewidth]{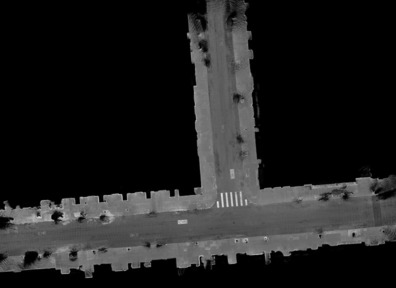}  & 
	\includegraphics[width=0.155\linewidth]{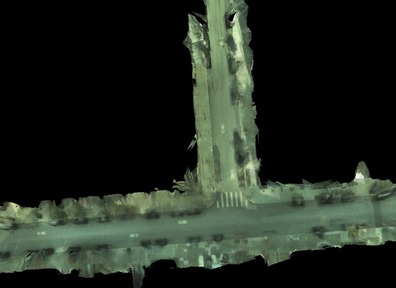}&  
	\includegraphics[width=0.155\linewidth]{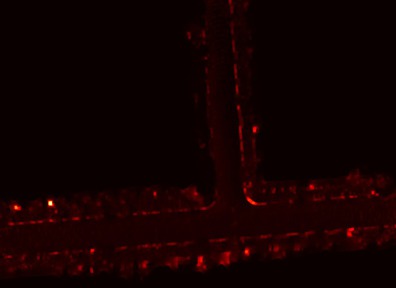} &
	\includegraphics[width=0.155\linewidth]{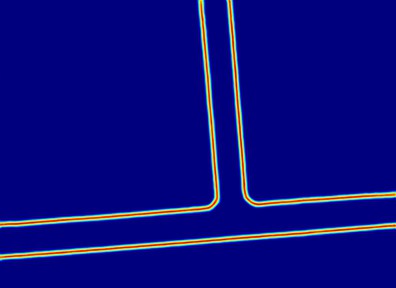}  & 
	\includegraphics[width=0.155\linewidth]{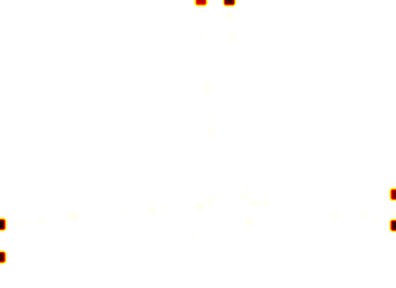}& 
	\includegraphics[width=0.155\linewidth]{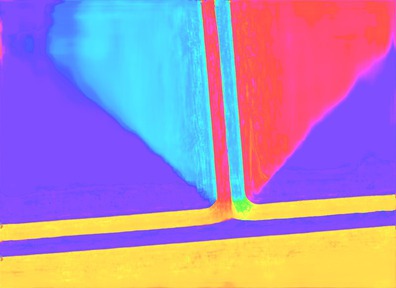} \\

	\includegraphics[width=0.155\linewidth]{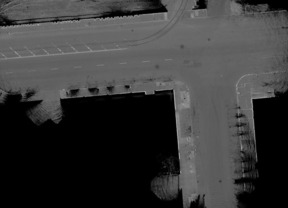}  & 
	\includegraphics[width=0.155\linewidth]{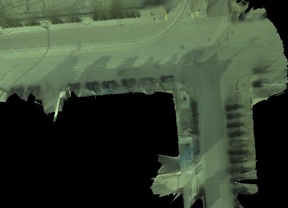}&  
	\includegraphics[width=0.155\linewidth]{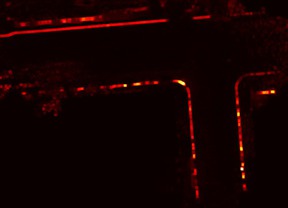} &
	\includegraphics[width=0.155\linewidth]{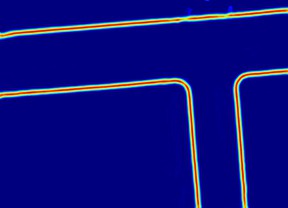}  & 
	\includegraphics[width=0.155\linewidth]{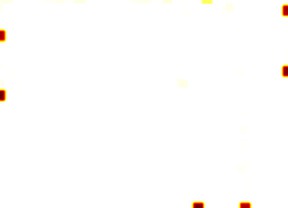}& 
	\includegraphics[width=0.155\linewidth]{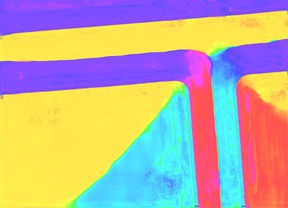} \\

	\includegraphics[width=0.155\linewidth]{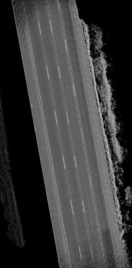}  & 
	\includegraphics[width=0.155\linewidth]{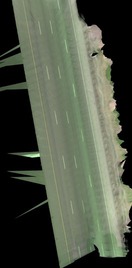}&  
	\includegraphics[width=0.155\linewidth]{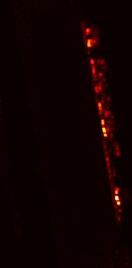} &
	\includegraphics[width=0.155\linewidth]{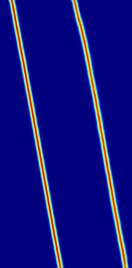}  & 
	\includegraphics[width=0.155\linewidth]{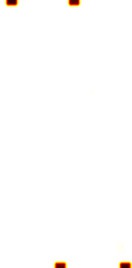}& 
	\includegraphics[width=0.155\linewidth]{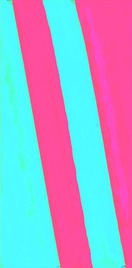} \\

	\includegraphics[width=0.155\linewidth]{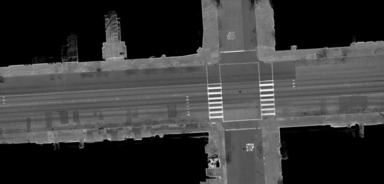}  & 
	\includegraphics[width=0.155\linewidth]{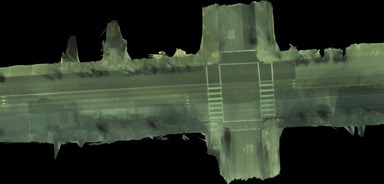}&  
	\includegraphics[width=0.155\linewidth]{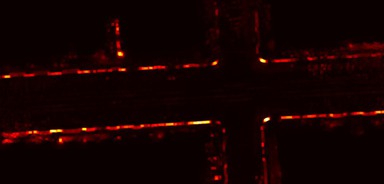} &
	\includegraphics[width=0.155\linewidth]{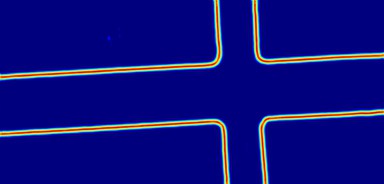}  & 
	\includegraphics[width=0.155\linewidth]{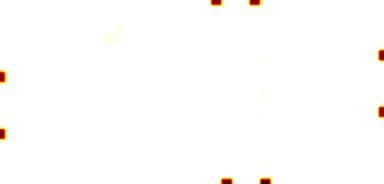}& 
	\includegraphics[width=0.155\linewidth]{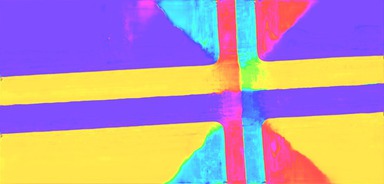} \\

	\includegraphics[width=0.155\linewidth]{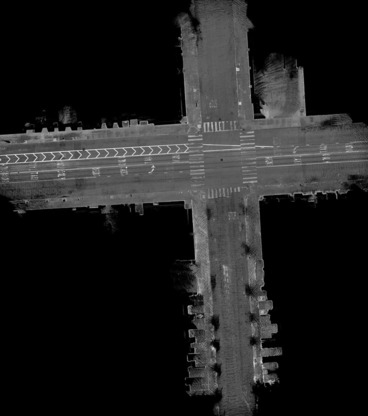}  & 
	\includegraphics[width=0.155\linewidth]{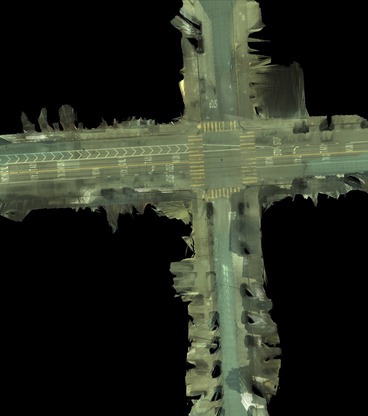}&  
	\includegraphics[width=0.155\linewidth]{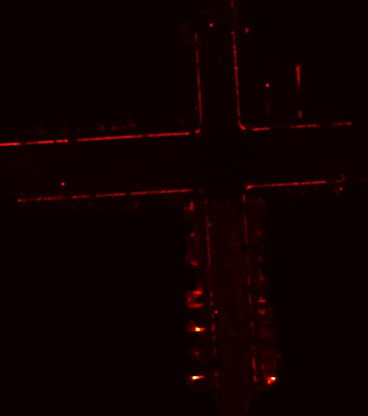} &
	\includegraphics[width=0.155\linewidth]{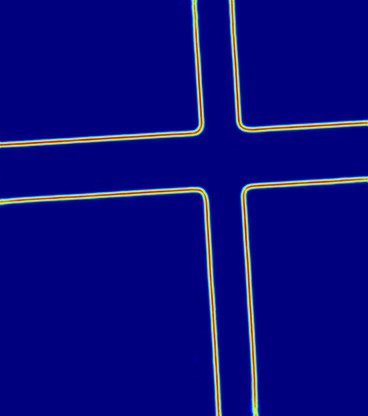}  & 
	\includegraphics[width=0.155\linewidth]{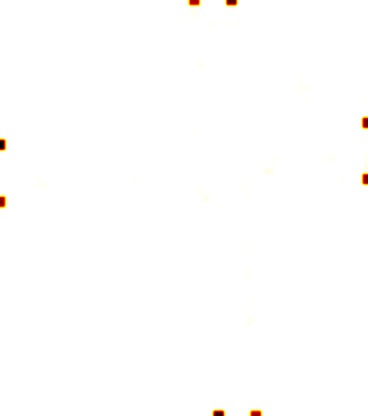}& 
	\includegraphics[width=0.155\linewidth]{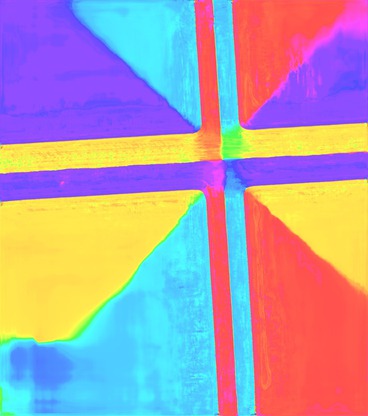} \\

	\includegraphics[width=0.155\linewidth]{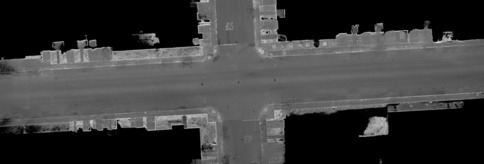}  & 
	\includegraphics[width=0.155\linewidth]{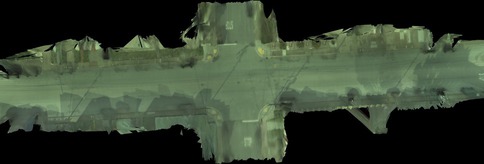}&  
	\includegraphics[width=0.155\linewidth]{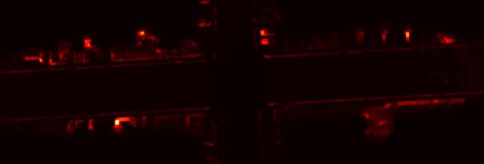} &
	\includegraphics[width=0.155\linewidth]{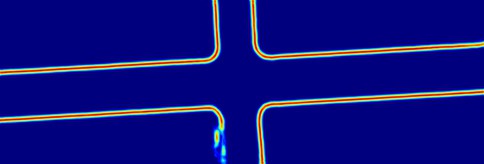}  & 
	\includegraphics[width=0.155\linewidth]{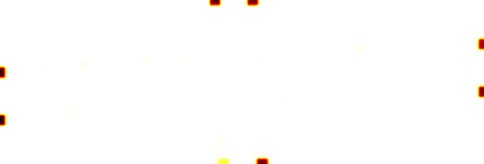}& 
	\includegraphics[width=0.155\linewidth]{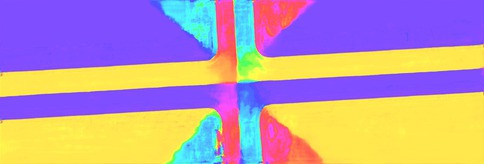} \\

	\includegraphics[width=0.155\linewidth]{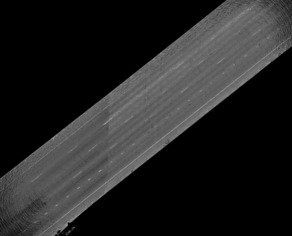}  & 
	\includegraphics[width=0.155\linewidth]{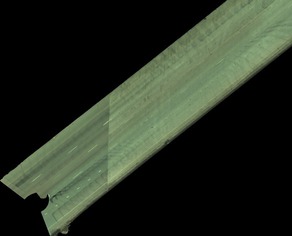}&  
	\includegraphics[width=0.155\linewidth]{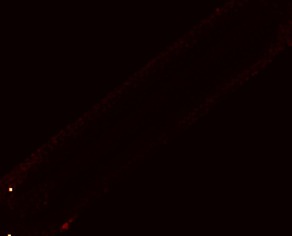} &
	\includegraphics[width=0.155\linewidth]{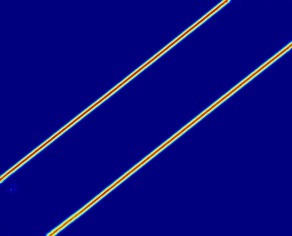}  & 
	\includegraphics[width=0.155\linewidth]{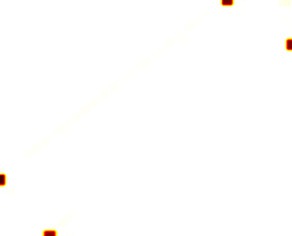}& 
	\includegraphics[width=0.155\linewidth]{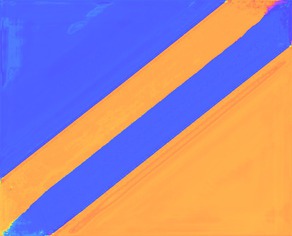} \\

	\raisebox{2px}{{(Lidar)}} &	
	\raisebox{2px}{{(Camera)}} &
	\raisebox{2px}{{(Elevation Gradient)}} &
	\raisebox{2px}{{(Detection Map)}} &
	\raisebox{2px}{{(Endpoints)}} &
	\raisebox{2px}{{(Direction Map)}} 

	\end{array}
	\]
	
	\caption{Deep Features: Columns \textbf{(1-3)} correspond to the  inputs and columns \textbf{(4-6)} correspond to the deep feature maps. The direction map shown here as a flow field \cite{color_code}.}
	\label{fig:feat3}
\end{figure*}

\begin{figure*}[h]
    \centering
	\[\arraycolsep=1.0pt
	\begin{array}{cccccc}
	
	\includegraphics[width=0.155\linewidth]{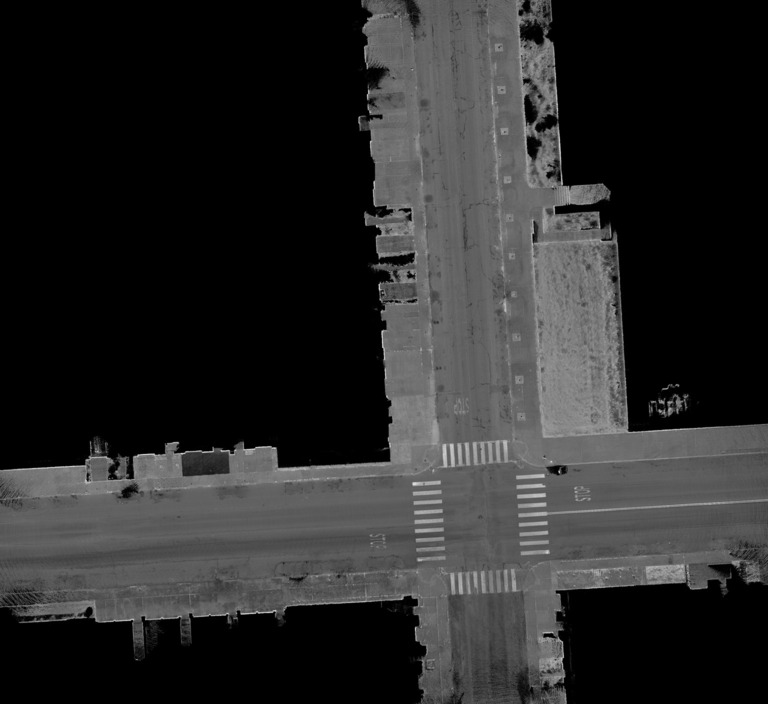}  & 
	\includegraphics[width=0.155\linewidth]{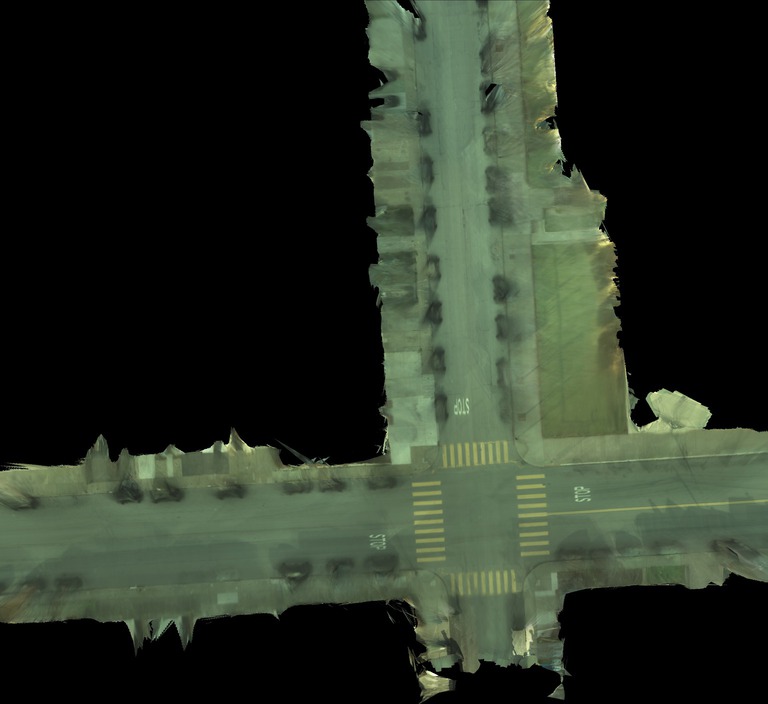}&  
	\includegraphics[width=0.155\linewidth]{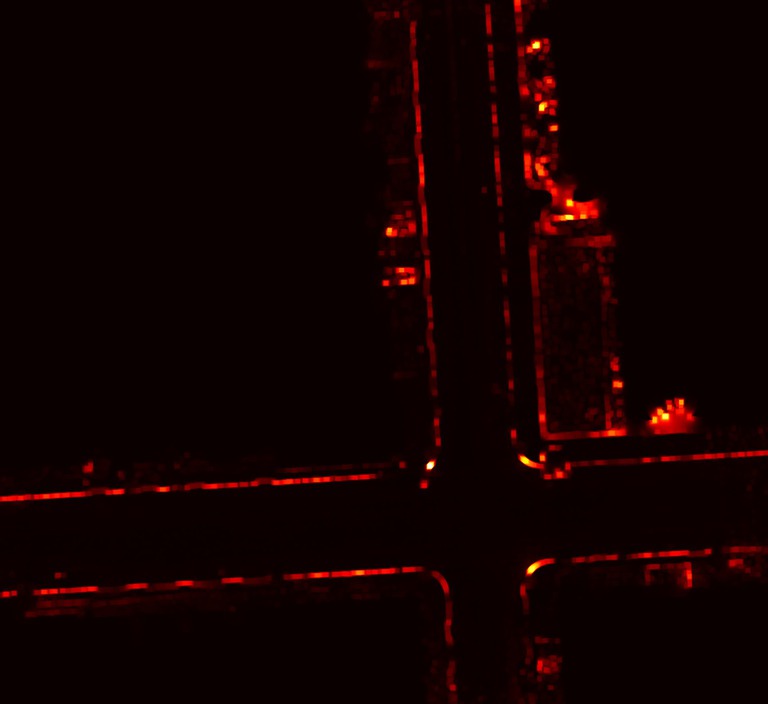} &
	\includegraphics[width=0.155\linewidth]{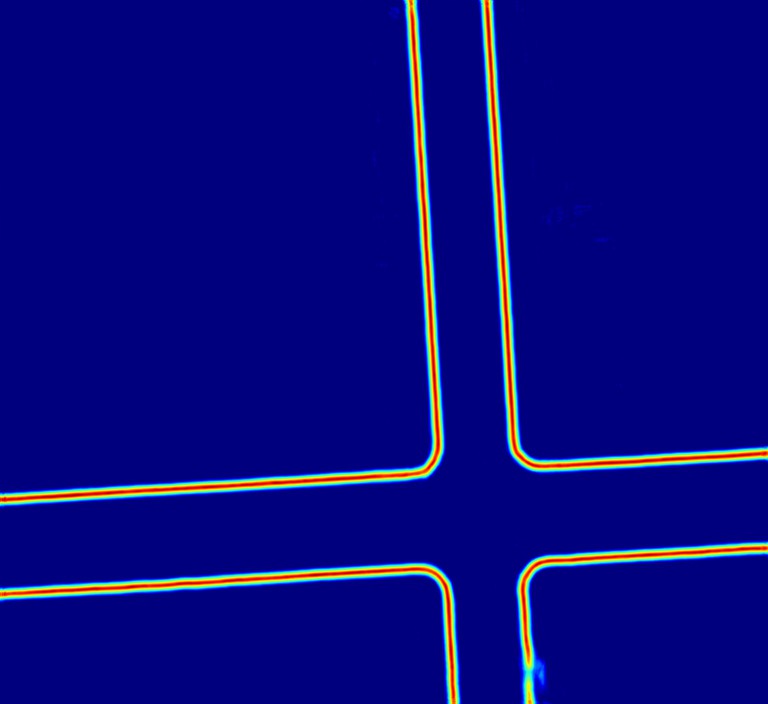}  & 
	\includegraphics[width=0.155\linewidth]{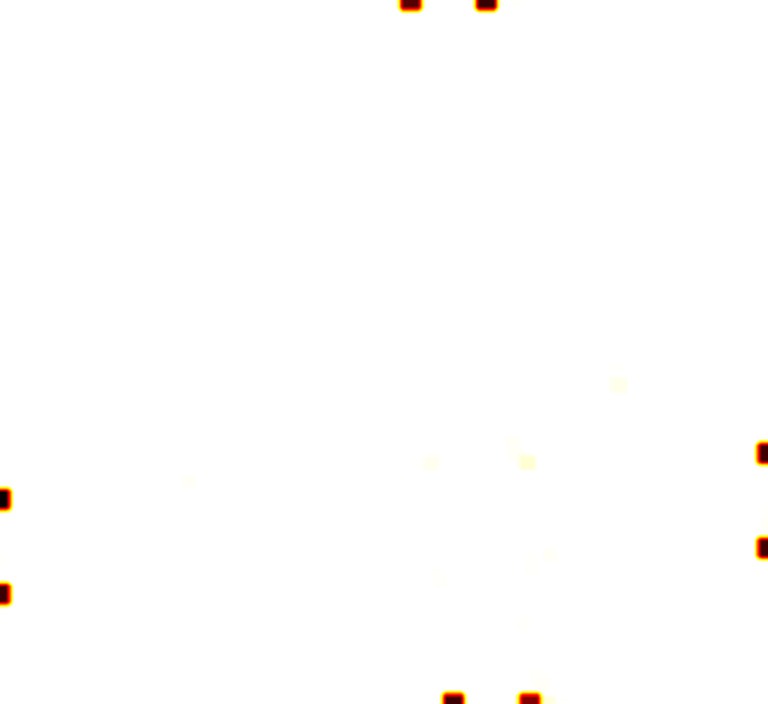}& 
	\includegraphics[width=0.155\linewidth]{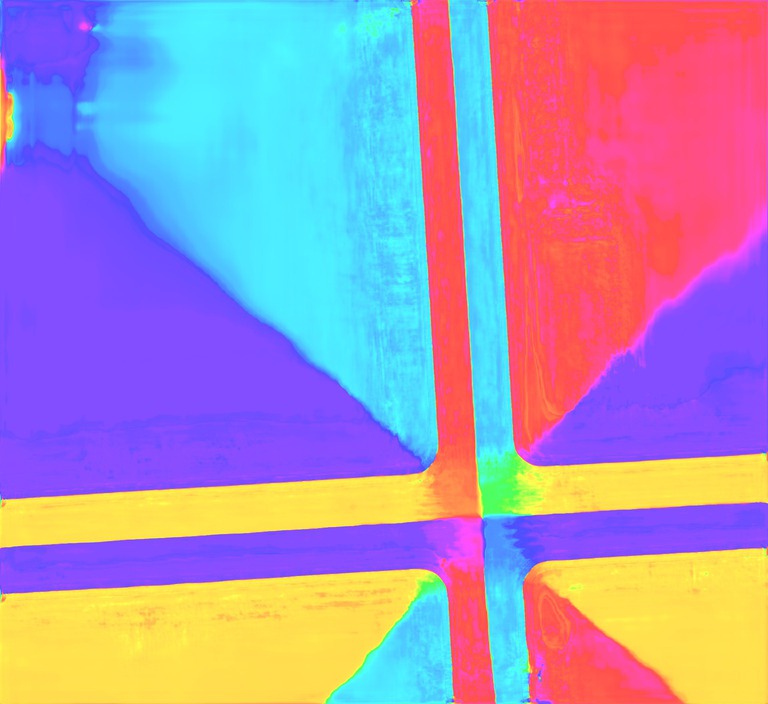} \\

	\includegraphics[width=0.155\linewidth]{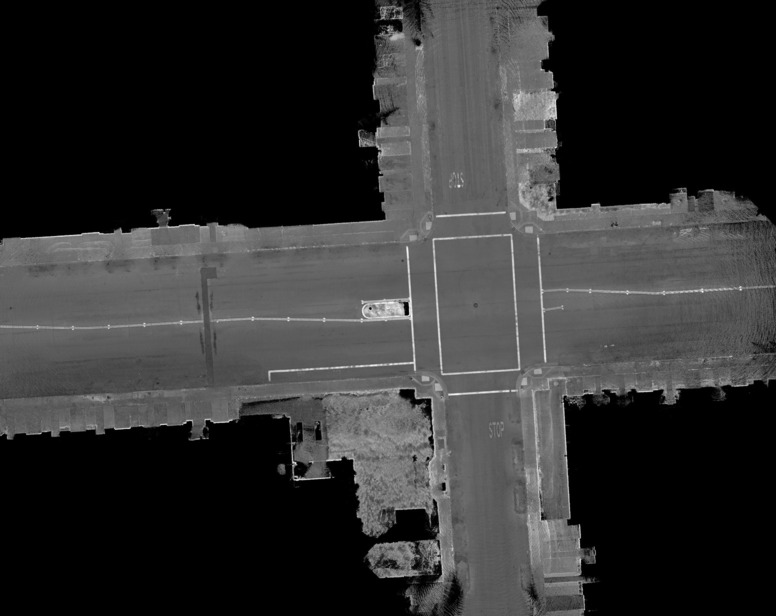}  & 
	\includegraphics[width=0.155\linewidth]{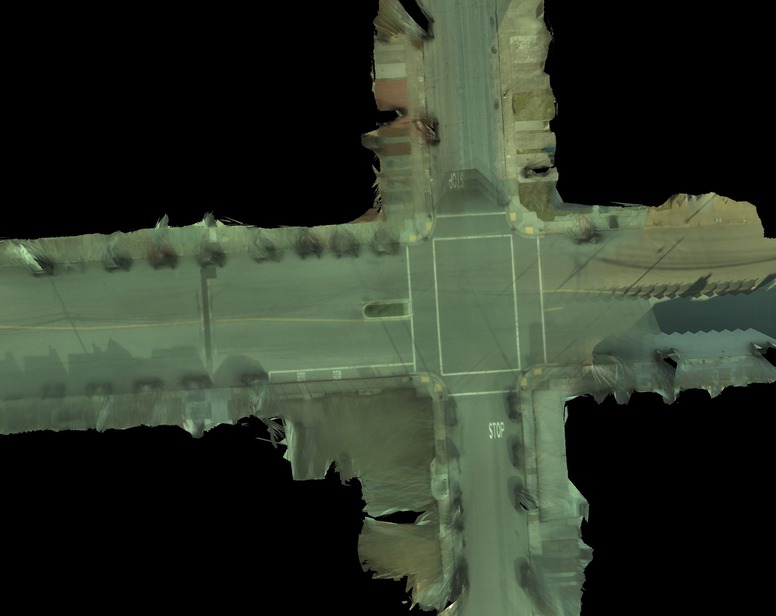}&  
	\includegraphics[width=0.155\linewidth]{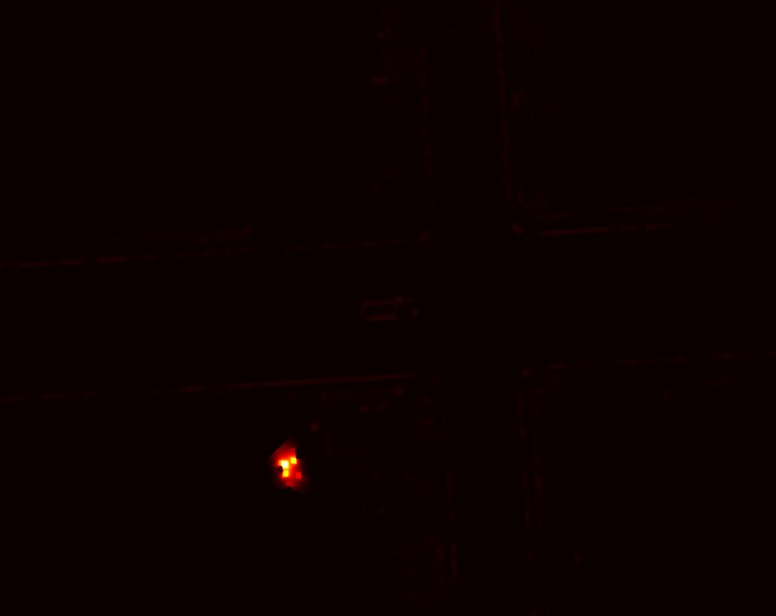} &
	\includegraphics[width=0.155\linewidth]{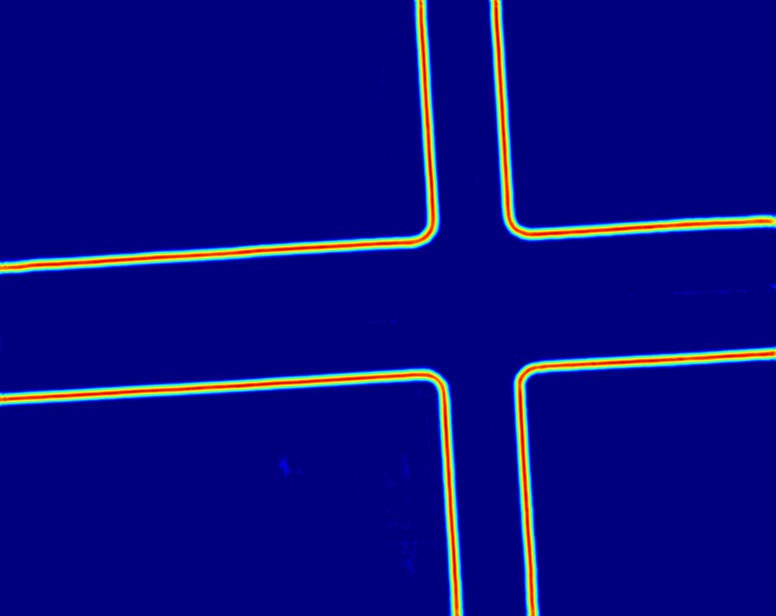}  & 
	\includegraphics[width=0.155\linewidth]{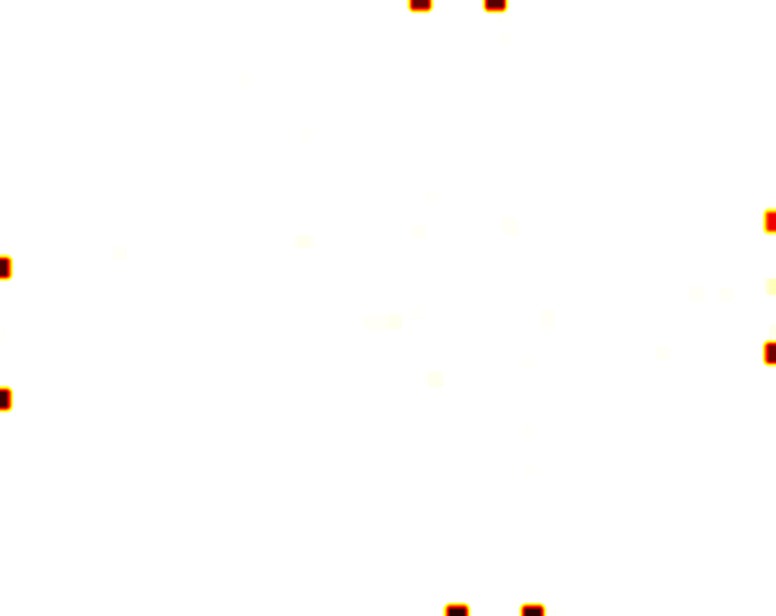}& 
	\includegraphics[width=0.155\linewidth]{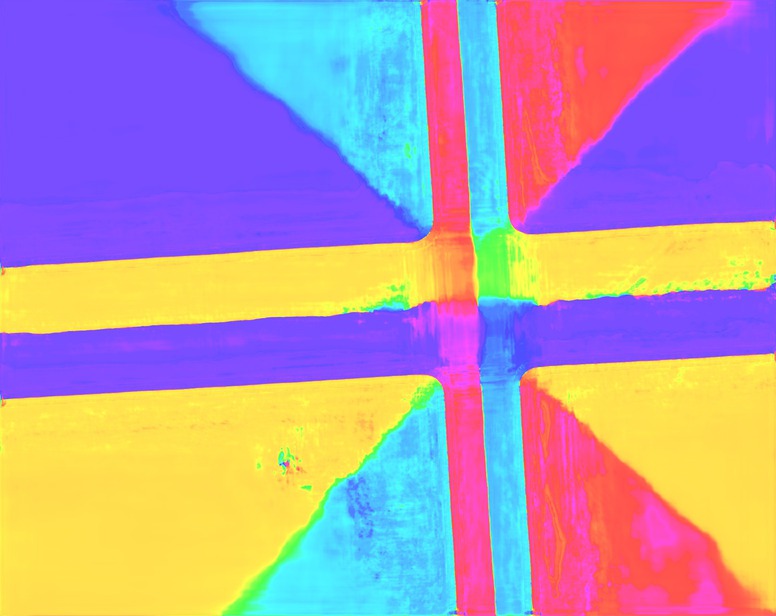} \\

	\includegraphics[width=0.155\linewidth]{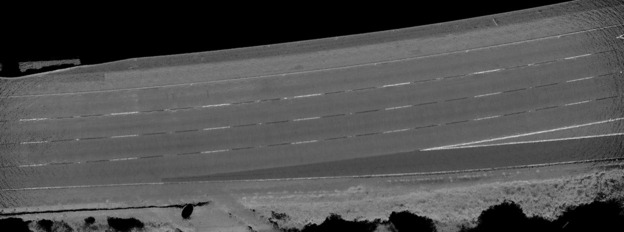}  & 
	\includegraphics[width=0.155\linewidth]{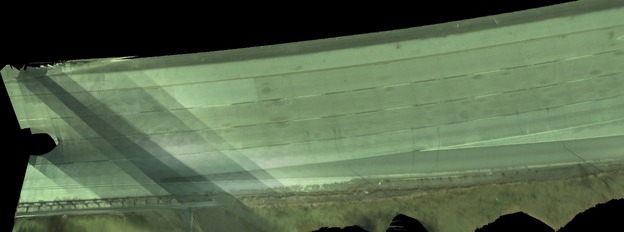}&  
	\includegraphics[width=0.155\linewidth]{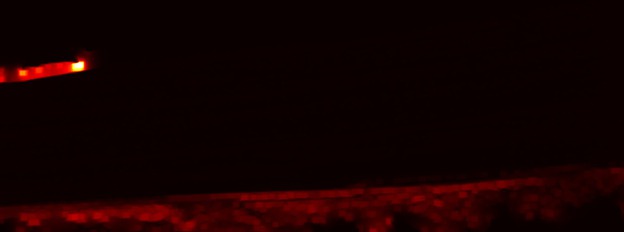} &
	\includegraphics[width=0.155\linewidth]{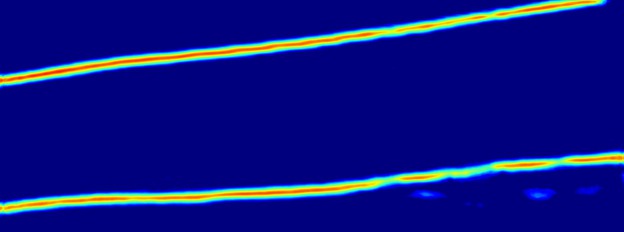}  & 
	\includegraphics[width=0.155\linewidth]{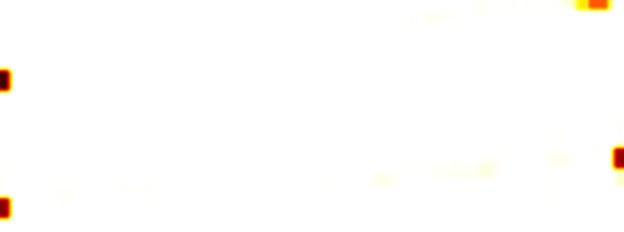}& 
	\includegraphics[width=0.155\linewidth]{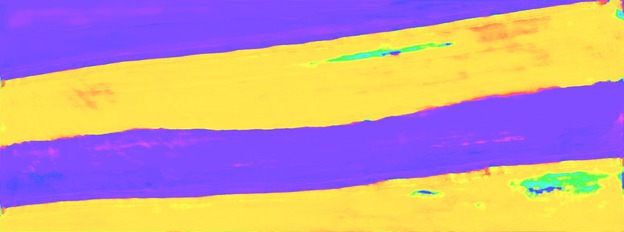} \\

	\includegraphics[width=0.155\linewidth]{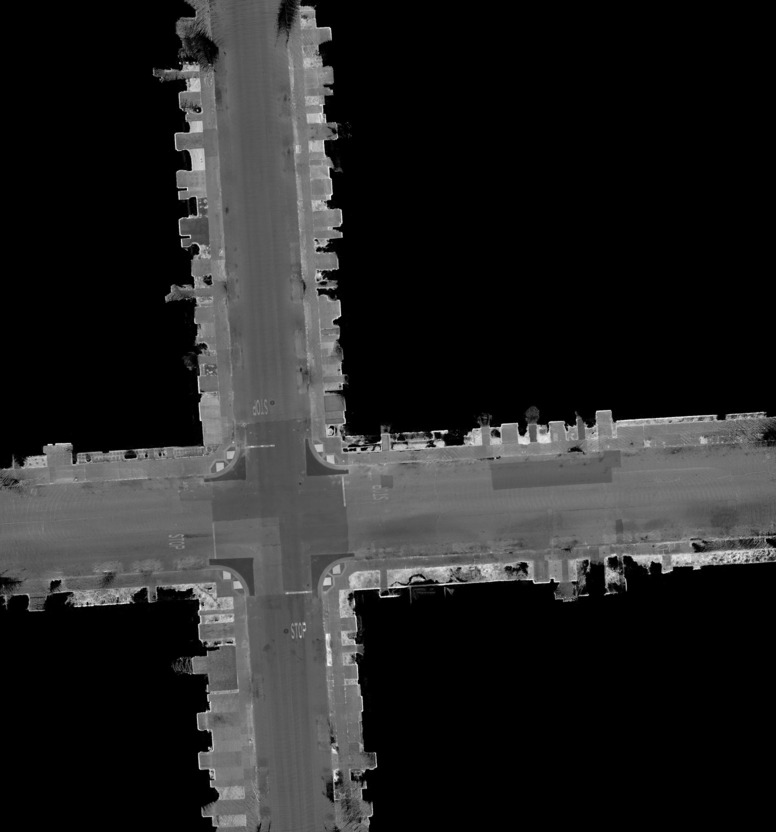}  & 
	\includegraphics[width=0.155\linewidth]{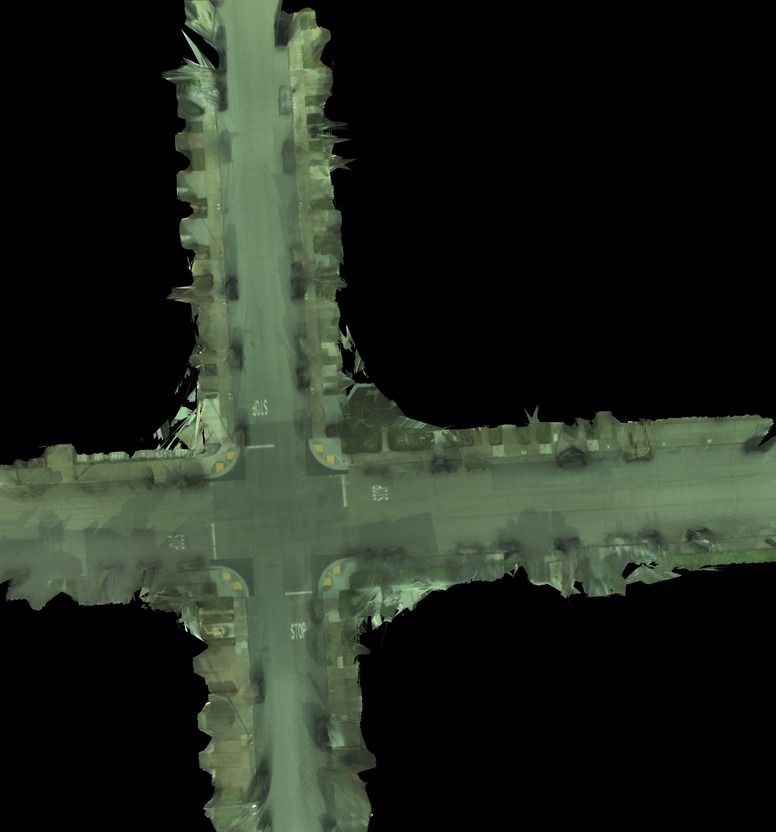}&  
	\includegraphics[width=0.155\linewidth]{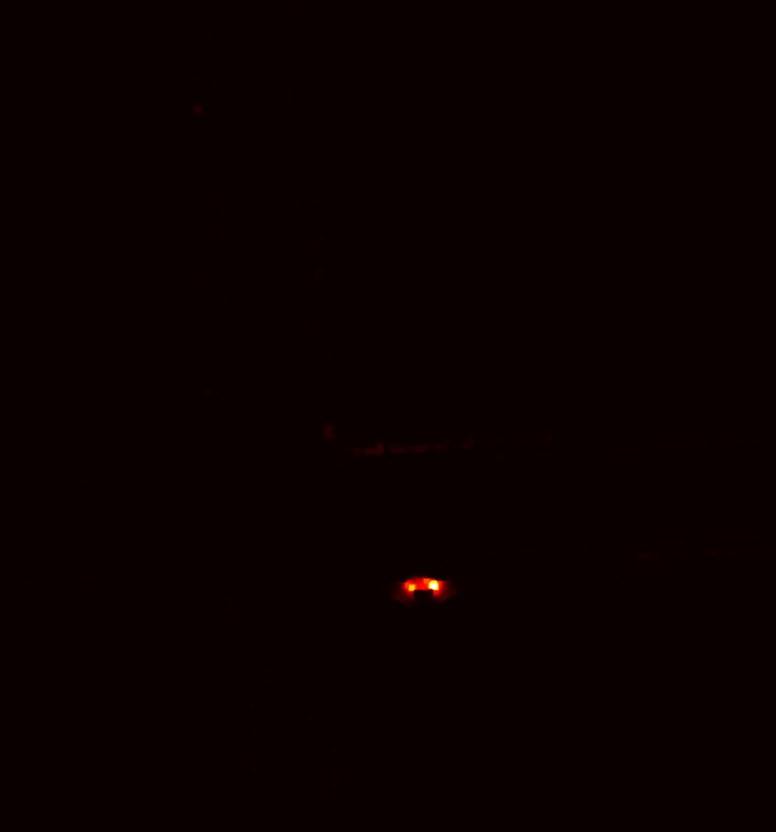} &
	\includegraphics[width=0.155\linewidth]{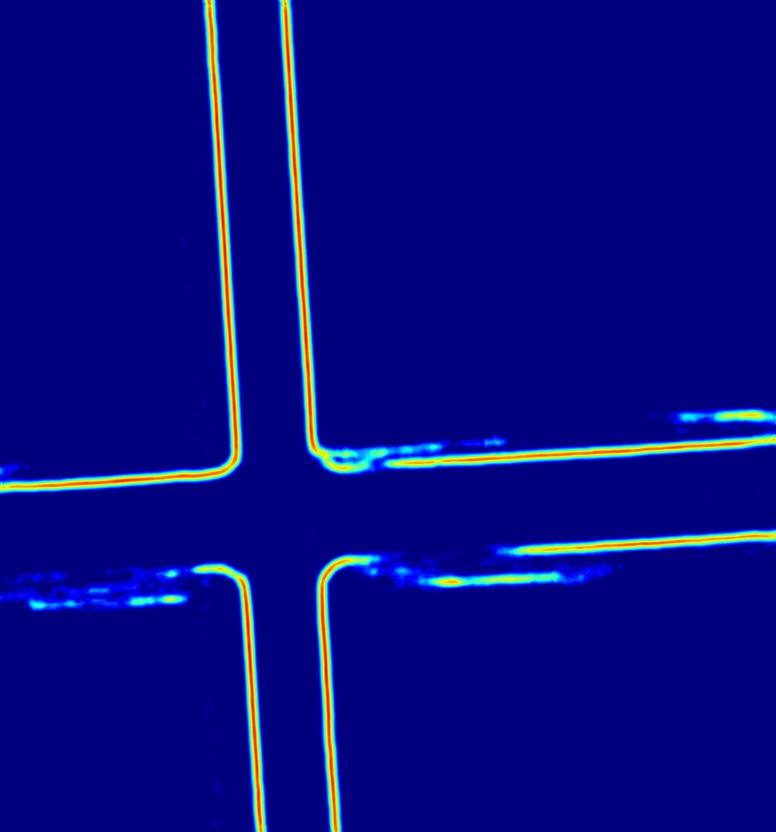}  & 
	\includegraphics[width=0.155\linewidth]{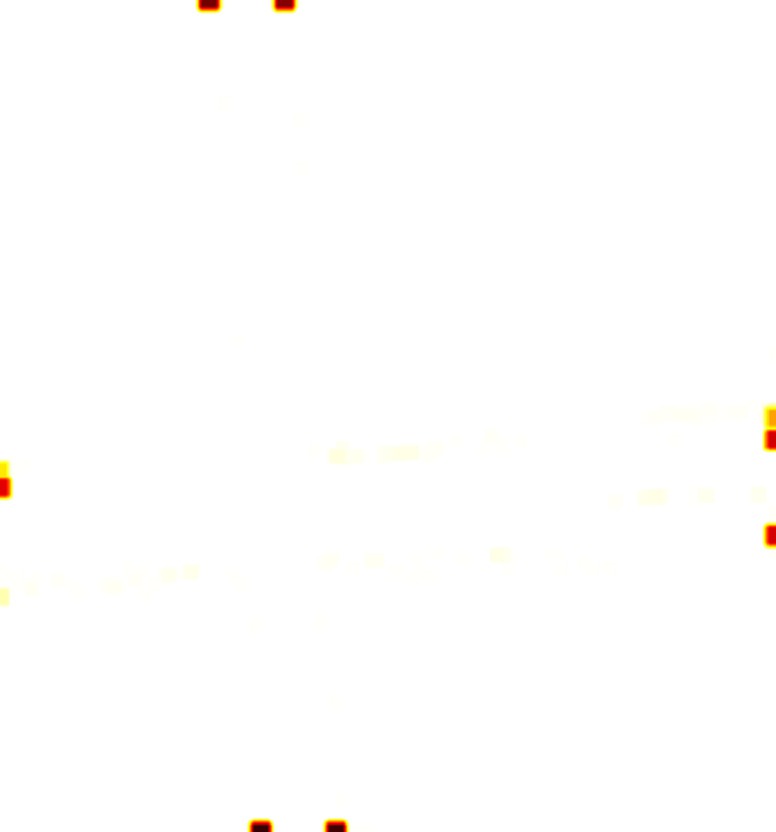}& 
	\includegraphics[width=0.155\linewidth]{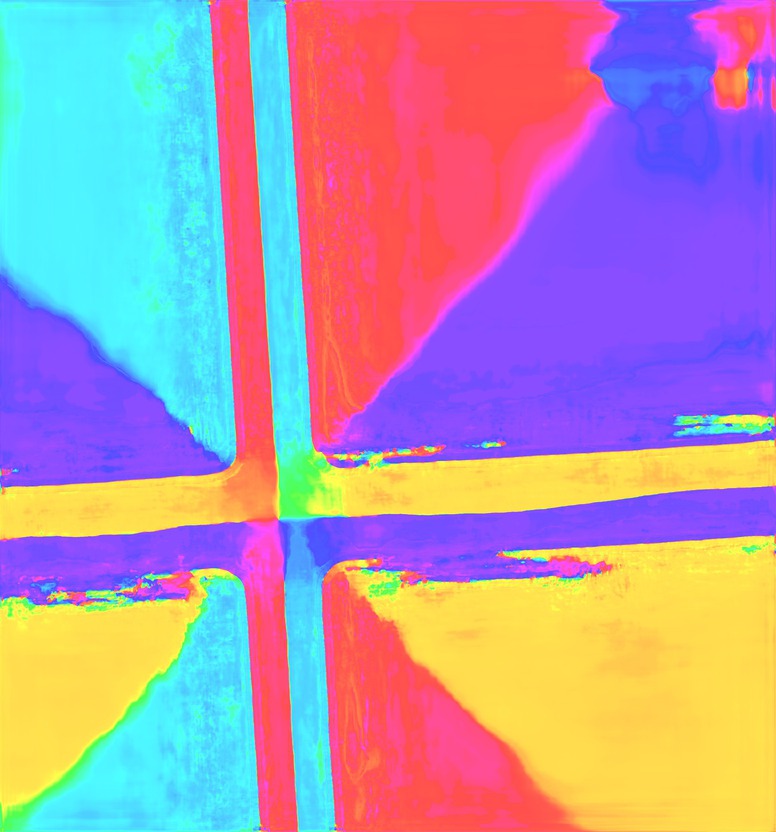} \\

	\includegraphics[width=0.155\linewidth]{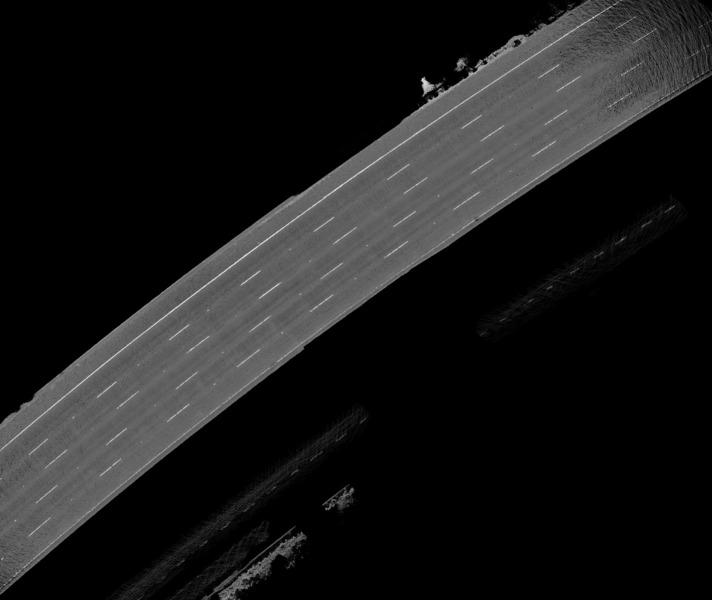}  & 
	\includegraphics[width=0.155\linewidth]{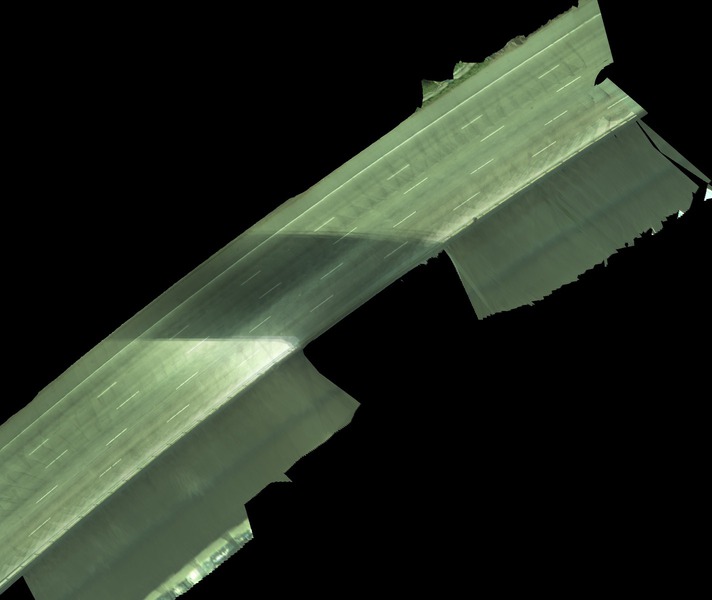}&  
	\includegraphics[width=0.155\linewidth]{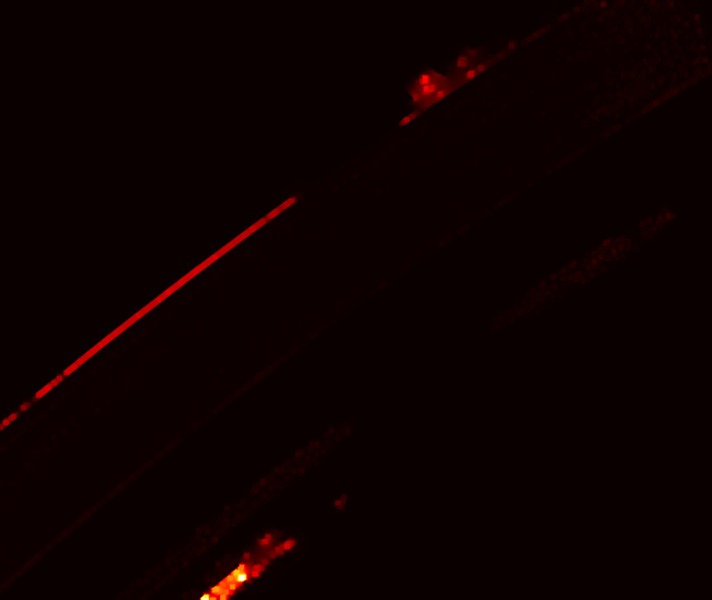} &
	\includegraphics[width=0.155\linewidth]{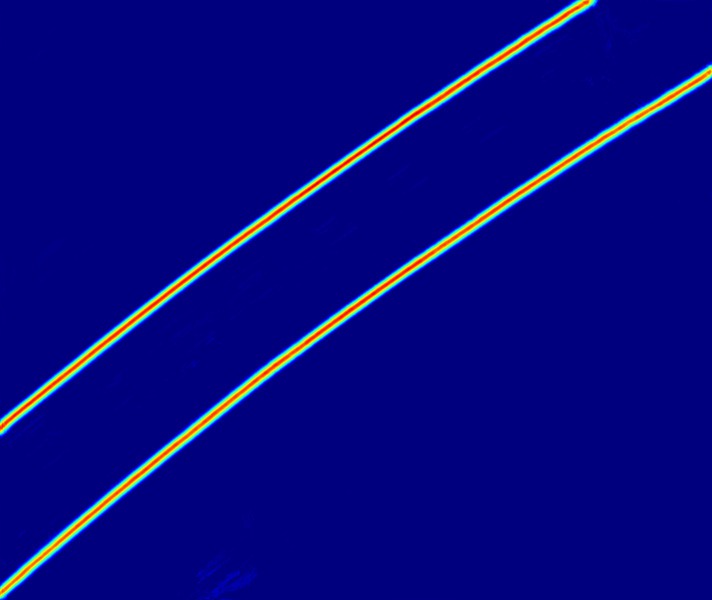}  & 
	\includegraphics[width=0.155\linewidth]{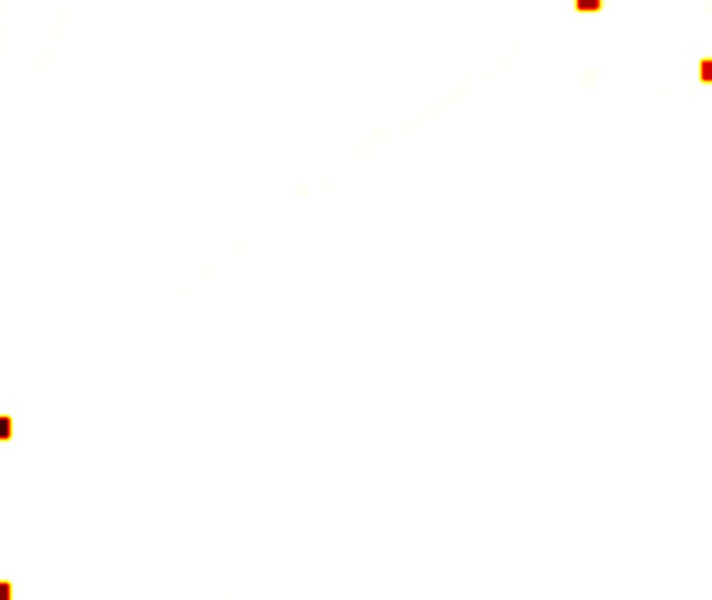}& 
	\includegraphics[width=0.155\linewidth]{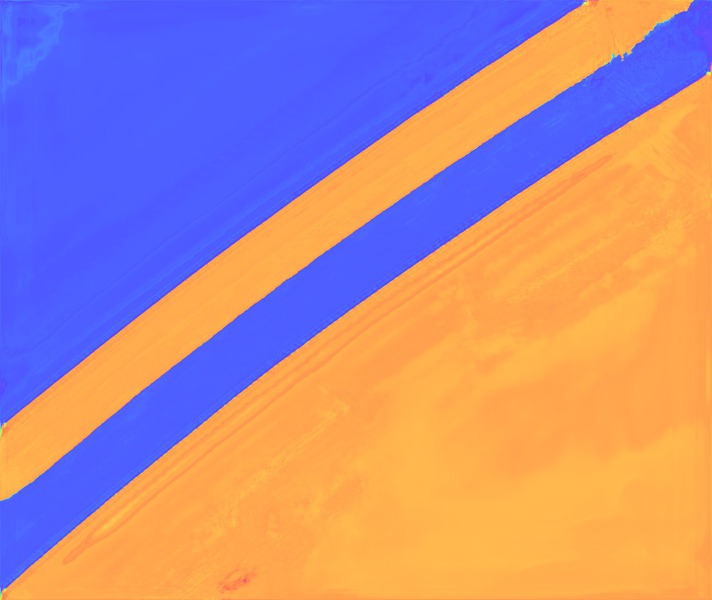} \\

	\includegraphics[width=0.155\linewidth]{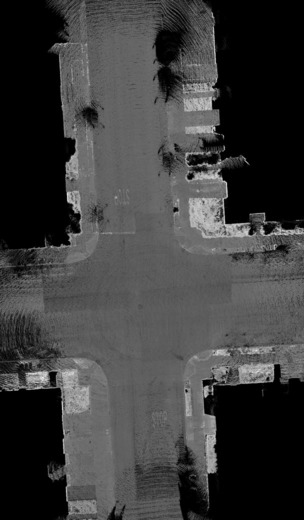}  & 
	\includegraphics[width=0.155\linewidth]{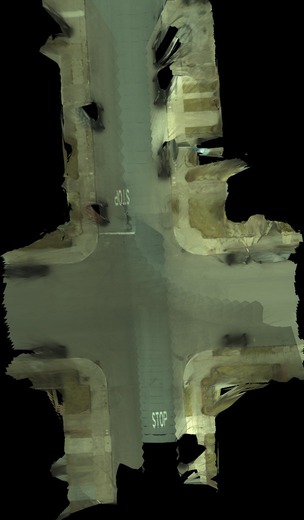}&  
	\includegraphics[width=0.155\linewidth]{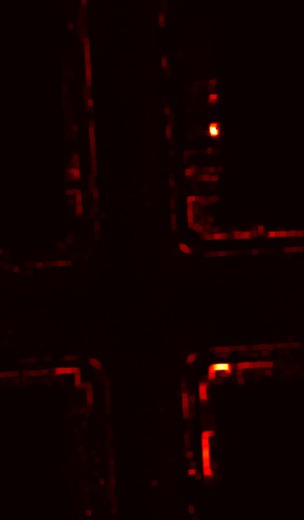} &
	\includegraphics[width=0.155\linewidth]{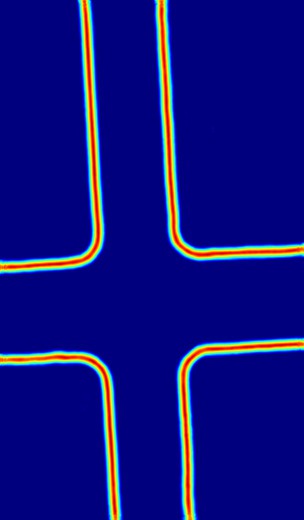}  & 
	\includegraphics[width=0.155\linewidth]{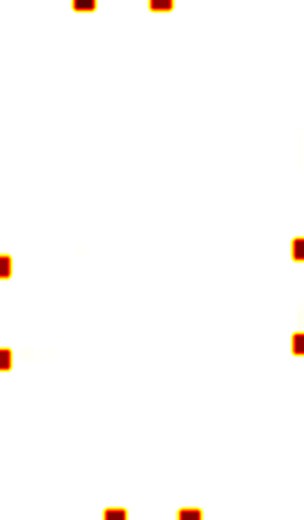}& 
	\includegraphics[width=0.155\linewidth]{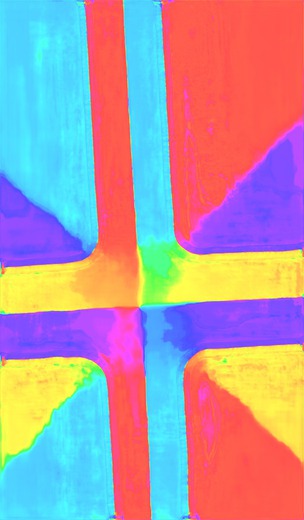} \\

	\includegraphics[width=0.155\linewidth]{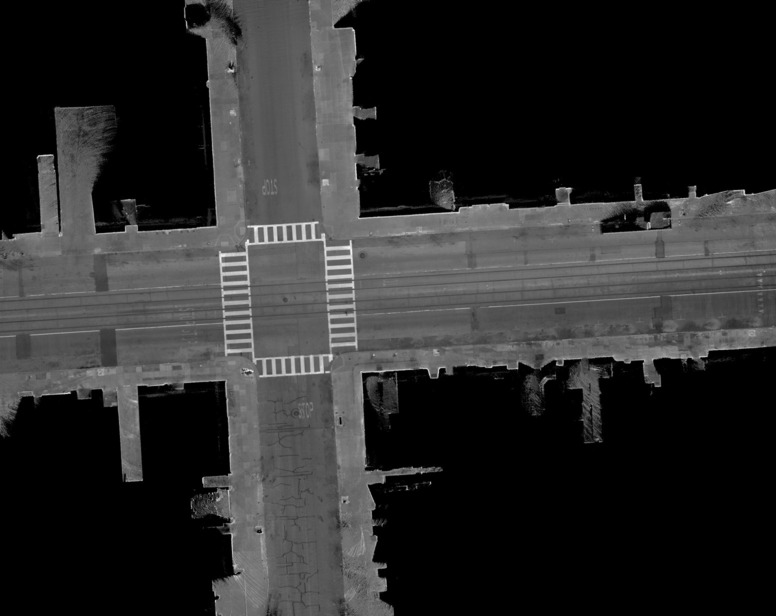}  & 
	\includegraphics[width=0.155\linewidth]{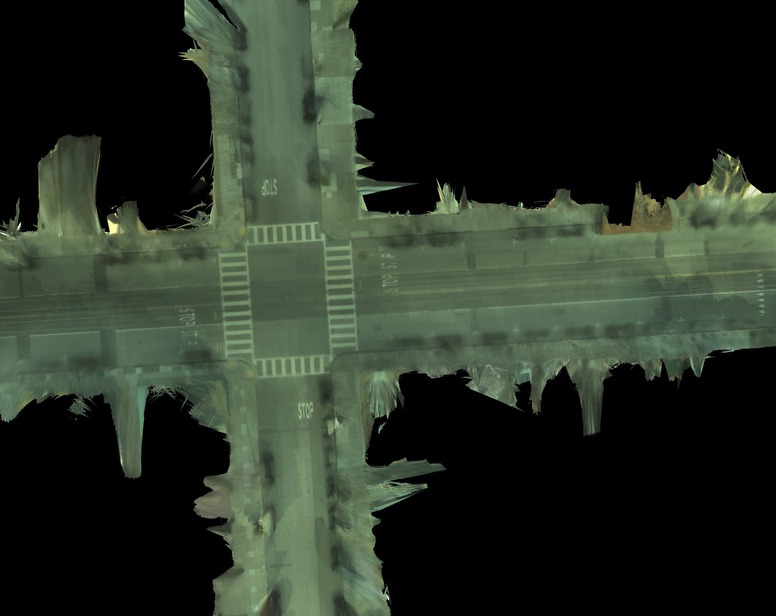}&  
	\includegraphics[width=0.155\linewidth]{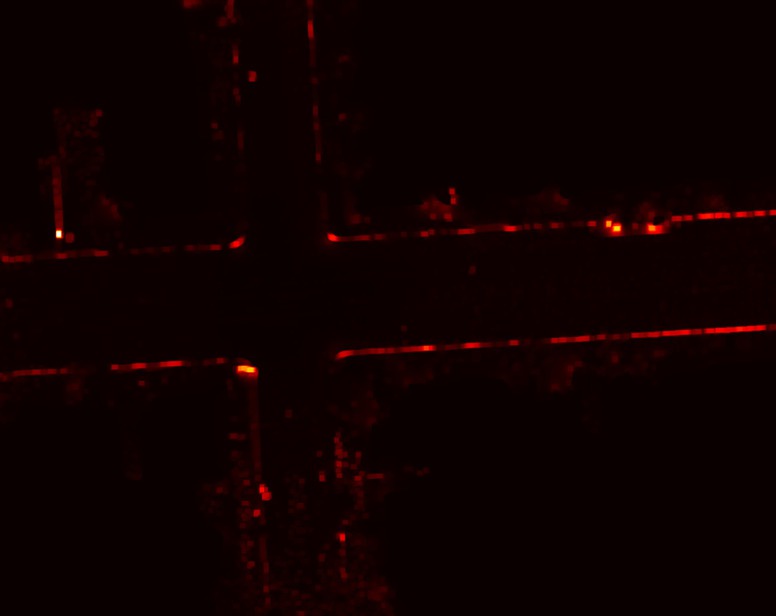} &
	\includegraphics[width=0.155\linewidth]{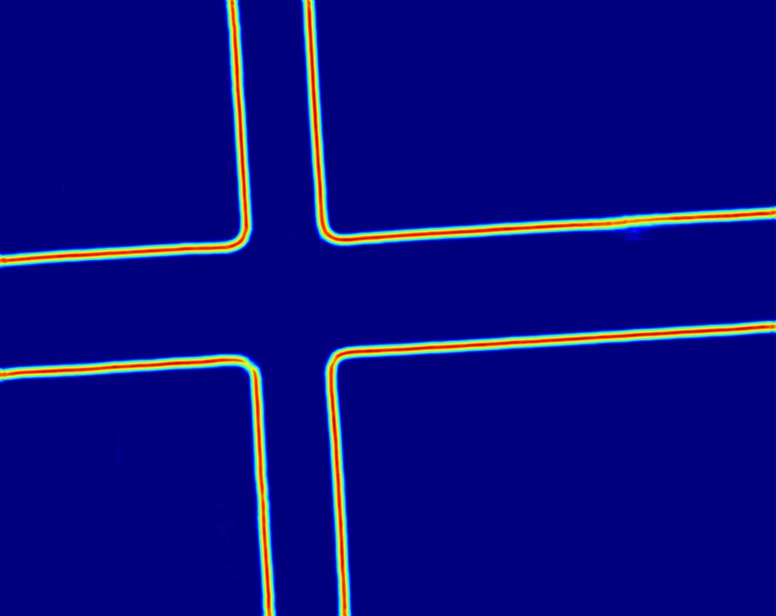}  & 
	\includegraphics[width=0.155\linewidth]{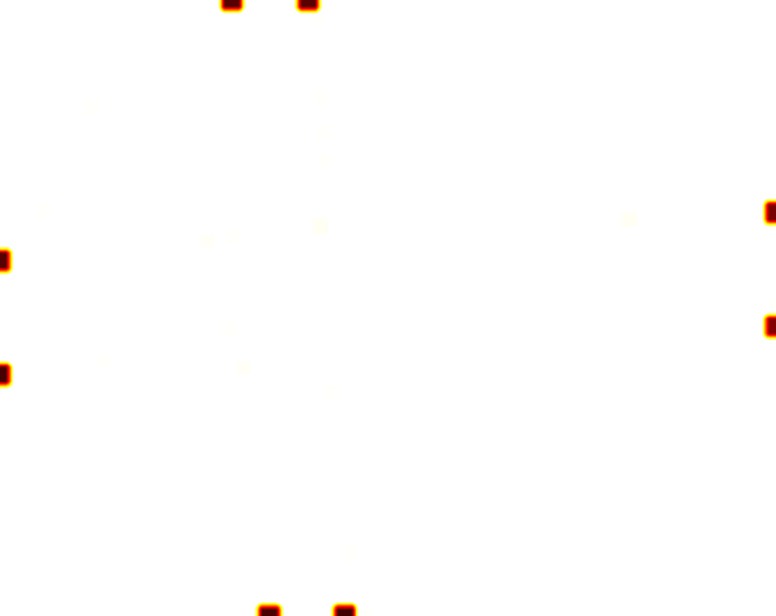}& 
	\includegraphics[width=0.155\linewidth]{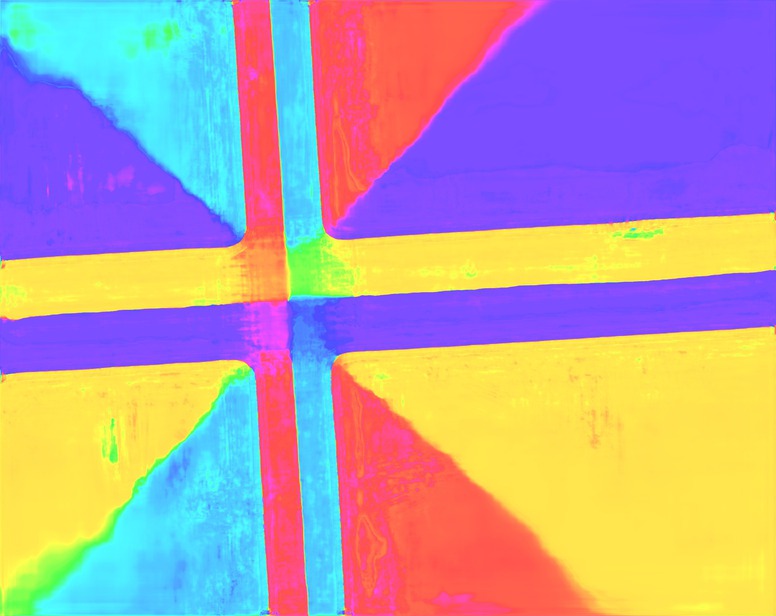} \\

	\raisebox{2px}{{(Lidar)}} &	
	\raisebox{2px}{{(Camera)}} &
	\raisebox{2px}{{(Elevation Gradient)}} &
	\raisebox{2px}{{(Detection Map)}} &
	\raisebox{2px}{{(Endpoints)}} &
	\raisebox{2px}{{(Direction Map)}}

	\end{array}
	\]

	\caption{Deep Features: Columns \textbf{(1-3)} correspond to the  inputs and columns \textbf{(4-6)} correspond to the deep feature maps. The direction map shown here as a flow field \cite{color_code}.}
	\label{fig:feat4}
\end{figure*}

\begin{figure*}[h]
	\[\arraycolsep=2pt
	\begin{array}{cc}

	\includegraphics[width=0.38\linewidth]{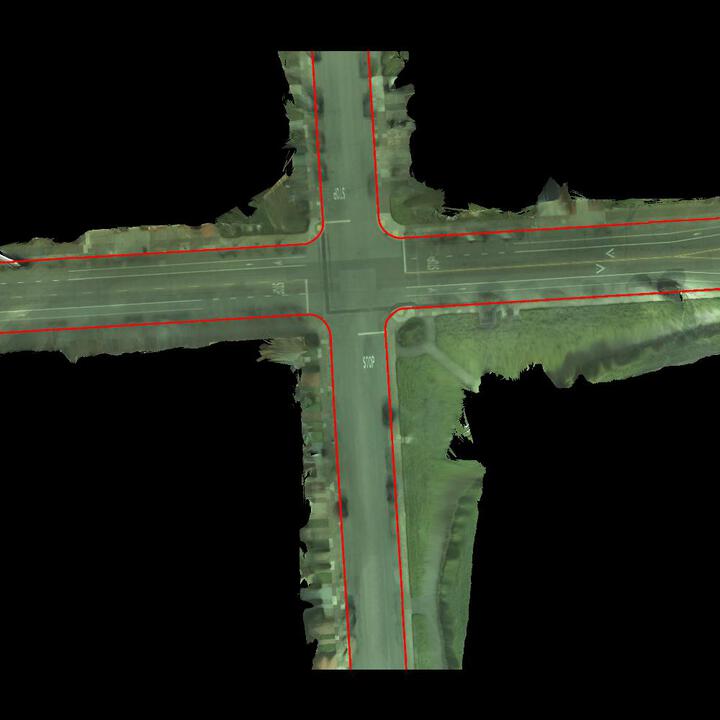}& 
	\includegraphics[width=0.38\linewidth]{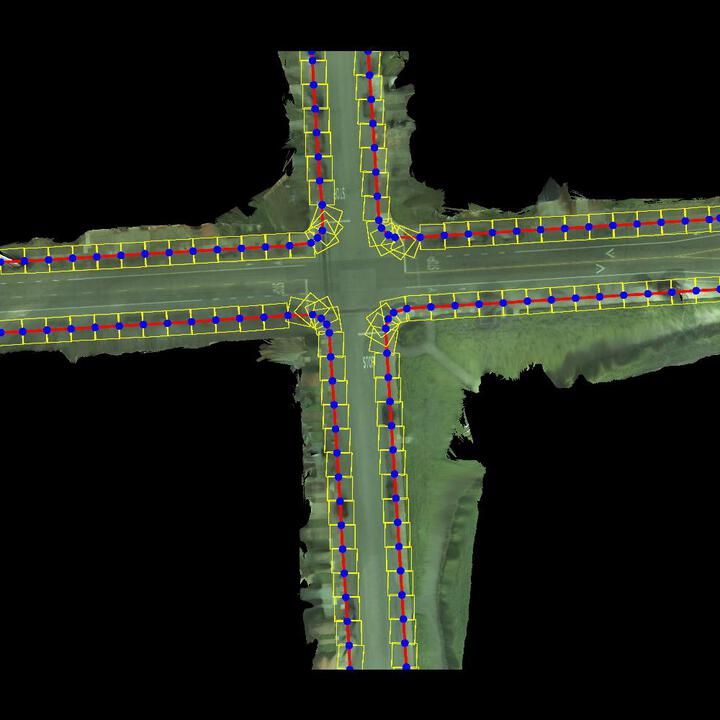} \\
	
	\includegraphics[width=0.38\linewidth]{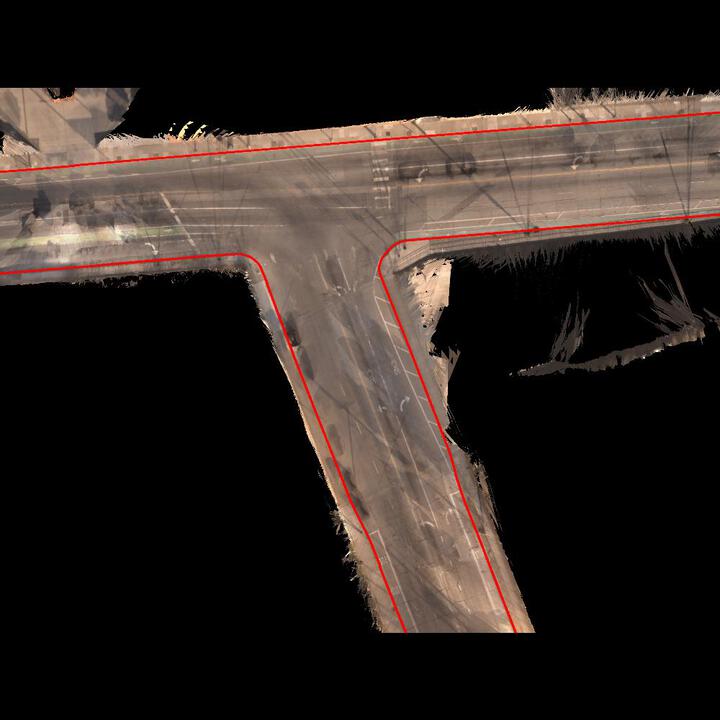}& 
	\includegraphics[width=0.38\linewidth]{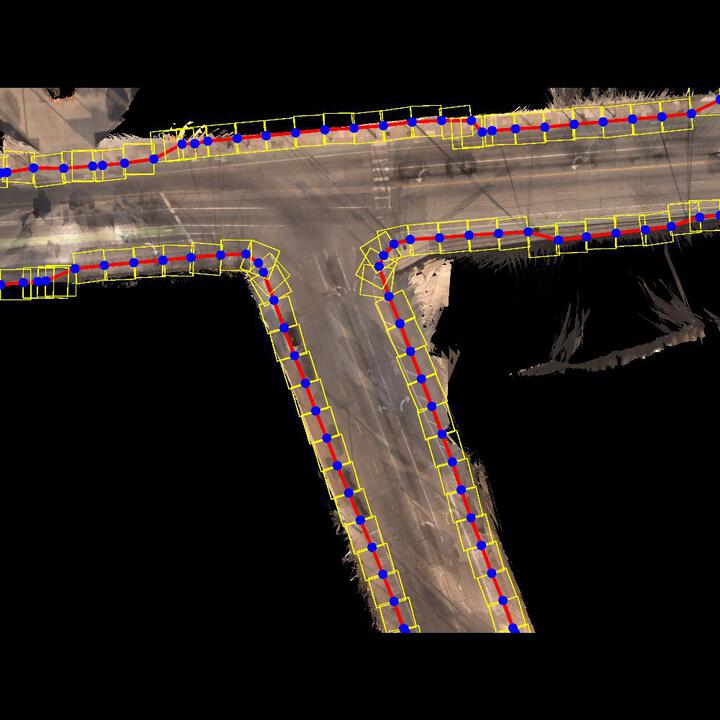} \\
	
	\includegraphics[width=0.38\linewidth]{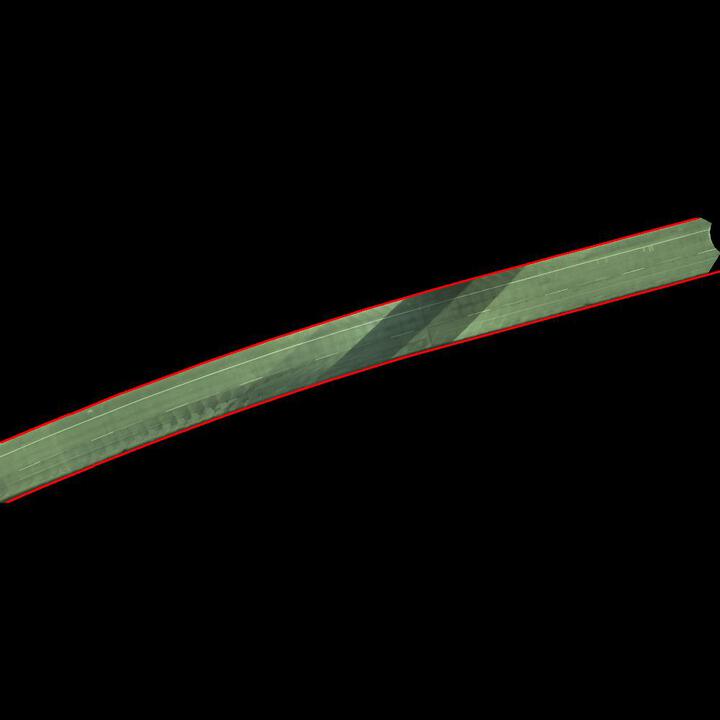}& 
	\includegraphics[width=0.38\linewidth]{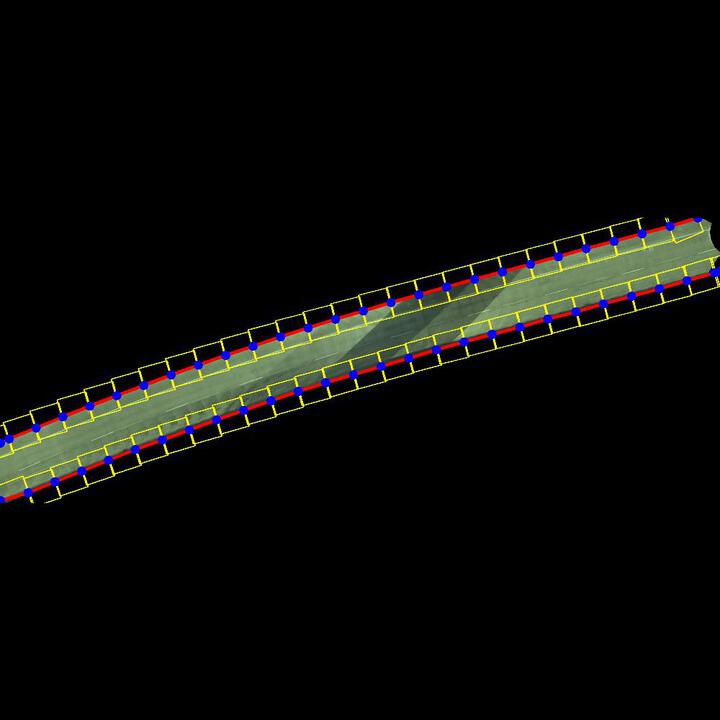} \\

	\raisebox{2px}{{(Ground Truth)}} &
	\raisebox{2px}{{(Prediction)}} 
	
	\end{array}
	\]
\vspace{-5mm}	
	\caption{Results: Column \textbf{1} corresponds to the ground truth boundaries and column \textbf{2} are our predicted boundaries.}
	\label{fig:results_1}
\end{figure*}

\begin{figure*}[h]
	\[\arraycolsep=2pt
	\begin{array}{cc}

	\includegraphics[width=0.38\linewidth]{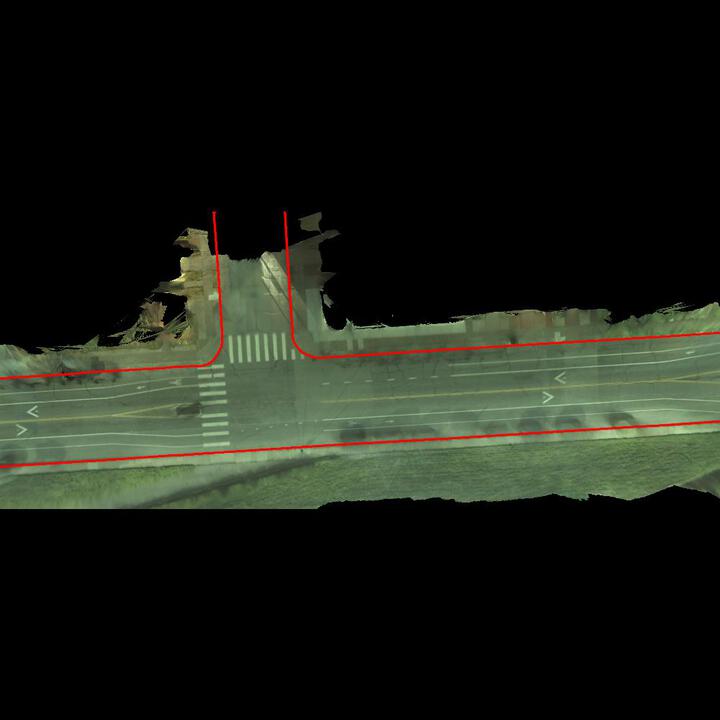}& 
	\includegraphics[width=0.38\linewidth]{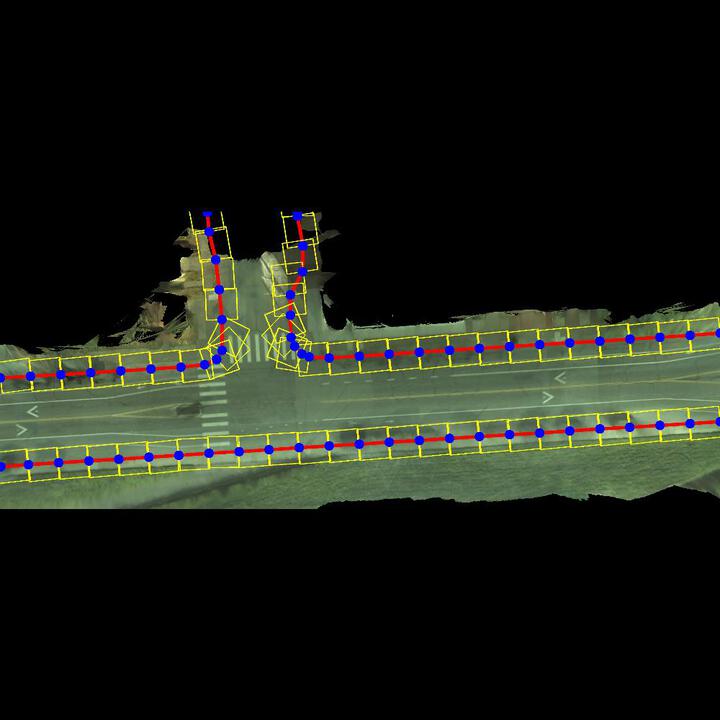} \\
	
	\includegraphics[width=0.38\linewidth]{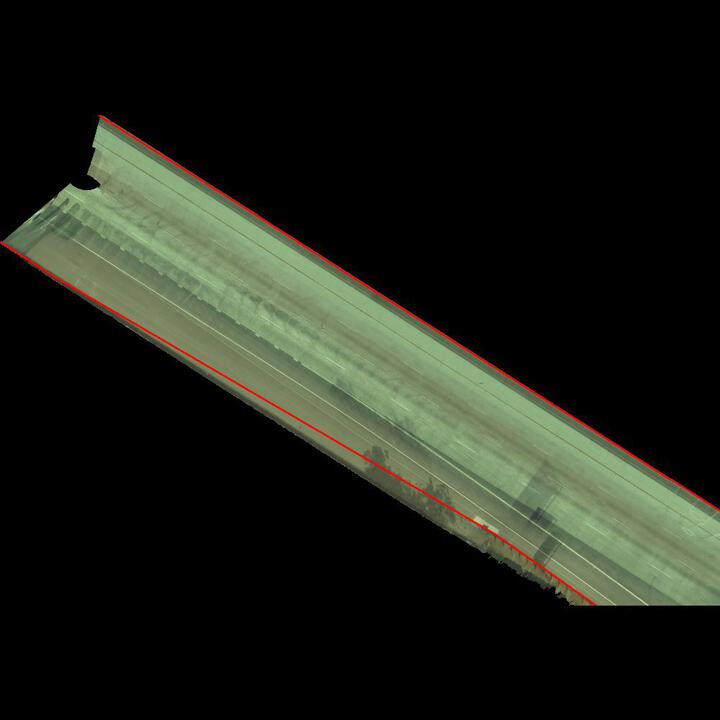}& 
	\includegraphics[width=0.38\linewidth]{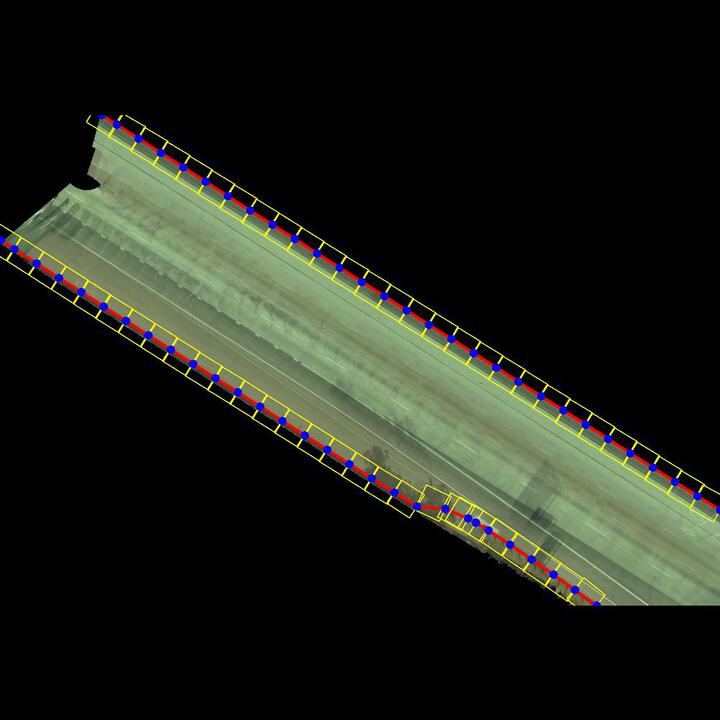} \\
	
	\includegraphics[width=0.38\linewidth]{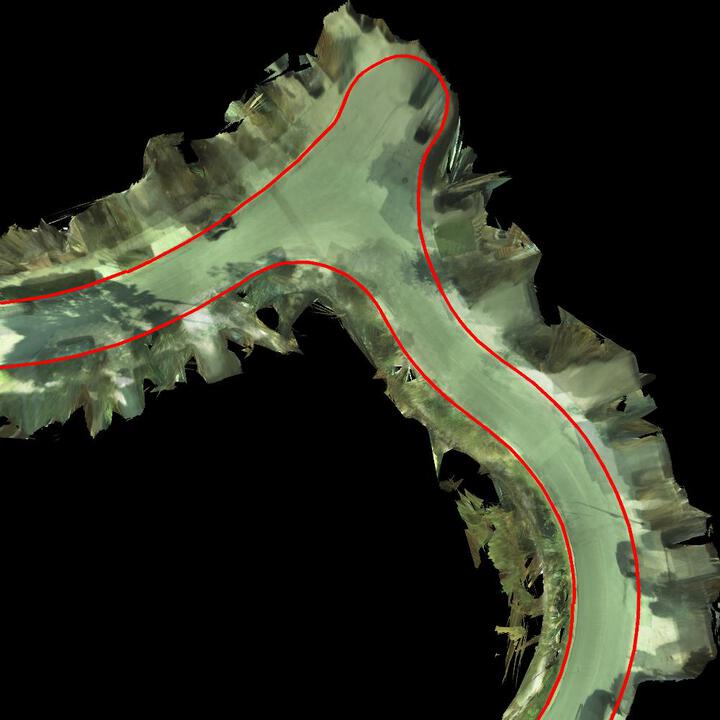}& 
	\includegraphics[width=0.38\linewidth]{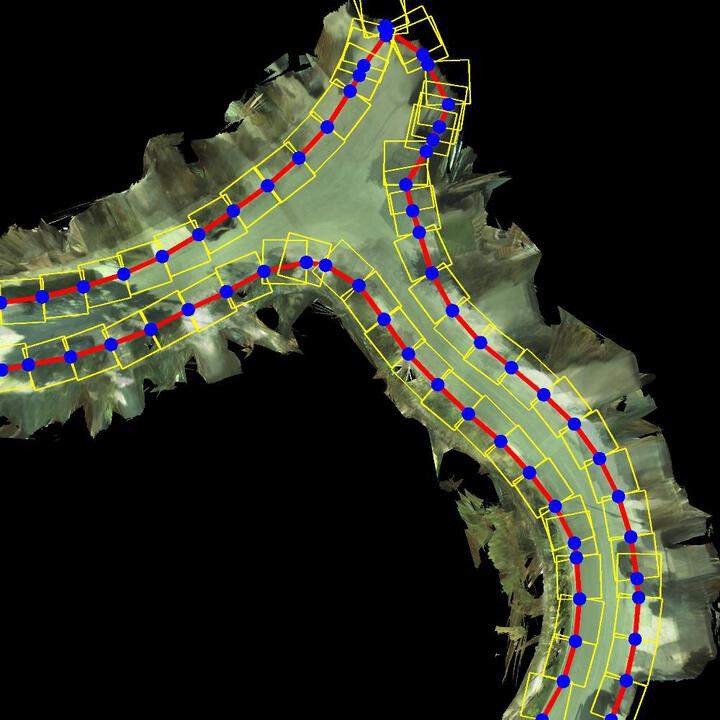} \\	

	\raisebox{2px}{{(Ground Truth)}} &
	\raisebox{2px}{{(Prediction)}} 
	
	\end{array}
	\]
	
	\caption{Results: Column \textbf{1} corresponds to the ground truth boundaries and column \textbf{2} are our predicted boundaries.}
	\label{fig:results_2}
\end{figure*}

\begin{figure*}[h]
	\[\arraycolsep=2pt
	\begin{array}{cc}

	\includegraphics[width=0.38\linewidth]{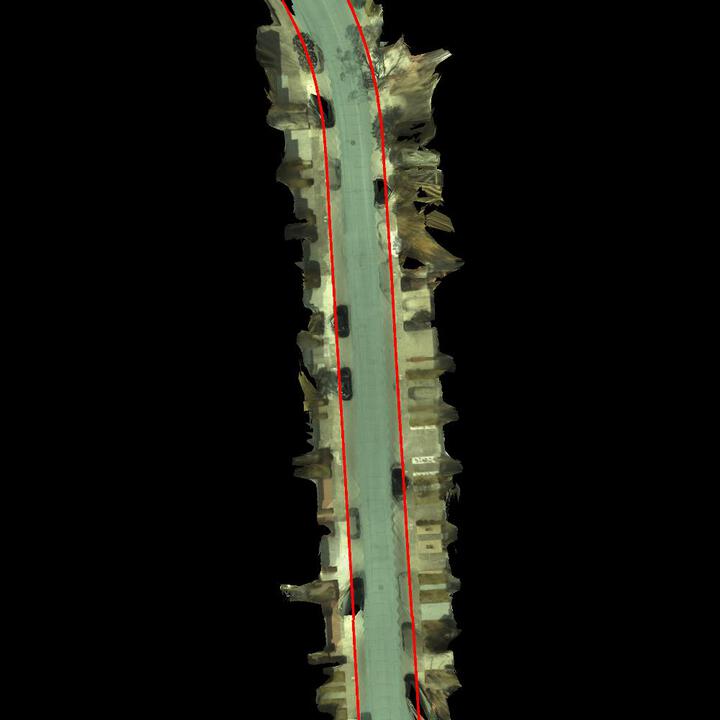}& 
	\includegraphics[width=0.38\linewidth]{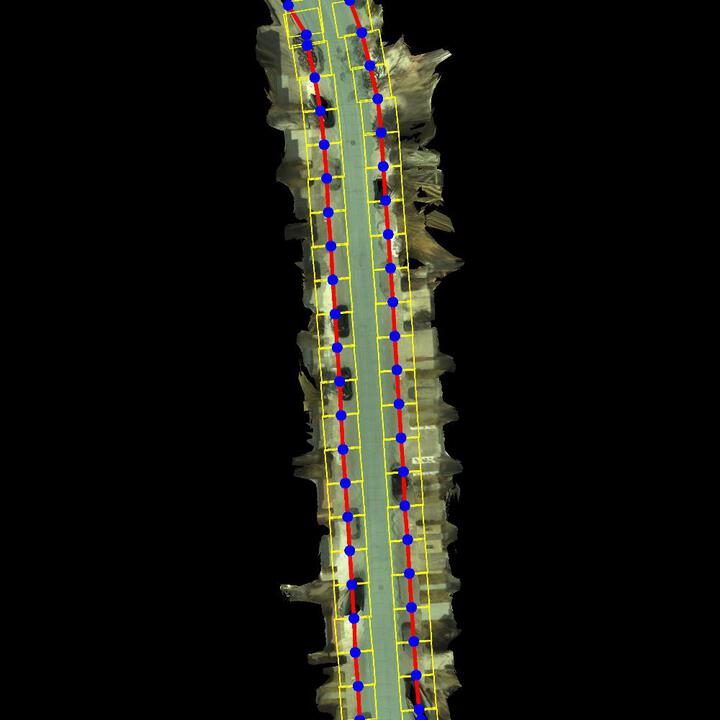} \\
	
	\includegraphics[width=0.38\linewidth]{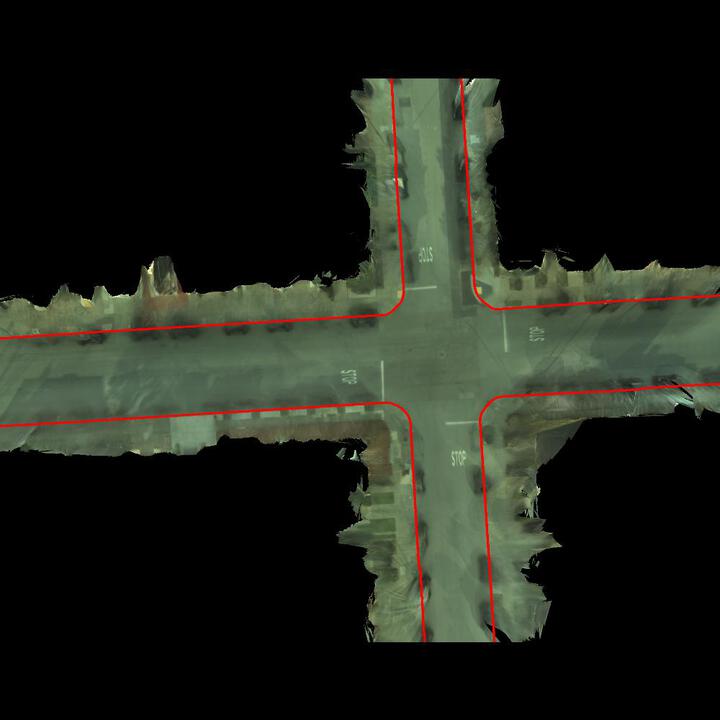}& 
	\includegraphics[width=0.38\linewidth]{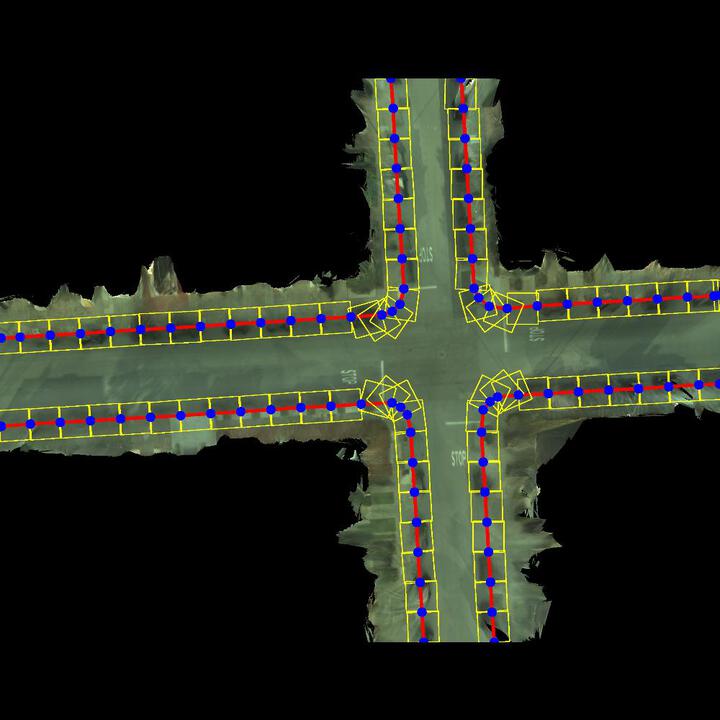} \\
	
	\includegraphics[width=0.38\linewidth]{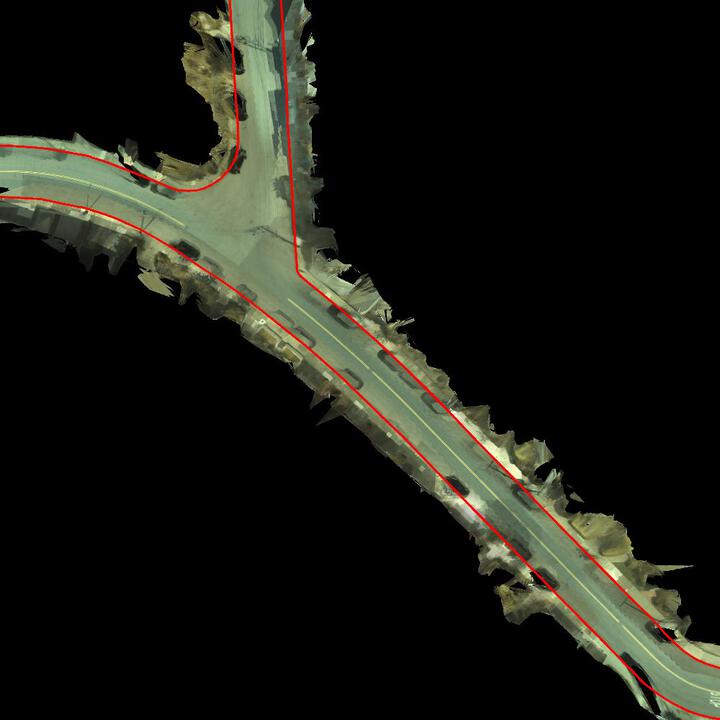}& 
	\includegraphics[width=0.38\linewidth]{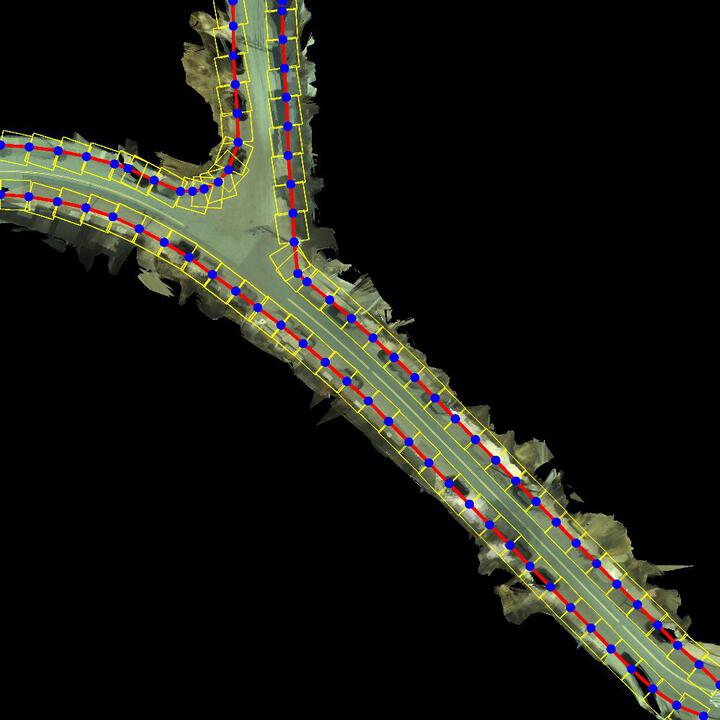} \\	

	\raisebox{2px}{{(Ground Truth)}} &
	\raisebox{2px}{{(Prediction)}} 
	
	\end{array}
	\]
	
	\caption{Results: Column \textbf{1} corresponds to the ground truth boundaries and column \textbf{2} are our predicted boundaries.}
	\label{fig:results_3}
\end{figure*}

\vspace{-2.0mm}
In Sec. \ref{sec:feat} of this supplementary material, we showcase the deep features obtained by our network in Figures \ref{fig:feat1},\ref{fig:feat2},\ref{fig:feat3} and \ref{fig:feat4}. In Sec. \ref{sec:results} we visualize the structured polyline predictions and the corresponding ground truth road boundaries in Figures \ref{fig:results_1}, \ref{fig:results_2} and \ref{fig:results_3}. Please also refer to a video we have attached for more visualizations.

\section{Deep Features}
\label{sec:feat}
\vspace{-2.0mm}
Here we visualize the inputs to our deep features model (lidar, camera, elevation gradient) and its output (detection map, end points, direction map). 
\vspace{-5.0mm}

\vspace{-1mm}

\clearpage

\clearpage

\section{Results}
\label{sec:results}
\vspace{-2.0mm}
Here we visualize the the ground truth boundaries (left column) and our corresponding predictions (right column). Our predicted polyline vertices are highlighted in blue dots and the yellow squares correspond to the rotated ROIs that our cSnake attends.  
\vspace{-4.0mm}

\vspace{-1.5mm}

\end{appendices}

\end{document}